\documentclass[10pt,twocolumn,letterpaper]{article}

\usepackage{cvpr}

\usepackage{amssymb}
\usepackage{tabularx}
\usepackage{float}
\usepackage{multirow}
\usepackage{booktabs}
\usepackage{siunitx}
\usepackage{pifont}
\usepackage{colortbl}
\usepackage{makecell}
\usepackage{soul}
\usepackage{array}
\colorlet{soulred}{yellow}

\definecolor{cvprblue}{rgb}{0.21,0.49,0.74}
\usepackage[pagebackref,breaklinks,colorlinks,allcolors=cvprblue]{hyperref}

\def\paperID{4480}
\def\confName{CVPR}
\def\confYear{2025}
\definecolor{yellow}{rgb}{1, 1, 0.7}
\definecolor{orange}{rgb}{1, 0.85, 0.7}
\definecolor{red}{rgb}{1, 0.7, 0.7}

\title{GSurf: Learning Signed Distance Fields from Splatting Opaque Gaussians for High-quality 3D Reconstruction}

\author{Baixin Xu$^1$ \quad
Jiangbei Hu$^2$ \quad
Jiaze Li$^1$ \quad
Ying He$^1$\thanks{Corresponding author: Y. He (yhe@ntu.edu.sg).}\\
$^1$College of Computing and Data Science, Nanyang Technological University\\
$^2$School of Software, Dalian University of Technology
}

\begin{document}
\maketitle

\begin{abstract}
High-fidelity surface reconstruction from multi-view images is a core problem in 3D computer vision. While neural implicit surfaces like SDFs offer smooth geometry, they are often bottlenecked by the computational intensity of volume rendering. Conversely, 3D Gaussian Splatting (3DGS) provides rapid training but lacks geometry continuity, often leading to fragmented surfaces. This paper presents a novel framework that integrates Signed Distance Fields directly into the splatting pipeline. By leveraging the continuous nature of SDFs to regularize Gaussian primitives, our method effectively fills geometric holes and suppresses noise inherent in sparse point clouds. Unlike hybrid approaches that rely on heavy volumetric sampling, our approach utilizes the efficiency of splatting to achieve faster convergence. Extensive evaluations demonstrate that our method produces high-quality surfaces with significantly fewer primitives, offering a more compact and efficient representation for both indoor and outdoor environments. The source code is available at \href{https://github.com/xubaixinxbx/Gsurf}{https://github.com/xubaixinxbx/Gsurf}.
\end{abstract}

\begin{figure*}[!htbp]
    \centering
    \includegraphics[width=0.11\linewidth]{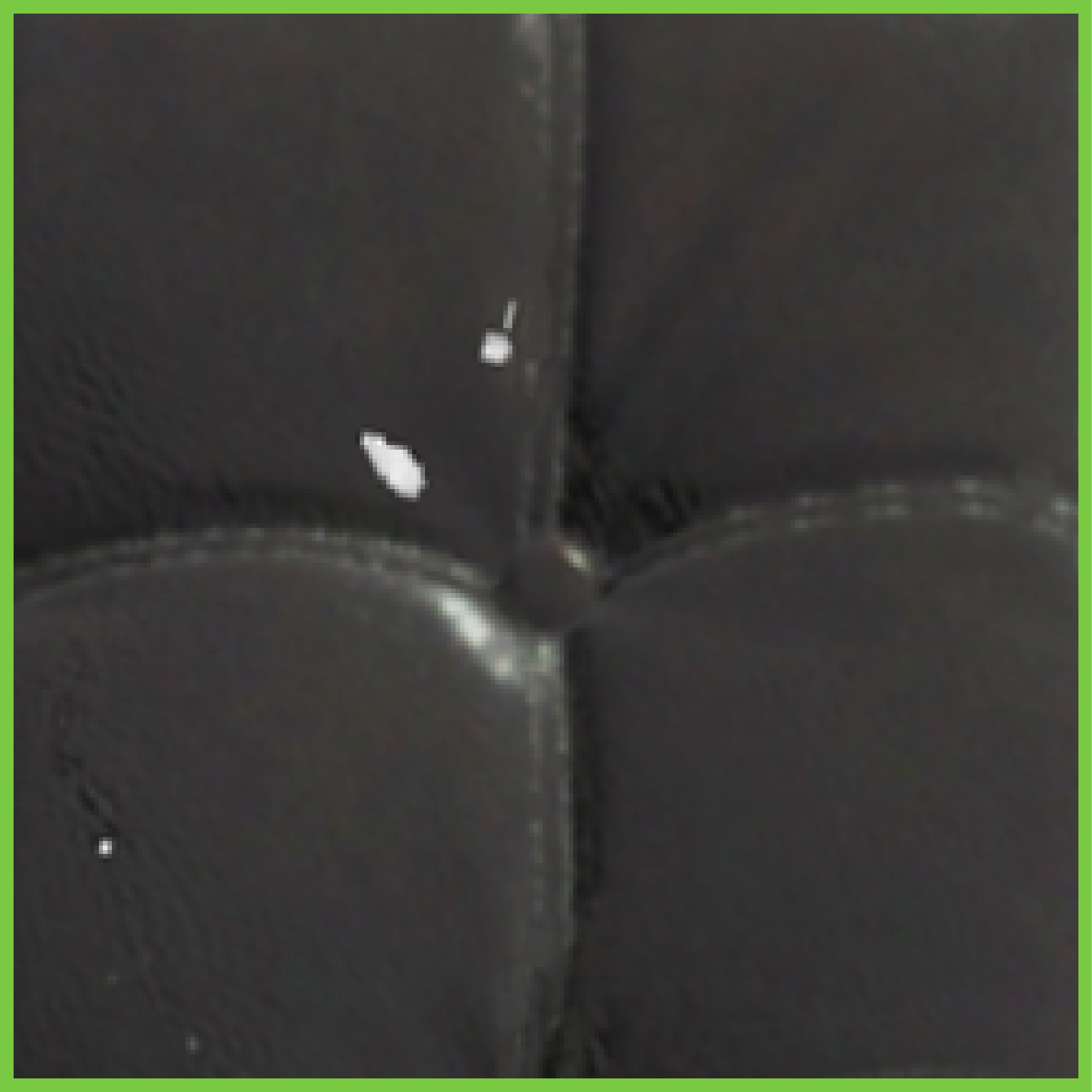}
    \includegraphics[width=0.11\linewidth]{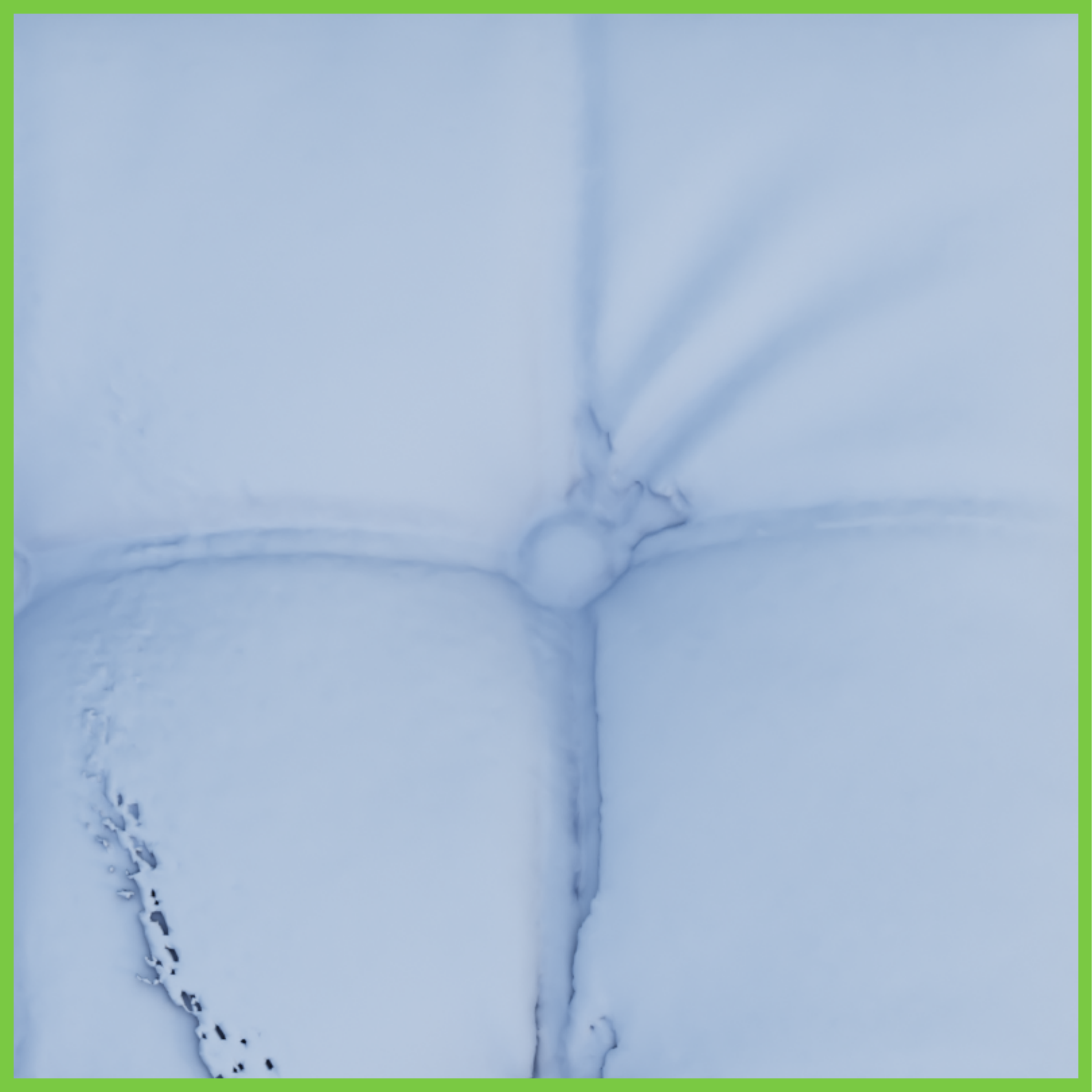}
    \includegraphics[width=0.11\linewidth]{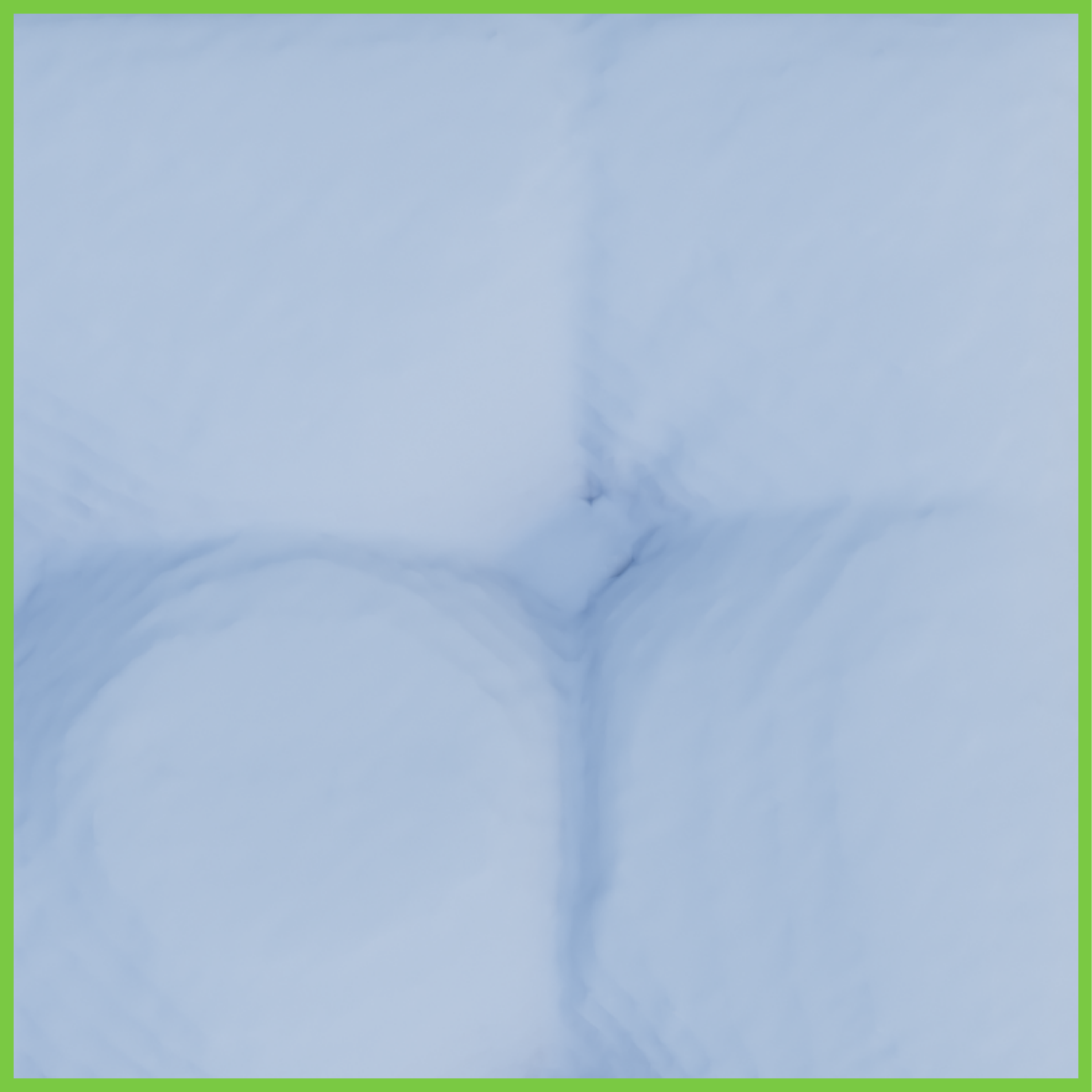}
    \includegraphics[width=0.11\linewidth]{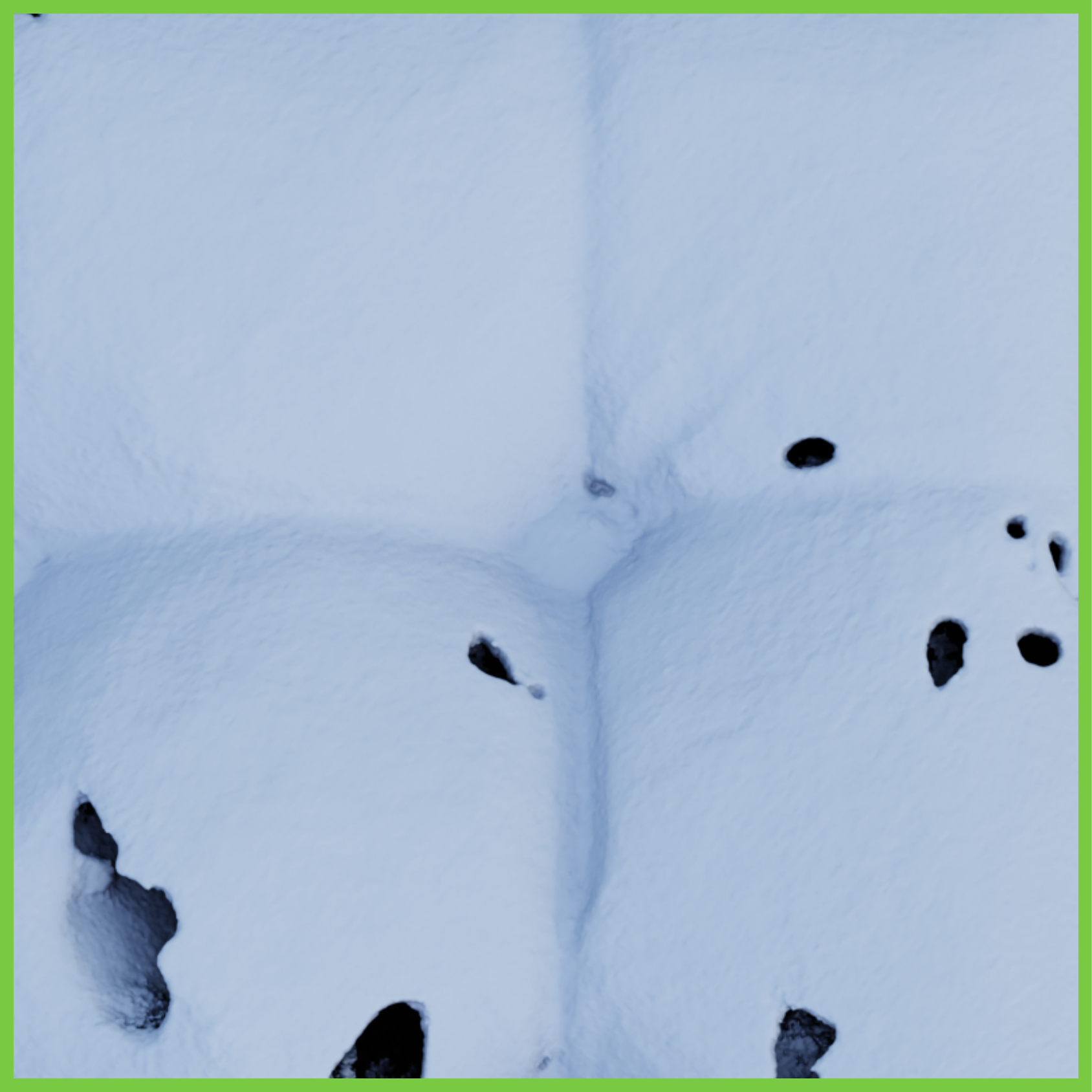}
    \includegraphics[width=0.11\linewidth]{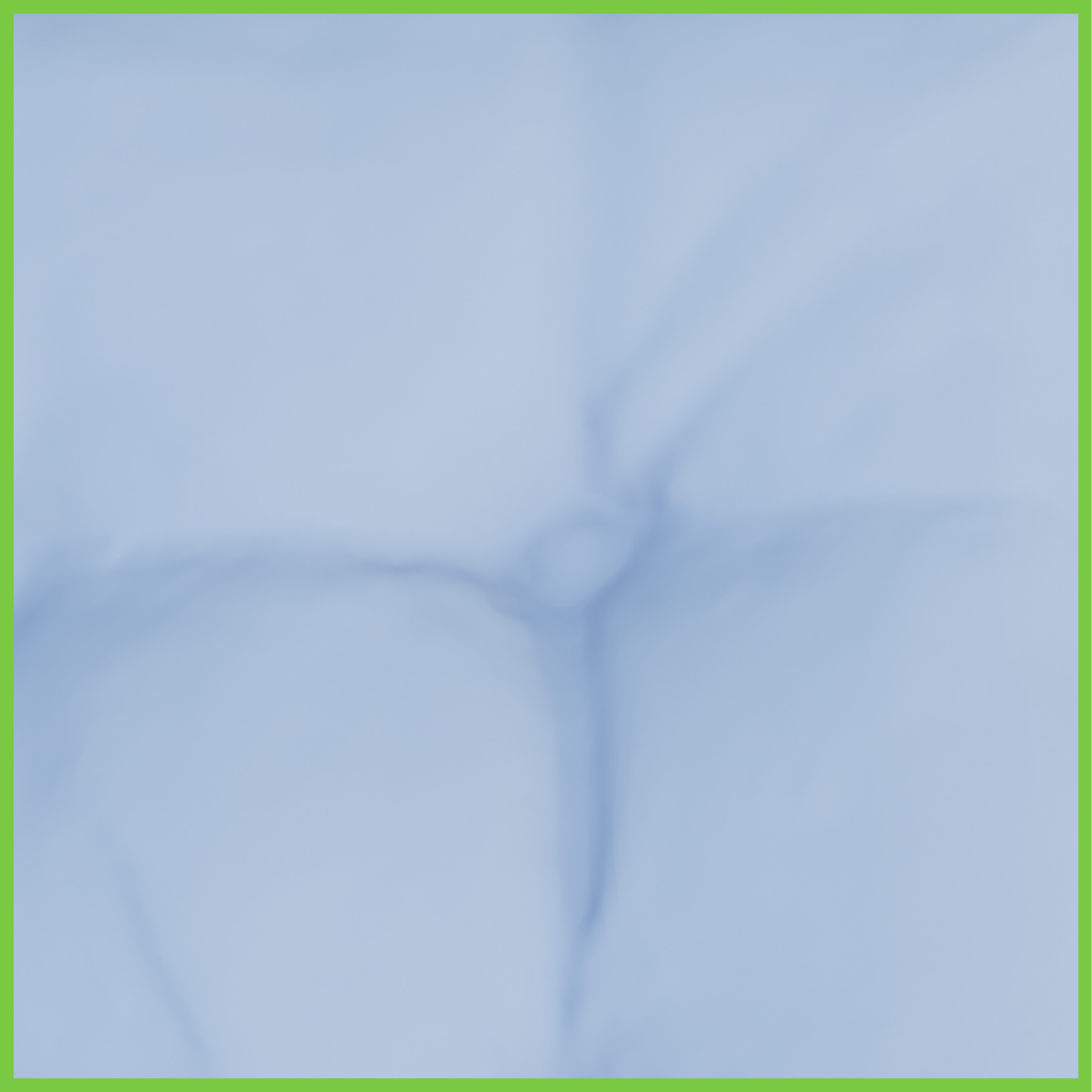}
    \includegraphics[width=0.11\linewidth]{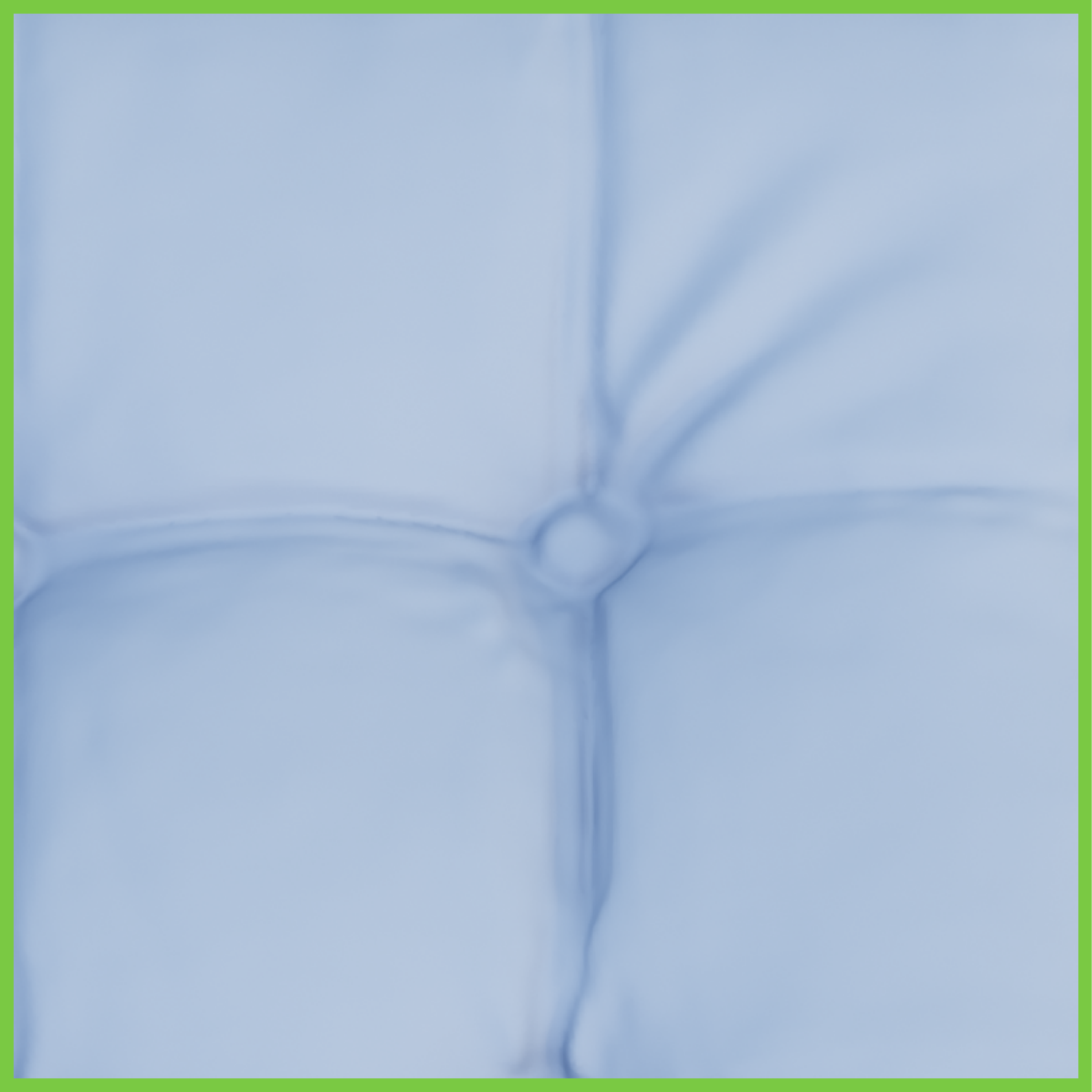}
    \includegraphics[width=0.11\linewidth]{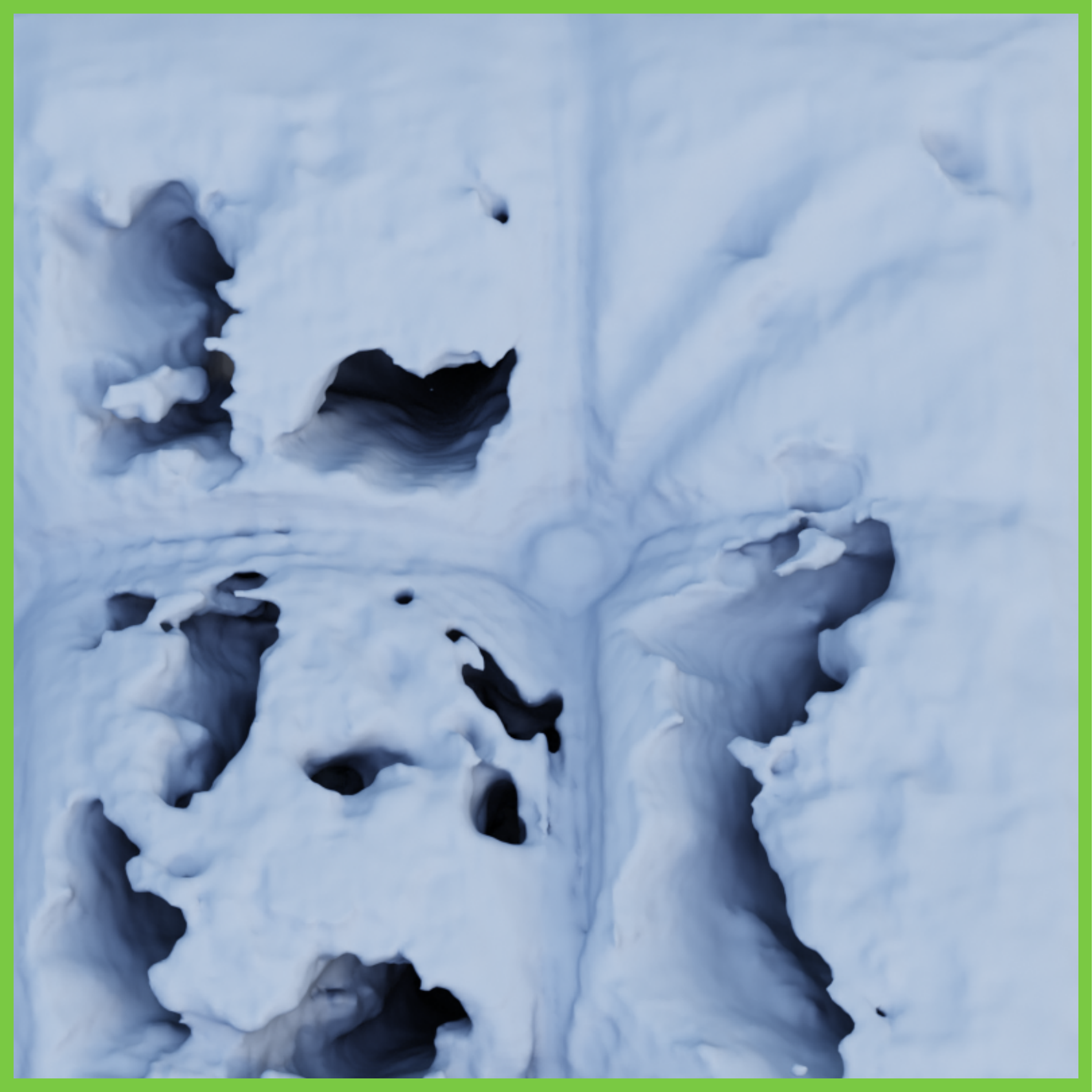}
    \includegraphics[width=0.11\linewidth]{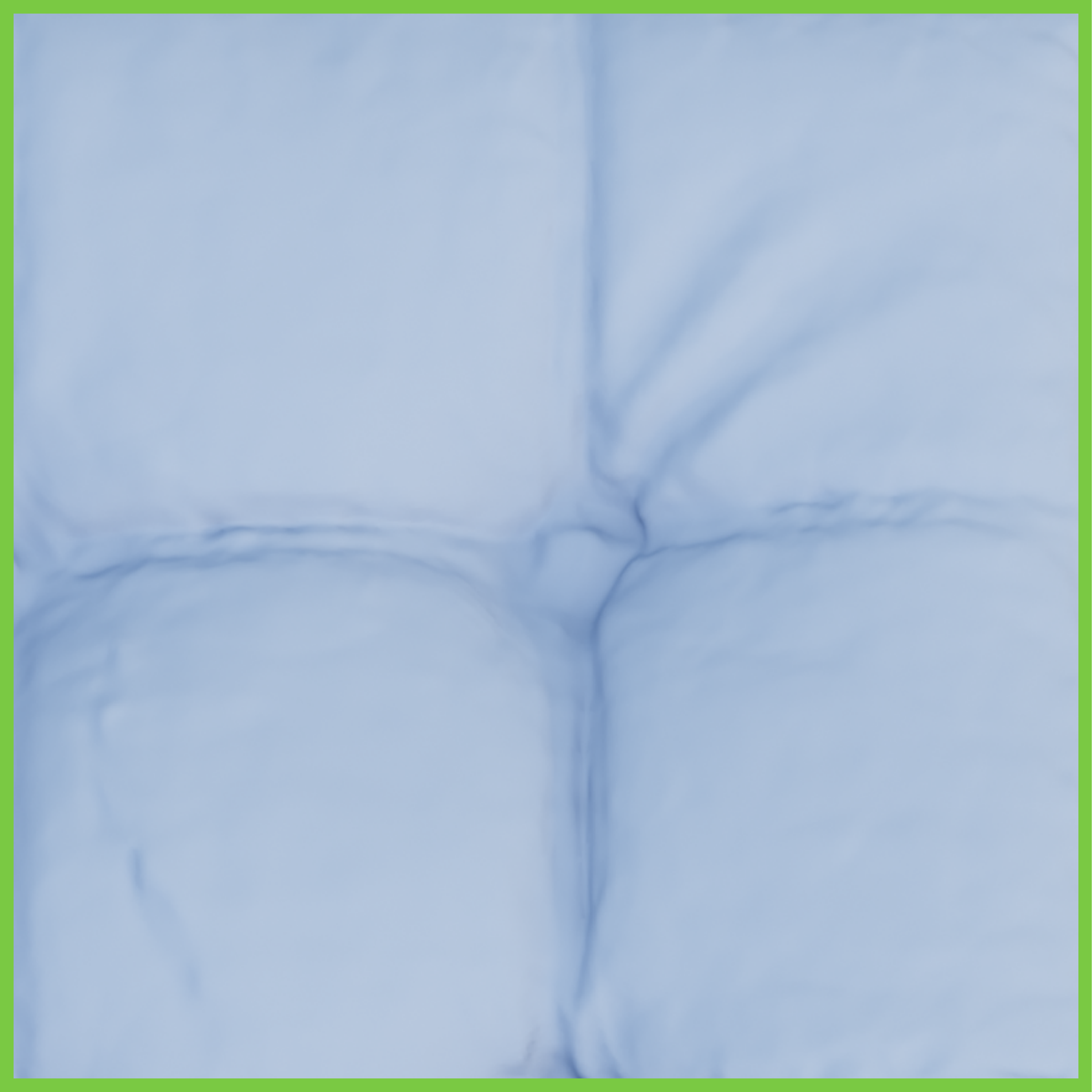}  
    \\
    \includegraphics[width=0.11\linewidth]{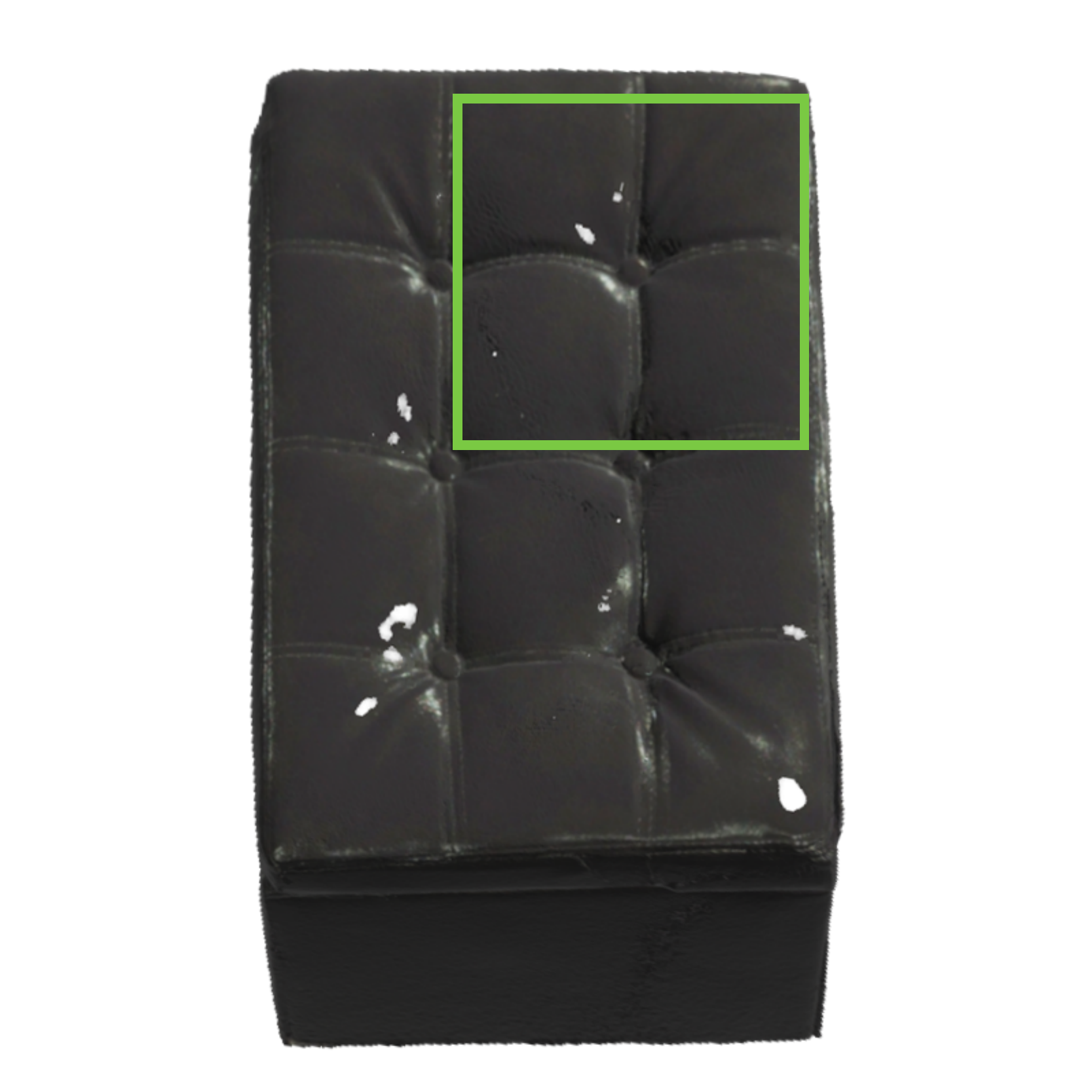}
    \includegraphics[width=0.11\linewidth]{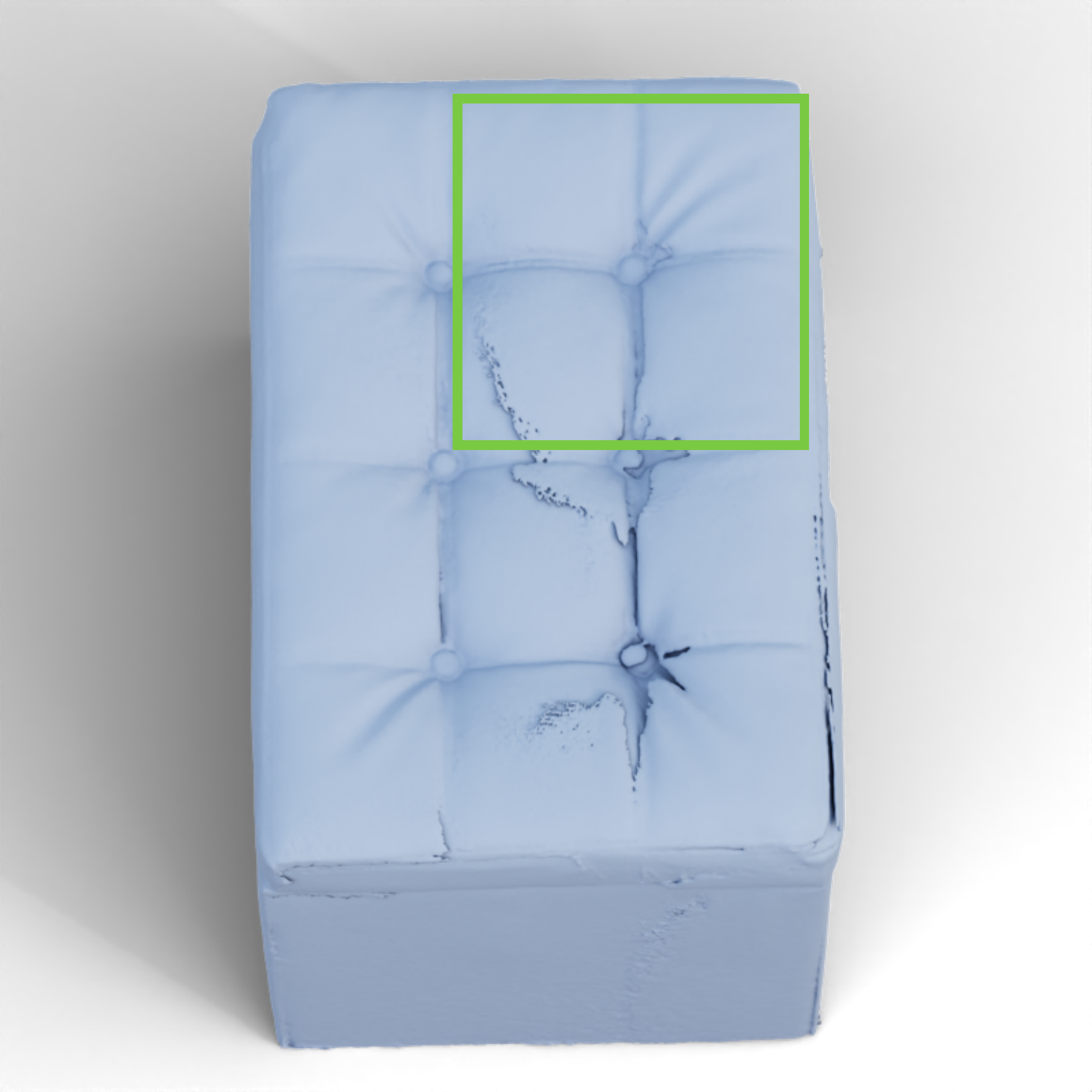}
    \includegraphics[width=0.11\linewidth]{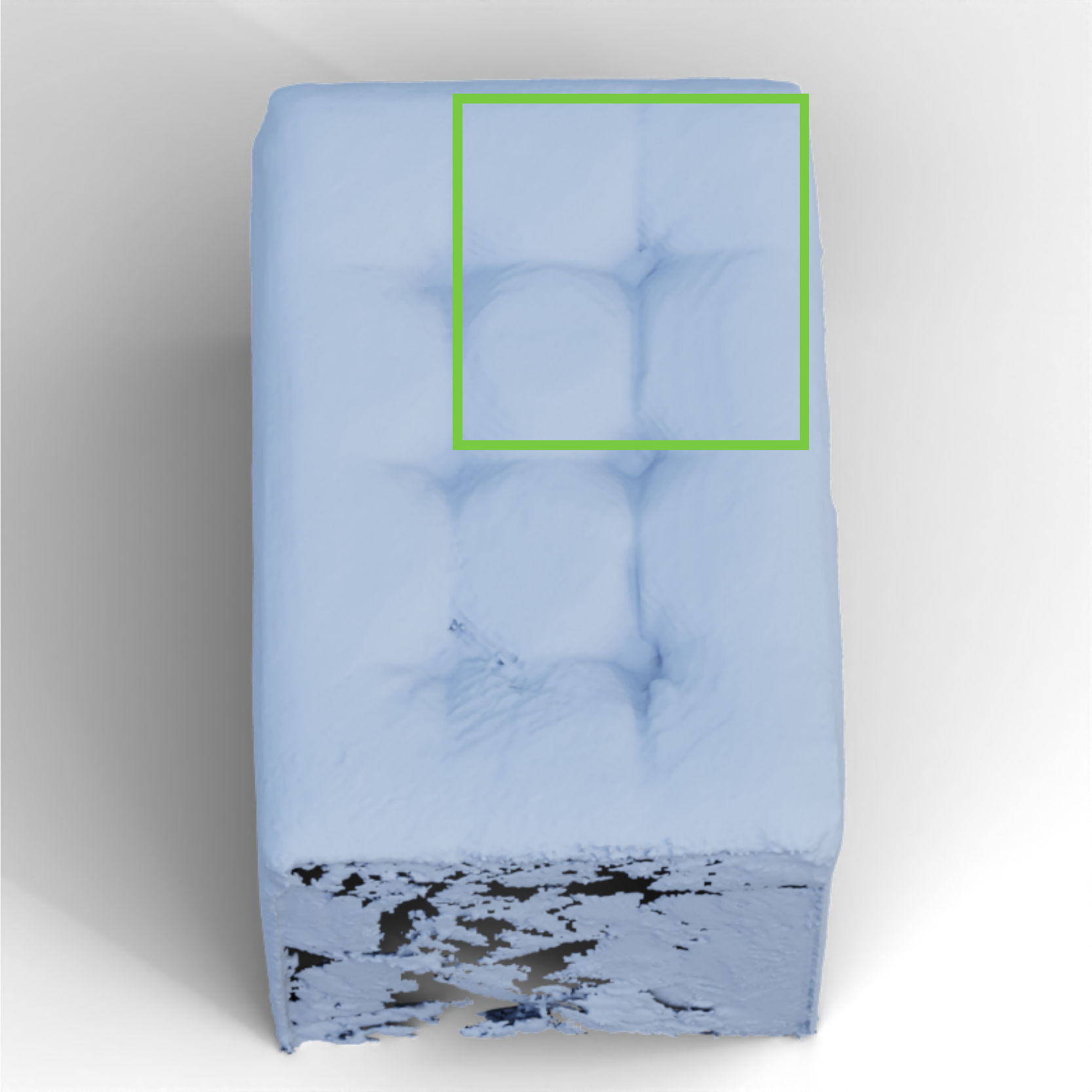}
    \includegraphics[width=0.11\linewidth]{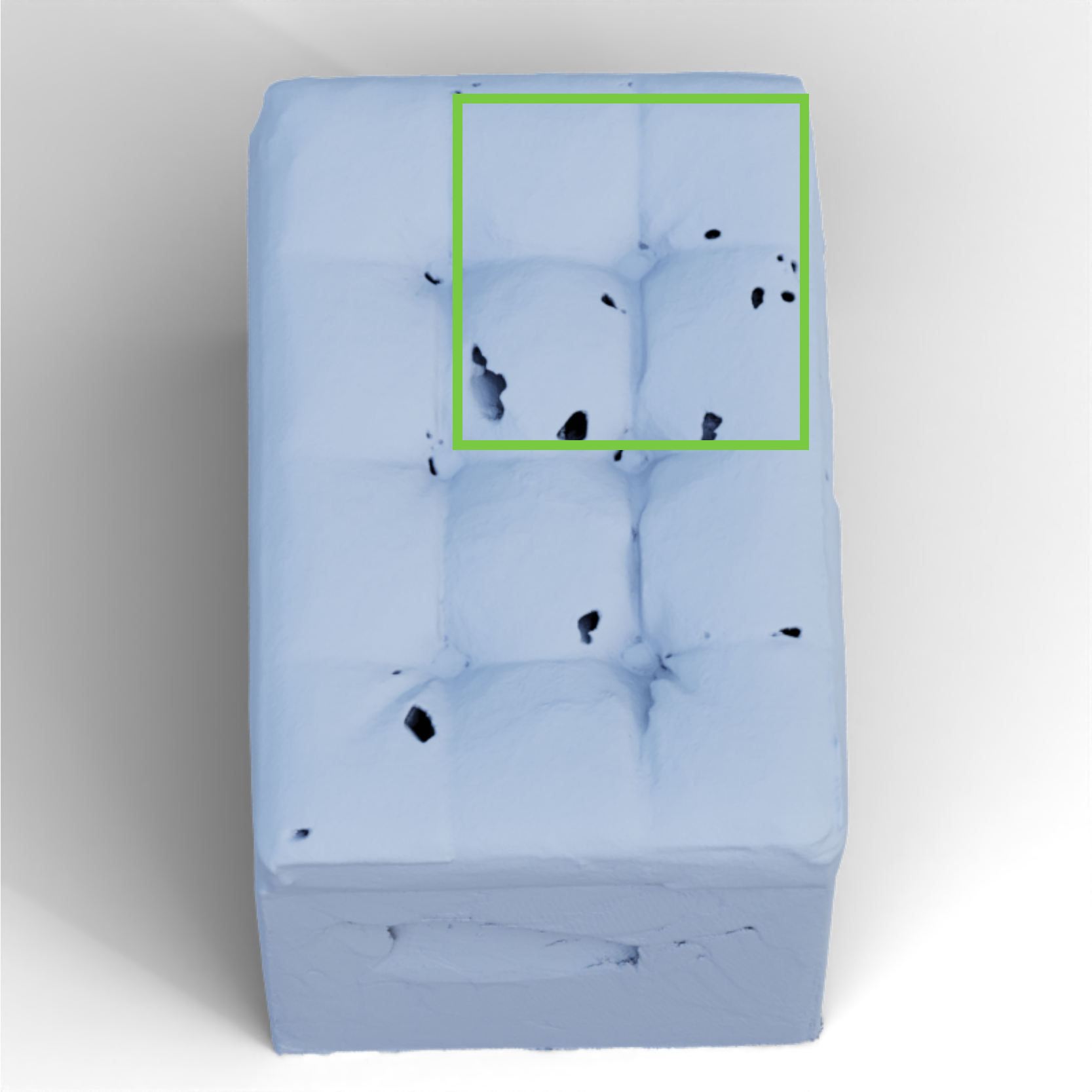}
    \includegraphics[width=0.11\linewidth]{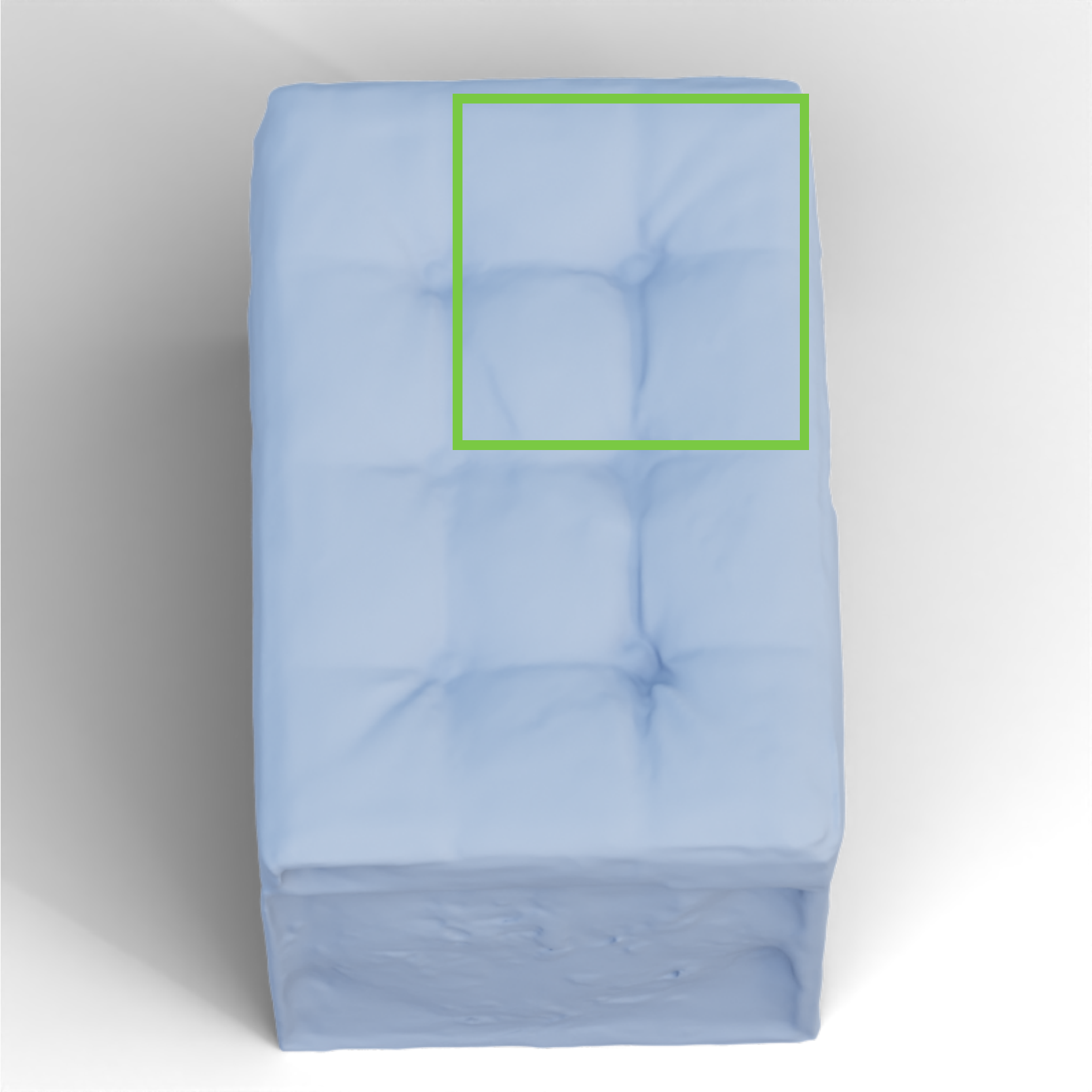}
    \includegraphics[width=0.11\linewidth]{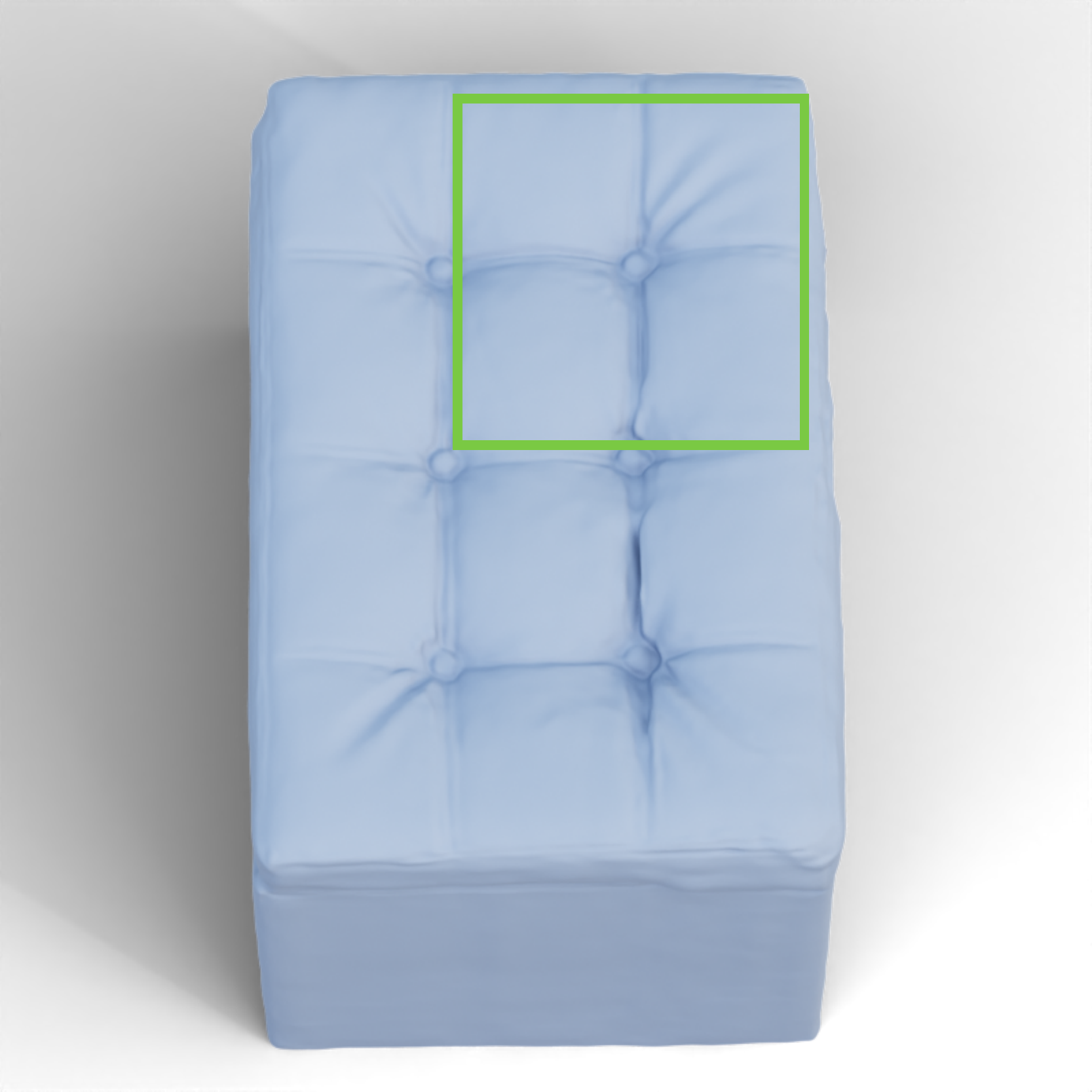}
    \includegraphics[width=0.11\linewidth]{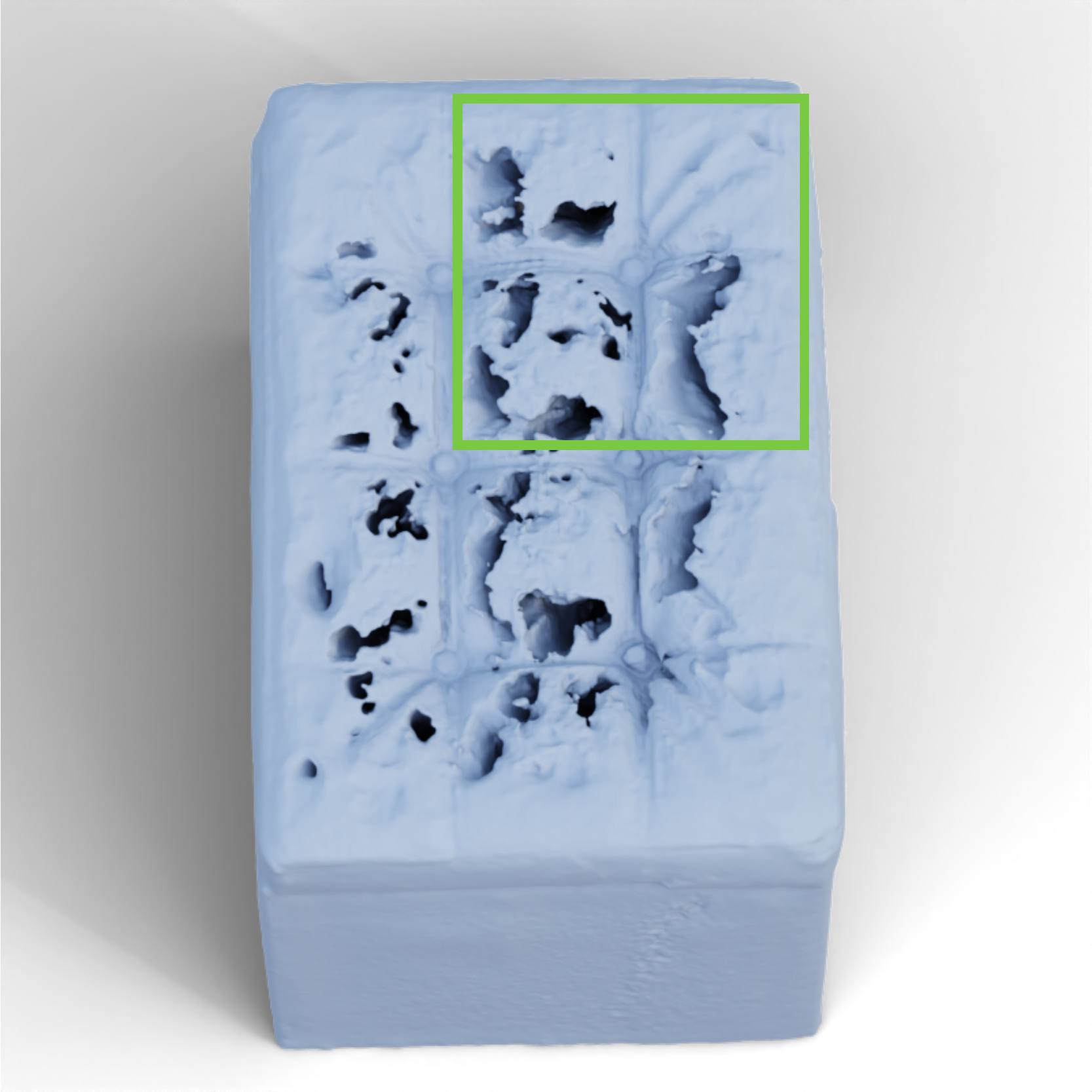}
    \includegraphics[width=0.11\linewidth]{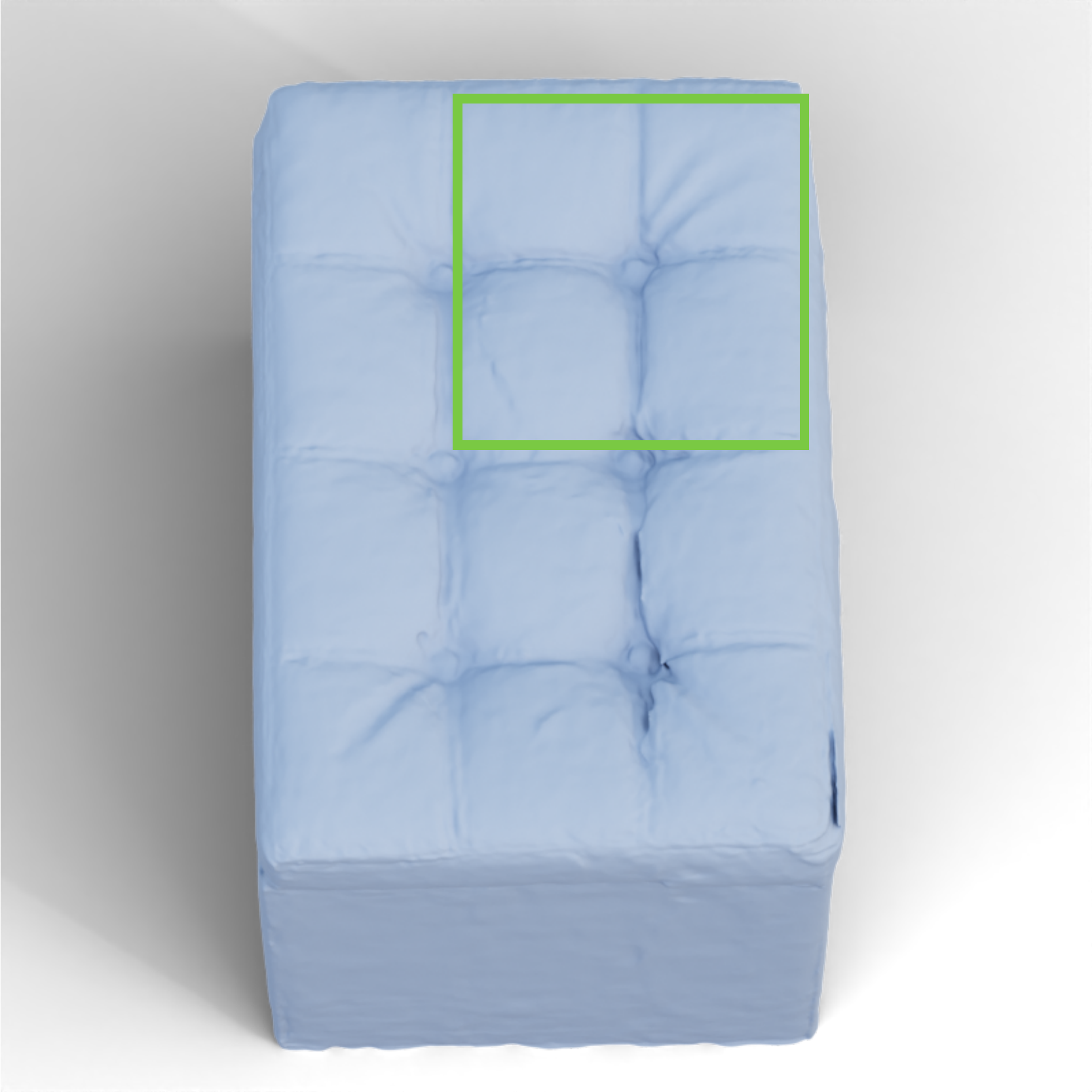}  
    \\
    \tiny
    \makebox[0.11\linewidth]{CD$(\times 10^3)\downarrow$, NC$\uparrow$}
    \makebox[0.11\linewidth]{}
    \makebox[0.11\linewidth]{16.21, 0.923}
    \makebox[0.11\linewidth]{10.25, 0.916}
    \makebox[0.11\linewidth]{17.06, 0.934}
    \makebox[0.11\linewidth]{17.90, 0.931}
    \makebox[0.11\linewidth]{12.45, 0.899}
    \makebox[0.11\linewidth]{9.10, 0.938}\\
     \makebox[0.11\linewidth]{Reference}
     \makebox[0.11\linewidth]{GT}
    \makebox[0.11\linewidth]{2DGS}
    \makebox[0.11\linewidth]{GOF}
    \makebox[0.11\linewidth]{ GaussianSurfels}
    \makebox[0.11\linewidth]{VolSDF}
    \makebox[0.11\linewidth]{Voxurf}
    \makebox[0.11\linewidth]{Ours}\\
    \caption{GSurf learns an SDF directly from Gaussian primitives, using its smoothness and continuity for robust reconstruction. It surpasses state-of-the-art GS-based methods~\cite{huang20242d,yu2024gof,dai2024gsurfel} and Voxurf~\cite{wu2022voxurf}, particularly in challenging scenarios such as varying lighting and reflective surfaces, while maintaining comparable surface smoothness and faster training than neural implicit methods like VolSDF~\cite{yariv2021volume}.}
    \label{fig:teaser}
\end{figure*}
\section{Introduction}
\label{sec:intro}

Reconstructing high-quality 3D geometry from multi-view images is a central problem in 3D vision, essential for applications such as virtual reality~\cite{ foo2023progressive}, robotics~\cite{wang2024nerf, zhong2025CoopTrack, xu2025component}, and animation, among others. Neural Radiance Fields (NeRF)~\cite{mildenhall2021nerf} have demonstrated impressive performance in novel view synthesis by combining implicit representations~\cite{mescheder2019occupancy, park2019deepsdf, saito2019pifu} with volume rendering. Numerous NeRF variants have extended its applications to geometry reconstruction~\cite{wang2021neus, yariv2021volume}, inverse rendering~\cite{jin2023tensoir, zhang2021nerfactor, zhang2023neilf++, dai2024mirres, jin2023robust}, editing~\cite{xu2024parameterization, srinivasan2025nuvo, bao2023sine, zhan2023gauge}, and generation~\cite{chan2022efficient, chan2021pigan}. However, these NeRF-based approaches encounter limitations in training and rendering efficiency due to the computational demands of implicit representations coupled with volume rendering.

To address these limitations, explicit 3D supervision and discrete representations, such as point-based methods~\cite{zhang2023pointneus, zheng2023pointavatar, chen2023neurbf}, have been introduced to improve efficiency. By transitioning from implicit to discrete representations, these approaches aim to reduce computational demands without compromising reconstruction quality. In this context, 3D Gaussian Splatting (3DGS)~\cite{kerbl20233d} introduces a discrete representation with CUDA-based rasterization, reducing training times to minutes and enabling real-time rendering. Beyond appearance modeling~\cite{cen2025segment, lin2025decoupling, zeng2025frequency}, 3DGS also enhances surface reconstruction~\cite{chen2025gigags,li2024monogsdf}. A common way for obtaining 3D geometry from 3DGS is to extract meshes by fusing rendered depth maps. This approach can achieve high-resolution geometry, but it often lacks robustness, leading to inconsistencies across different views and producing artifacts in regions with insufficient depth information. Another approach is to train an SDF network in conjunction with a volume rendering branch~\cite{chen2023neusg, yu2024gsdf}. Despite the improved geometry quality, incorporating volume rendering with Gaussian splatting introduces redundancy in the differentiable rendering pipeline, resulting in extended training times, ranging from 2 hours~\cite{yu2024gsdf} to as much as 16 hours~\cite{chen2023neusg}.

In this paper, we introduce a novel reconstruction method, GSurf, which combines SDF, a continuous representation, with 3D Gaussians, a discrete representation, achieving a balance between the accuracy of implicit representations and the computational efficiency of explicit ones. Specifically, we train the SDF directly under Gaussian supervision, eliminating the need for volume rendering and thereby maintaining high training efficiency. To improve the alignment of Gaussians with the surface, we regularize opacity using an entropy-based loss, which effectively reduces redundant transparent Gaussians. Additionally, we incorporate an MLP to integrate geometry cues--such as normals and geometric features--into the appearance modeling, moving beyond a sole reliance on spherical harmonics. This approach enhances geometric detail compared to existing GS-based reconstruction methods. \Cref{fig:teaser} shows an example of reconstructed surfaces by our method and state-of-the-art approaches.

In summary, this paper makes the following contributions:
\begin{itemize}
    \item We propose GSurf, a single-branch framework that jointly optimizes Gaussian primitives and an SDF without an additional volume-rendering branch. 
    \item We introduce direct SDF supervision form Gaussian centroids, together with opacity entropy regularization and pruning, to obtain a compact set of surface-supporting Gaussian primitives.
    \item We incorporate SDF-derived normals and geometric features into appearance modeling, improving geometric detail and view-dependent rendering quality. 
\end{itemize}

\section{Related Work}
\label{sec:relatedworks}
\paragraph{Implicit function-based reconstruction from point clouds}
Reconstructing 3D surfaces from point clouds is a fundamental problem in computer graphics that has been studied for over three decades. Classical methods typically compute an implicit function whose level set at a specified iso-value represents the target surface~\cite{hoppe1992surface,ohtake2003multi,kazhdan2006poisson,kazhdan2013screened,lin2022surface,hou2022iterative,xu2023globally,DBLP:journals/tog/LiuLCHXWWQH25}. With the rise of 3D deep learning, combining implicit functions, especially signed distance fields, with neural networks has become a mainstream approach~\cite{park2019deepsdf,Liu2021MLS,Wang_2022,On-SurfacePriors,neuralimls2023wang, hu2024if}. Common strategies for regularizing SDFs include the Eikonal loss~\cite{icml2020_2086}, the neural pull loss~\cite{NeuralPull}, and the singular Hessian loss~\cite{wang2023neuralsingular}. 

\paragraph{Multi-view reconstruction} NeuS~\cite{wang2021neus} and VolSDF~\cite{yariv2021volume} are seminal works that enable volume rendering using SDFs by mapping signed distances to densities. Using the continuity and smoothness of SDFs, these methods improve the quality of reconstruction compared to density-based methods such as NeRF~\cite{mildenhall2021nerf}. They have inspired further efforts aimed at runtime acceleration~\cite{neus2,geoneus}, improved geometric fidelity~\cite{wang2022hf}, and robustness to sparse-view input~\cite{sparseneus,xu2023deformable, komarichev2023diffsvr}. To address the computational cost of volume rendering, recent works adopt hybrid structures, such as regular grids~\cite{wu2022voxurf,sun2022direct,fridovich2022plenoxels}, octrees~\cite{li2024real,gsoctree-li2024}, triplanes~\cite{chan2022efficient,wang2023pet}, and hash grids~\cite{mueller2022instant,li2023neuralangelo,wang2023adaptive,reiser2024binary}, to replace or enhance MLPs in NeRFs.

\begin{table*}[!ht]
\centering
\setlength\tabcolsep{2pt}
\caption{\textbf{Comparison of neural surface reconstruction methods.} We explicitly contrast our method with prior works, detailing how geometric guidance is injected. While hybrid methods bottleneck efficiency via ray-marching ($O(N_{\text{ray}} N_{\text{smp}})$), both GS-Pull and our method operate exclusively on Gaussian primitives ($O(N_{\text{gs}})$). Crucially, unlike GS-Pull which relies on local pulling operations without explicit opacity constraints, we enforce an \textbf{Entropy Opacity Regularization}. This actively binarizes primitives into an opaque bounding shell, inherently satisfying conditions for stable \textbf{Eikonal} and \textbf{Normal} regularization directly at the zero-level set, avoiding the surface degradation observed in prior single-branch attempts.}
\label{tab:method_comparison}
\resizebox{\textwidth}{!}{
\begin{tabular}{l l c c c c c c}
\toprule
\textbf{Category} & \textbf{Method} & \textbf{\makecell{Computation\\Graph}} & \textbf{Primary SDF Supervision} & \textbf{SDF Regularization} & \textbf{\makecell{MLP Queries\\(Complexity)}} & \textbf{Opacity Reg.} & \textbf{Appearance} \\
\midrule
NeRF-based & VolSDF~\citep{yariv2021volume} & Vol.\ Rend. & Volumetric Weights (Ray) & Eikonal & $O(N_{\text{ray}} N_{\text{smp}})$ & N/A & Imp.\ (Cond.) \\
\midrule
\multirow{6}{*}{GS-based} 
 & 2DGS~\citep{huang20242d} & \multirow{6}{*}{Splatting} & \multirow{6}{*}{N/A} & \multirow{6}{*}{N/A} & \multirow{6}{*}{None} & None & \multirow{6}{*}{Exp.\ (Decoupled)} \\
 & GOF~\citep{yu2024gof} & & & & & Filter & \\
 & GSurfels~\citep{dai2024gsurfel} & & & & & Bell curve & \\
 & PGSR~\citep{chen2024pgsr} & & & & & None & \\
 & RTG-SLAM~\citep{peng2024rtg} & & & & & Fixed Value & \\
 & VCR-GauS~\citep{chen2024vcr} & & & & & None & \\
\midrule
\multirow{5}{*}{GS+SDF} 
 & NeuSG~\citep{chen2023neusg} & Dual-Branch & Vol. Weights + Aux. Centroids & Eikonal + Normal & $O(N_{\text{ray}} N_{\text{smp}})$ & None & Exp.\ (Decoupled) \\
 & GSDF~\citep{yu2024gsdf} & Dual-Branch & Volumetric Weights (Ray) & Eikonal + Normal & $O(N_{\text{ray}} N_{\text{smp}})$ & None & Exp.\ (Decoupled) \\
 & 3DGSR~\citep{lyu20243dgsr} & Dual-Branch & Volumetric Weights (Ray) & Eikonal + Normal & $O(N_{\text{ray}} N_{\text{smp}})$ & None & Exp.\ (Decoupled) \\
 \cmidrule{2-8}
 & GS-Pull~\citep{zhang2024gspull} & Splatting & Distance to Gaussian Disk & Neural Pulling & $O(N_{\text{gs}})$ & None & Exp.\ (Decoupled) \\
 & \textbf{Ours} & \textbf{Splatting} & \textbf{Zero-level Set at Centroids} & \textbf{Eikonal + Normal} & $\textbf{\ensuremath{O(N_{\text{gs}})}}$ & \textbf{Entropy} & \textbf{Imp.\ (Cond.)} \\
\bottomrule
\end{tabular}
}
\end{table*}

\paragraph{Reconstruction with 3DGS}
3D Gaussian Splatting~\cite{kerbl20233d} was originally proposed for real-time novel view synthesis, enabling fast training and rendering via CUDA-based rasterization. However, it does not explicitly reconstruct surface geometry. To address this, 2D Gaussian Splatting (2DGS)~\cite{huang20242d} uses flat 2D disks to better align with scene surfaces.

Most existing works aim to improve depth accuracy~\cite{wolf2024gs2mesh,chen2024vcr,chen2024pgsr,zhang2024spiking} and extract meshes via TSDF fusion. Other methods such as SuGaR~\cite{guedon2024sugar} and GaussianSurfels~\cite{dai2024gsurfel} apply Poisson reconstruction~\cite{kazhdan2006poisson,kazhdan2013screened}. However, geometry extracted from depth maps typically suffers from holes, surface noise, and poor smoothness. Moreover, the recovered geometry cannot directly influence the training of Gaussians. 
GOF~\cite{yu2024gof} bypasses TSDF or Poisson fusion by directly extracting geometry as a level set from the Gaussian opacity field. Several recent works~\cite{Wu2024gsrec,lyu20243dgsr,yu2024gsdf,xiang2025gaussianroom} incorporate SDFs into Gaussian splatting, using Marching Cubes for smoother geometry.

As summarized in \Cref{tab:method_comparison}, most 3DGS-based methods either (1) decouple geometry and appearance, preventing RGB loss from directly supervising geometry, or (2) combine splatting with volume rendering, leading to redundant computation and increased training cost. In contrast, our method reconstructs SDFs using an entropy-based opacity regularization, which effectively reduces semi-transparent Gaussian primitives. Furthermore, by incorporating both viewing direction and geometry cues into appearance rendering, GSurf achieves sharper detail without requiring a dual rendering pipeline. Compared to GS-Pull~\cite{zhang2024gspull}, a state-of-the-art reconstruction model that deforms point clouds toward a learned SDF but often produces oversmoothed surfaces, our geometry-guided appearance modeling yields more accurate and detailed reconstructions.

Several recent methods~\cite{Wu2024gsrec, lyu20243dgsr, yu2024gsdf, xiang2025gaussianroom} integrate SDFs into Gaussian splatting, extracting meshes with Marching Cubes to produce smoother and more complete geometry. We note that most 3DGS methods either decouple geometry and appearance, preventing direct backpropagation from RGB loss to geometry, or combine Gaussian splatting and volume rendering, introducing redundant computation and slowing down training. In contrast, our approach reconstructs SDFs using an opacity prior and models appearance with both view and geometry cues, achieving finer detail without dual rendering. Compared to GS-Pull~\cite{zhang2024gspull}, which pulls points toward a learned SDF and tends to oversmooth, our method directly integrates geometric cues into appearance, producing sharper and more accurate reconstructions.

\section{Method}
\begin{figure*}[!htbp]
    \centering
    \includegraphics[width=0.975\linewidth]{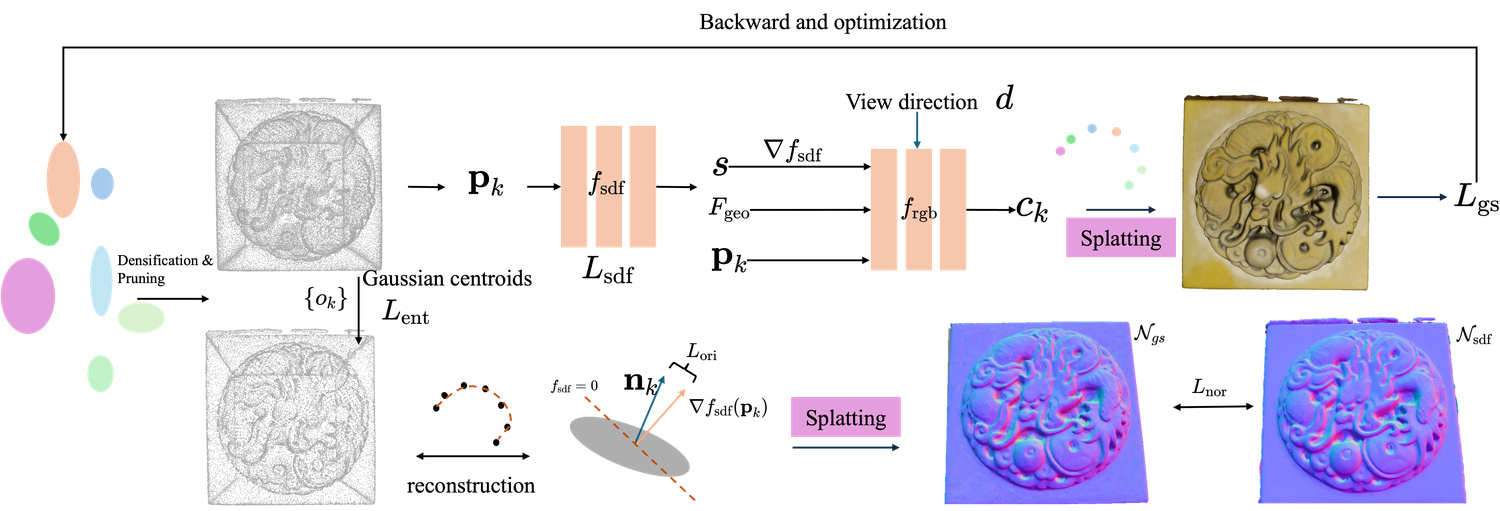}  
    \caption{Algorithmic pipeline. During the optimization of Gaussian primitives, we feed Gaussian centroids to learn an SDF supervised by the $\textbf{L}_\text{sdf}$ while simultaneously regularizing opacities with an entropy loss $\textbf{L}_\text{ent}$. Gaussian positions, normals, geometric features, and viewing directions are then fed into the appearance module to obtain view-dependent colors via the image reconstruction loss $\textbf{L}_\text{gs}$. Finally, the 3D Gaussians are rasterized into image space to produce images.}
    \label{fig:pipeline}
\end{figure*}

\subsection{Overview}
Given a set of posed multi-view RGB images, denoted as $\mathcal{I}$, our objective is to reconstruct the following components:
\begin{itemize}
    \item A set of Gaussian primitives, each characterized by opacity, scale, and rotation, collectively defining the object in a discrete manner;
    \item A signed distance field, which defines the object's geometry in a continuous manner; and 
    \item An appearance field, representing the object's radiance. 
\end{itemize} 
Both the signed distance field and the appearance field are parameterized using MLPs. In the original 3DGS rendering pipeline, Gaussian primitives are not required to lie exactly on the surface, and off-surface primitives, which are semi-transparent, can still contribute to rendering. While these off-surface primitives are acceptable for rendering purposes, they can adversely affect 3D reconstruction and may introduce significant distortions in the reconstructed surface. Therefore, a primary goal in GSurf is to minimize the number of off-surface primitives. To achieve this, we regularize the Gaussian primitives using an opacity entropy loss, which drives opacities toward binary value, and periodic pruning removes the low-opacity branch; consequently, the retained primitives form a compact, mostly opaque surface-supporting set. \Cref{fig:pipeline} illustrates the algorithmic pipeline of GSurf, with technical details provided in the following subsections.

\subsection{Gaussian Splatting}
\label{subsec:gs}
We adopt 2DGS~\cite{huang20242d} for rendering, which represents 3D objects with a set of 2D Gaussian primitives (i.e. elliptical disks). Each primitive is parameterized by a central position $\mathbf{p}_k \in \mathbb{R}^3$, two principal tangent vectors $\mathbf{t}_u, \mathbf{t}_v \in \mathbb{R}^3$, and two scaling factors $s_u, s_v \in \mathbb{R}$. Let $\mathbf{R}_k=[\mathbf{t}_u,\mathbf{t}_v,\mathbf{t}_u\times\mathbf{t}_v]$ be the rotation matrix, and $\mathbf{S}_k=\mathrm{diag}(s_u,s_v,0)$ be the diagonal matrix for scaling. This setup defines a 2D Gaussian within a local tangent plane in the  world coordinate system as:
\begin{equation}
\nonumber 
    \mathbf{P}(u,v)=\mathbf{p}_{k}+s_{u}\mathbf{t}_{u} u+s_{v}\mathbf{t}_{v}v =\mathbf{H}(u, v, 1,1)^{T},
\end{equation}
where \begin{displaymath}
\mathbf{H}=\left[
\begin{array}{cc}
\mathbf{R}_k\mathbf{S}_k & \mathbf{p}_k\\
\mathbf{0} & 1
\end{array}
\right]
\end{displaymath}
is a $4\times 4$ transformation matrix. For each point $\mathbf{u}=(u,v)^T$, its 2D Gaussian value is computed using the standard Gaussian formula:
\begin{equation}
\nonumber 
    \mathcal{G}(\mathbf{u}) = \exp\left(-\frac{u^2+v^2}{2}\right).
\end{equation}
These 2D Gaussians are then sorted based on the depth of their centers. To render a pixel $x$, we cast a ray into the scene and accumulate color using volumetric alpha blending as follows:
\[
    \mathbf{c}(x)=\sum_{i=1} w_i \mathbf{c}_i,
\]
where \[w_i = o_i \mathcal{G}_i\left(\mathbf{u}(x)\right) \prod_{j=1}^{i-1}\left(1-o_j \mathcal{G}_j(\mathbf{u}(x))\right),
\]
and $o_i$ represents the opacity.

Supervised with a color loss $L_\text{rgb}$, 2DGS also introduces regularization terms for depth distortion $L_{\text{dep}}$ and depth-normal consistency $L_{\text{dnc}}$, formulated as:
\begin{equation}
    L_\text{gs} = L_\text{rgb} + L_\text{dep} +  L_\text{dnc}.
\end{equation}
We refer the reader to~\cite{huang20242d} for more details about these terms.

It is important to note that we have made two modifications to 2DGS to tailor it to our task. First, in our setting, the color $\mathbf{c}_i$ is not represented by spherical harmonics (SH) as in vanilla-GS~\cite{kerbl20233d, huang20242d}; instead, we use an MLP to predict $\mathbf{c}_i$. Second, we require that the majority of Gaussian disks have high opacity, ensuring they align closely with the surface of the target object. This differs from 2DGS and 3DGS, which typically contain many semi-transparent primitives spread throughout the entire space. Details on the appearance modeling and the regularization strategy for opacity are provided in a later section.

\begin{table*}[!htbp]
    \centering
    \caption{Quantitative evaluation on the DTU dataset. We compare our method with volume rendering-based and Gaussian splatting-based reconstruction methods using CD ($\times10^3, \downarrow$).}
    \label{tab:dtu}
    \resizebox{\textwidth}{!}{\begin{tabular}{c|ccccccccccccccc|c}
        \toprule
        Method & 24 & 37 & 40 & 55 & 63 & 65 & 69 & 83 & 97 & 105 & 106 & 110 & 114 & 118 & 122 & Mean\\
        \midrule
        NeuS~\cite{wang2021neus} & 1.00 & 1.37 & 0.93 & 0.43 & 1.10 &  0.65 &  0.57 & 1.48 &  1.09 & 0.83 &  0.52 &  1.20 &  0.35 &  0.49 & 0.54 & 0.84 \\
        VolSDF~\cite{yariv2021volume} & 1.14 & 1.26 & 0.81 & 0.49 & 1.25 & 0.70 & 0.72 & 1.29 & 1.18 & 0.70 & 0.66 & 1.08 & 0.42 & 0.61 & 0.55 & 0.86 \\
        3DGS~\cite{kerbl20233d} & 2.14 & 1.53 & 2.08 & 1.68 & 3.49 & 2.21 & 1.43 & 2.07 & 2.22 & 1.75 & 1.79 & 2.55 & 1.53 & 1.52 & 1.50 & 1.97 \\
        SuGaR~\cite{guedon2024sugar} & 1.47 & 1.33 & 1.13 & 0.61 & 2.25 & 1.71 & 1.15 & 1.63 & 1.62 & 1.07 & 0.79 & 2.45 & 0.98 & 0.88 & 0.79 & 1.32 \\
        DN-Splatter~\cite{turkulainen2024dnsplatter} & 1.60 & 2.03 & 1.42 & 1.44 & 2.37 & 2.11 & 1.62 &  1.95 & 1.88 & 1.48 & 1.63 & 1.82 & 1.20 & 1.50 & 1.40 & 1.70 \\
        2DGS~\cite{huang20242d} & 0.48 & 0.91 & 0.39 & 0.39 & 1.01 & 0.83 & 0.81 & 1.36 & 1.27 & 0.76 & 0.70 & 1.40 & 0.40 & 0.76 & 0.52 & 0.80 \\
        GOF~\cite{yu2024gof}  & 0.50 & 0.82 & 0.37 & 0.37 & 1.12 & 0.74 & 0.73 & 1.18 & 1.29 & 0.68 & 0.77 & 0.90 & 0.42 & 0.66 & 0.49 & 0.74  \\
        GSurfels~\cite{dai2024gsurfel} & 0.66 & 0.93 & 0.54 & 0.41 & 1.06 & 1.14 & 0.85 & 1.29 & 1.53 & 0.79 & 0.82 & 1.58 & 0.45 & 0.66 & 0.53 & 0.88 \\
        Ours & 0.52 & 1.00 & 0.48 & 0.47 & 0.79 & 1.27 & 0.82 & 1.24 & 1.27 & 0.65 & 0.89 & 1.26 & 0.52 & 0.76 & 0.73 & 0.84 \\
        \bottomrule
    \end{tabular}}
\end{table*}

\subsection{Learning SDF from Gaussians}
\label{subsec:sdf}

Inspired by recent advances in shape representation using SDF~\cite{wang2023neuralsingular, On-SurfacePriors, park2019deepsdf, neuralimls2023wang, Liu2021MLS,Wang_2022} and their application in neural rendering (e.g., NeuS~\cite{wang2021neus} and VolSDF~\cite{yariv2021volume}), we propose to fit an SDF directly to the Gaussian centroids. This offers two key advantages: (1) It enables geometric regularization using SDF-based losses, such as the Eikonal loss, which helps address issues like holes and surface irregularities commonly observed in TSDF fusion. (2) The learned SDF provides rich geometric cues that enhance Gaussian appearance modeling, unlike TSDF-based methods that rely solely on opacity and spherical harmonics and cannot directly affect appearance learning during training.

Unlike NeuS-based methods, which are purely trained from 2D images and lack direct 3D supervision, our approach with Gaussians provides explicit 3D supervision, similar to surface reconstruction from point clouds. Given the central positions of the Gaussians, ${\mathbf{p}_k}$, as discrete representations of the 3D object, our goal is to learn an SDF parameterized by an MLP $f_\text{sdf}$. 

Specifically, we impose the following conditions on the SDF:
\begin{itemize}
    \item The position constraint $L_\text{pos}= \sum_k |f_\text{sdf}(\mathbf{p}_k)|$ encourages the zero-level set to align with the Gaussians.
    \item The Eikonal constraint $ L_\text{eik} = \Sigma_{\mathbf{x}\in\chi_{\text{eik}}} (\| \nabla f_\text{sdf}(\mathbf{x})\|_2 - 1)^2$ enforces unit gradient norm and promotes SDF smoothness, where $\chi_{\text{eik}}$ is a set of samples in the space. \item Off-surface constraint: $L_\text{off} = \sum_k \exp{ \left(-\alpha |f_\text{sdf}(\mathbf{q}_k)| \right) }$, applied to query points $\mathbf{q}_k$ that are uniformly sampled in the space. This constraint ensures function values for off-surface points remain relatively large, keeping them away from zero. We set $ \alpha=100$ in our implementation. \item Orientation constraint: $L_\text{ori} = 1 - \left|\mathbf{n}_k \cdot \frac{\nabla f_\text{sdf}(\mathbf{p}_k)}{\|\nabla f_\text{sdf}(\mathbf{p}_k)\|} \right|$, which enforces alignment between the Gaussian orientation $\mathbf{n}_k$ and the SDF's gradient. Here, $\mathbf{n}_k$ is defined by the shortest axis of the disk.
\item  Normal map constraint: $L_\text{nor}=1-\mathcal{N}_{\text{gs}}\cdot\mathcal{N}_{\text{sdf}}$, which minimizes the discrepancy between the normal maps $\mathcal{N}_{\text{sdf}}$ and $\mathcal{N}_{\text{gs}}$ rendered from the SDF and 2DGS.
\end{itemize}

Combining all terms, the total SDF loss is defined as:
\begin{equation}
    L_\text{sdf} = \lambda_1 L_\text{pos} + \lambda_2 L_\text{eik} + \lambda_3 L_\text{off} + \lambda_4 L_\text{ori} + \lambda_5 L_\text{nor}.
\end{equation}

\begin{figure*}[!htbp]
    \centering
    \includegraphics[width=0.11\linewidth]{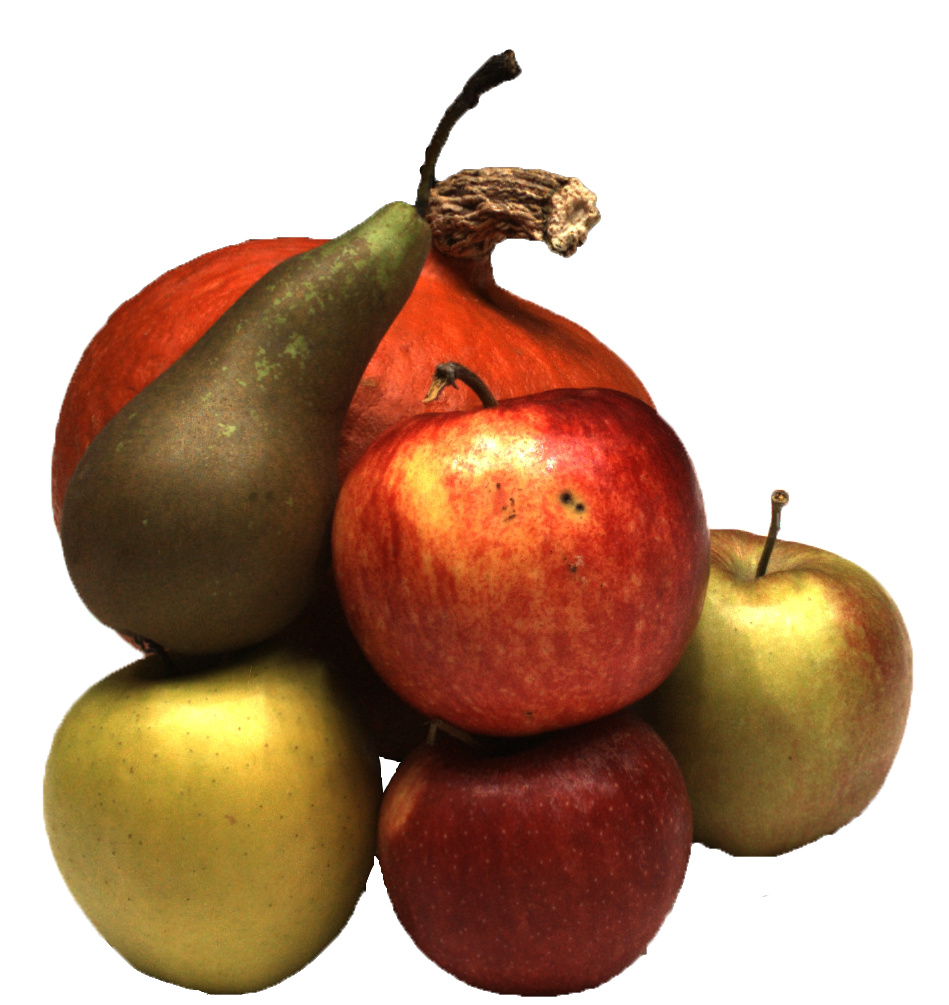}
    \includegraphics[width=0.11\linewidth]{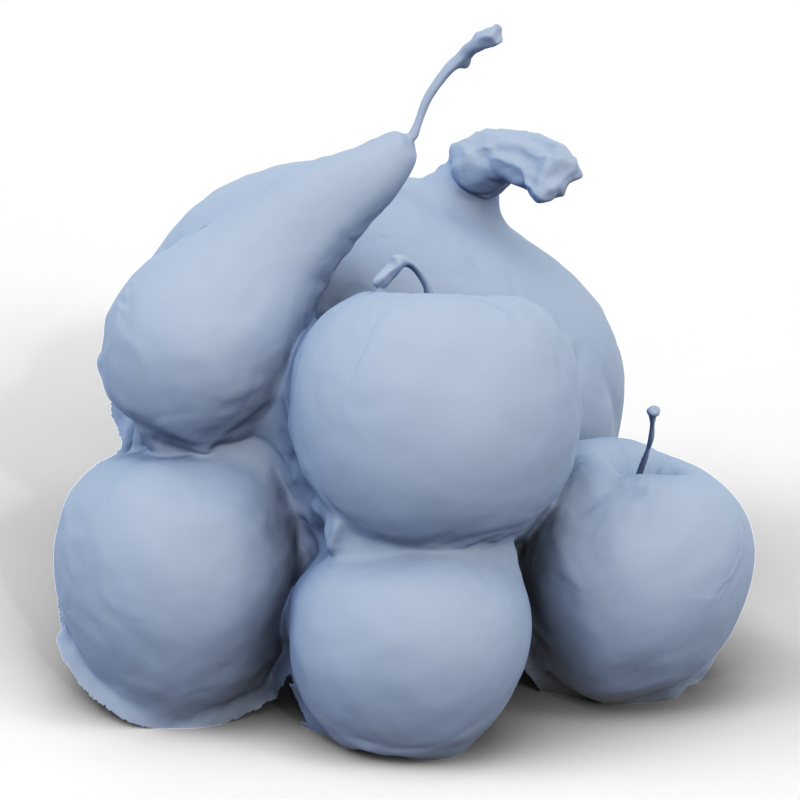}
    \includegraphics[width=0.11\linewidth]{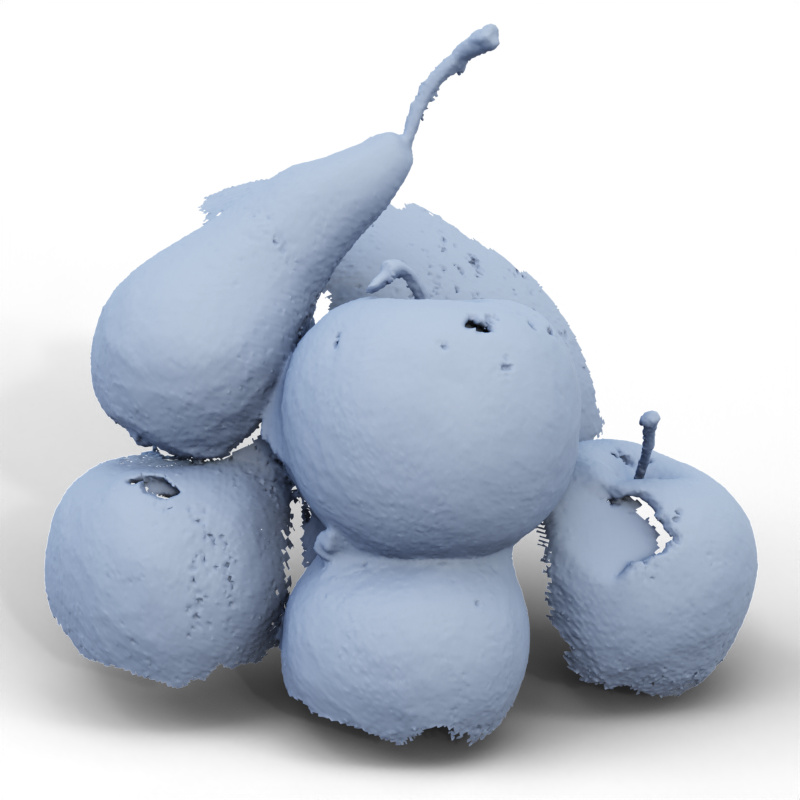}
    \includegraphics[width=0.11\linewidth]{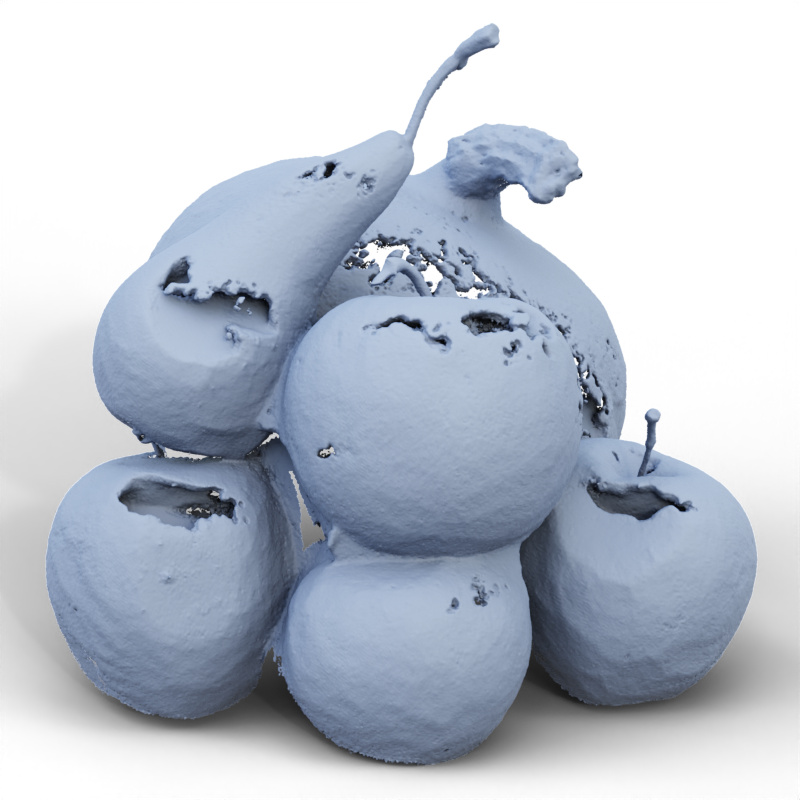}
    \includegraphics[width=0.11\linewidth]{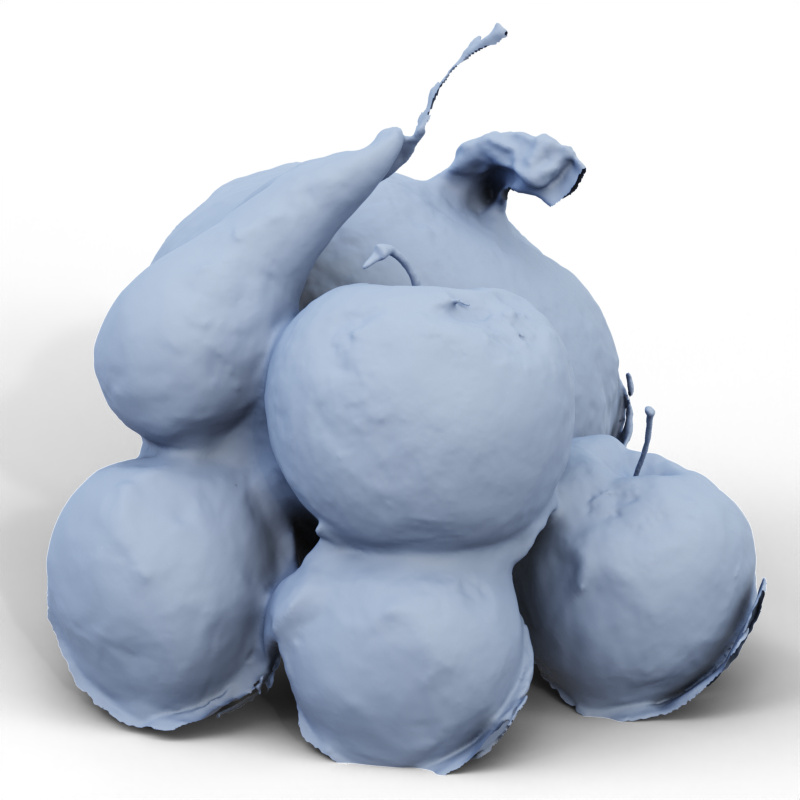}
    \includegraphics[width=0.11\linewidth]{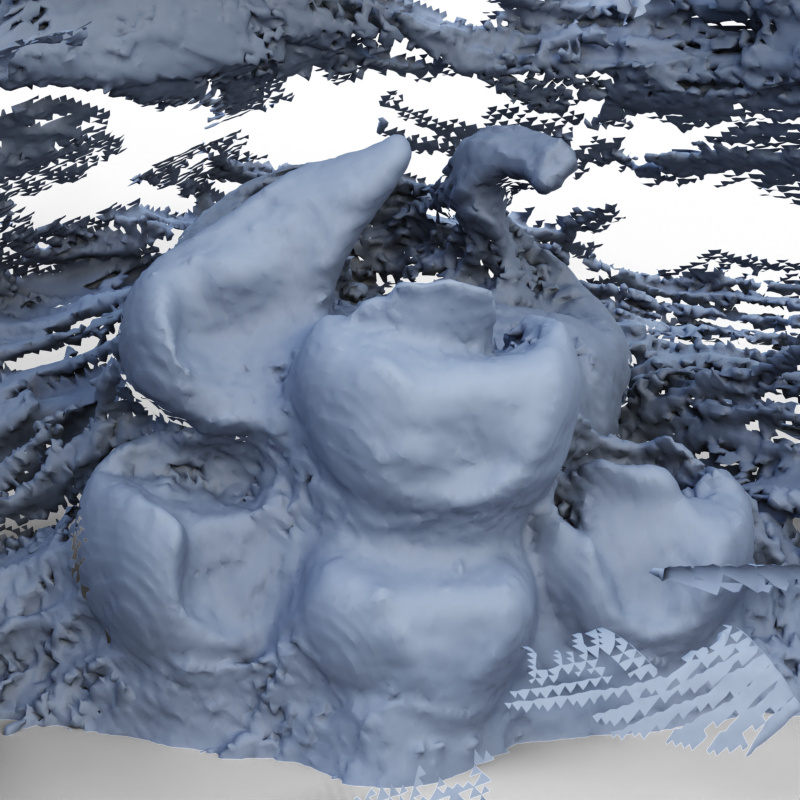}
    \includegraphics[width=0.11\linewidth]{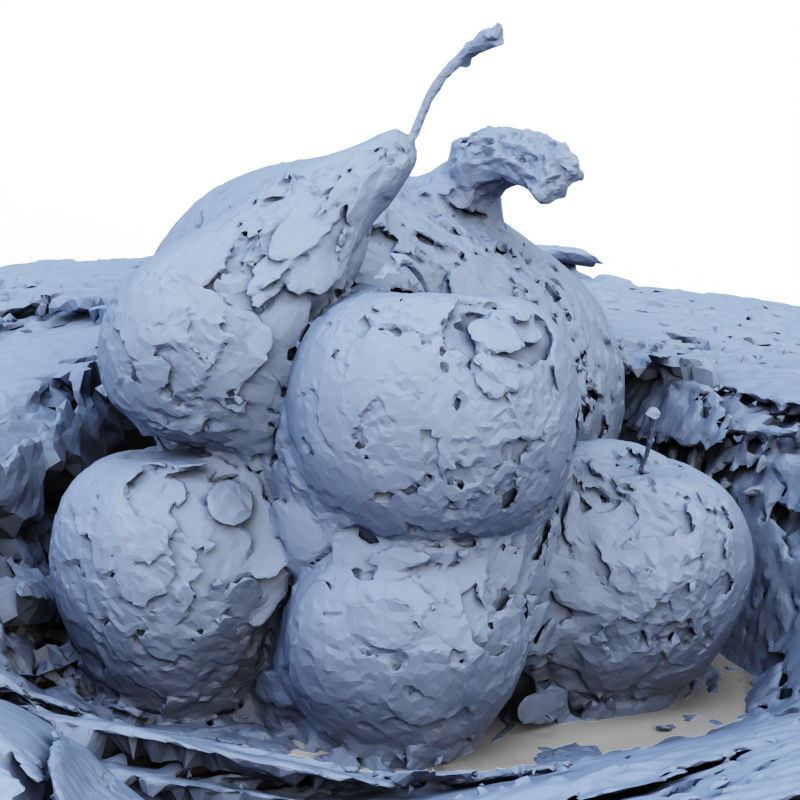}
    \includegraphics[width=0.11\linewidth]{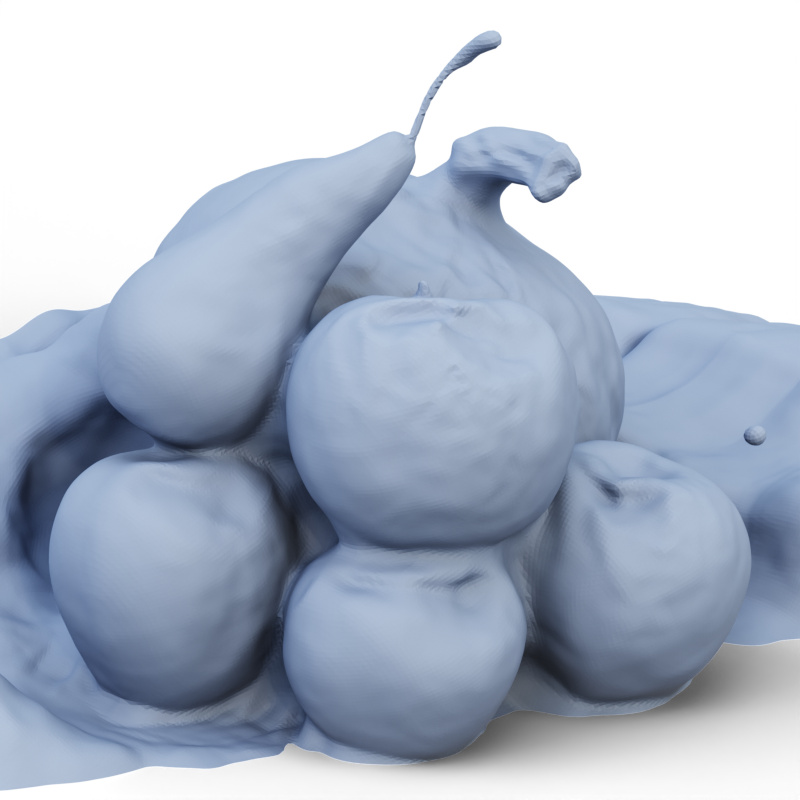}
    \\
    \includegraphics[width=0.11\linewidth]{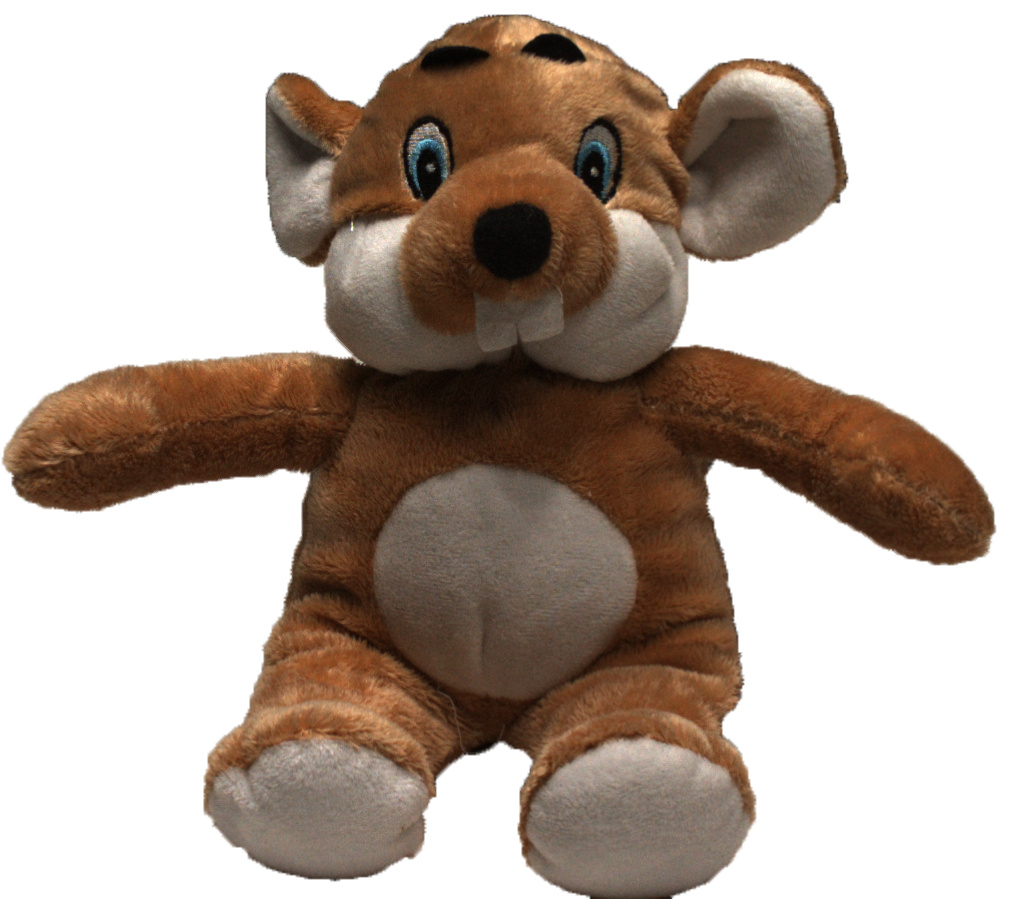}
    \includegraphics[width=0.11\linewidth]{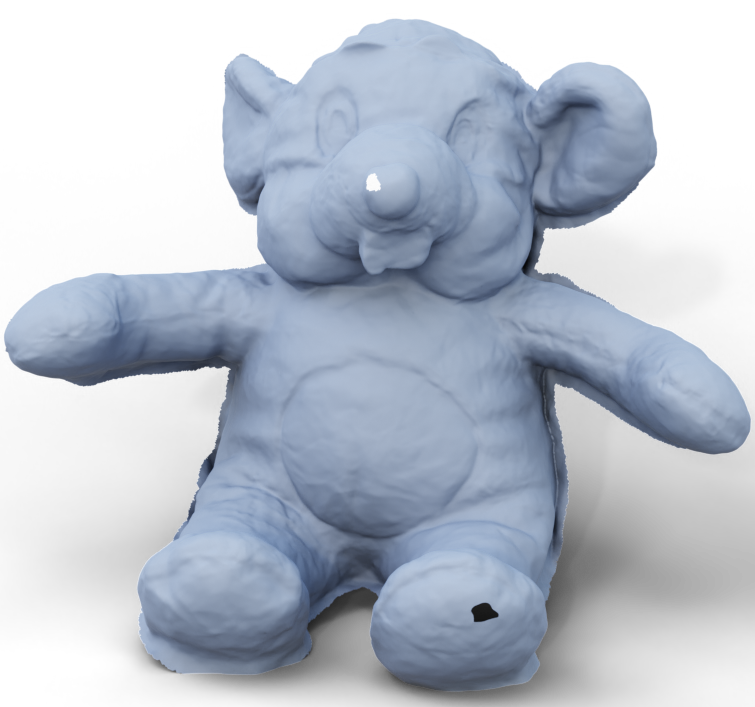}
    \includegraphics[width=0.11\linewidth]{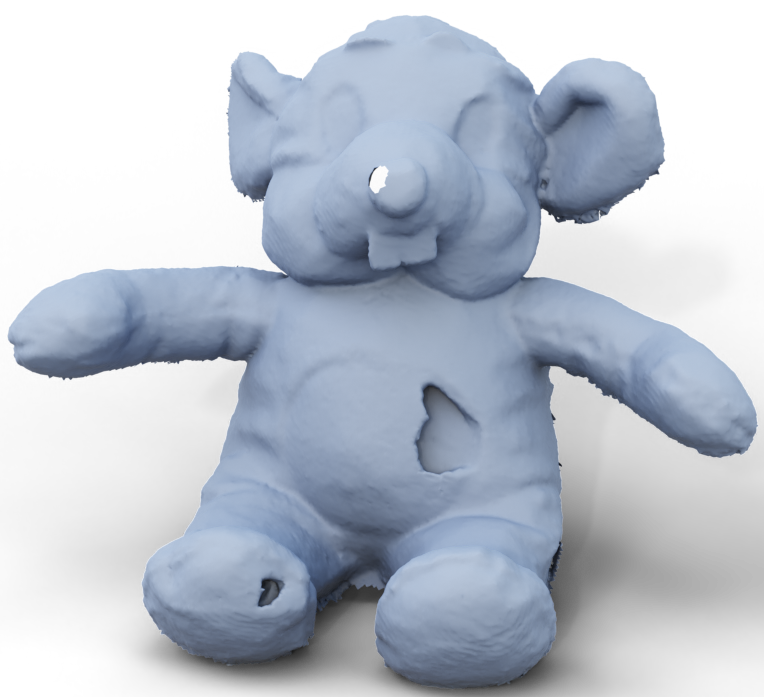}
    \includegraphics[width=0.11\linewidth]{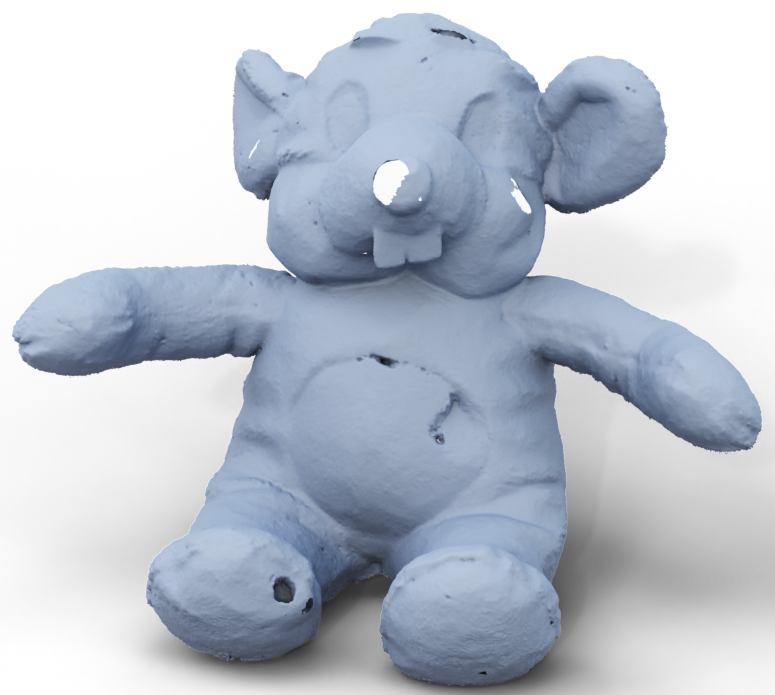}
    \includegraphics[width=0.11\linewidth]{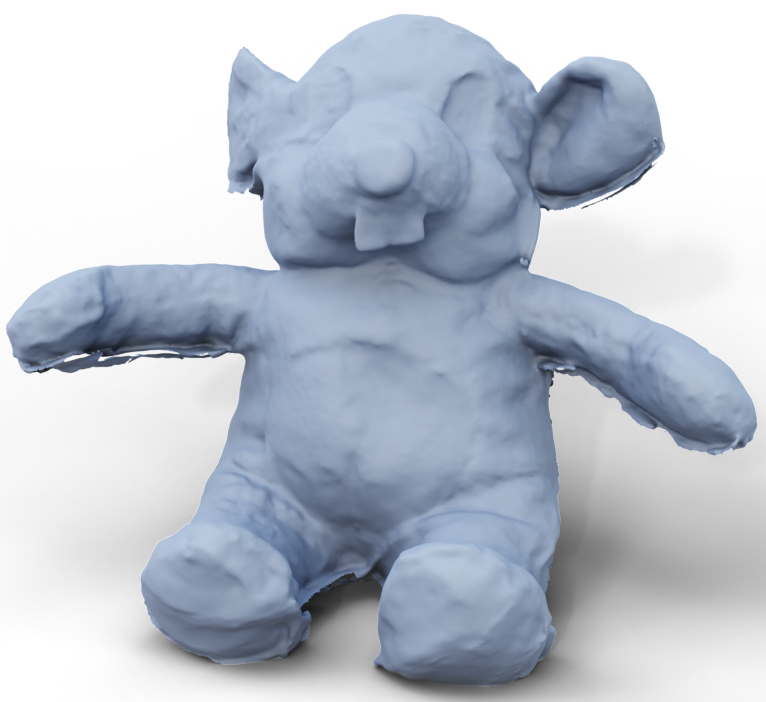}
    \includegraphics[width=0.11\linewidth]{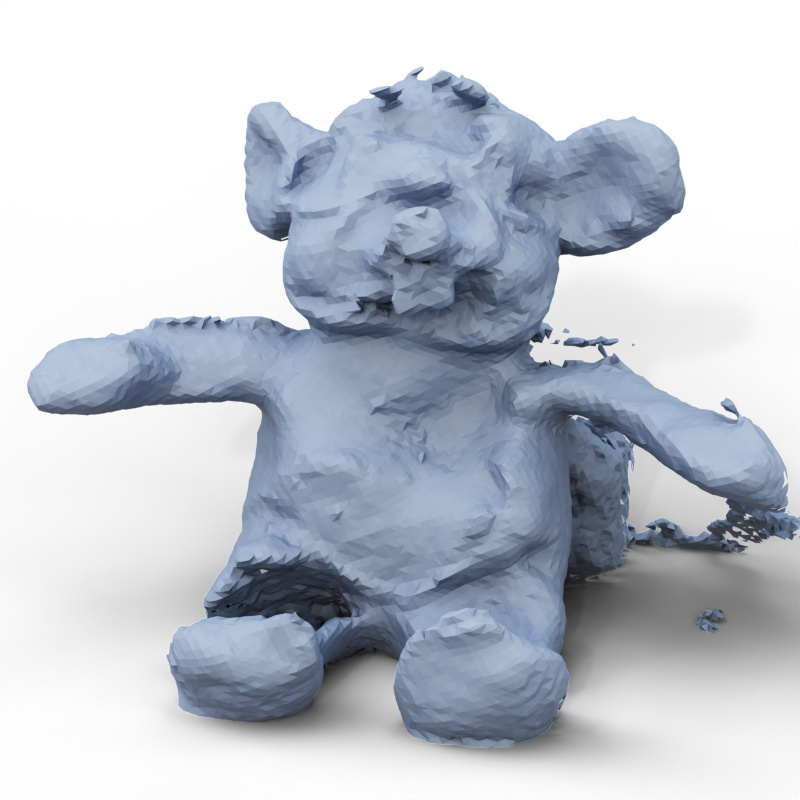}
    \includegraphics[width=0.11\linewidth]{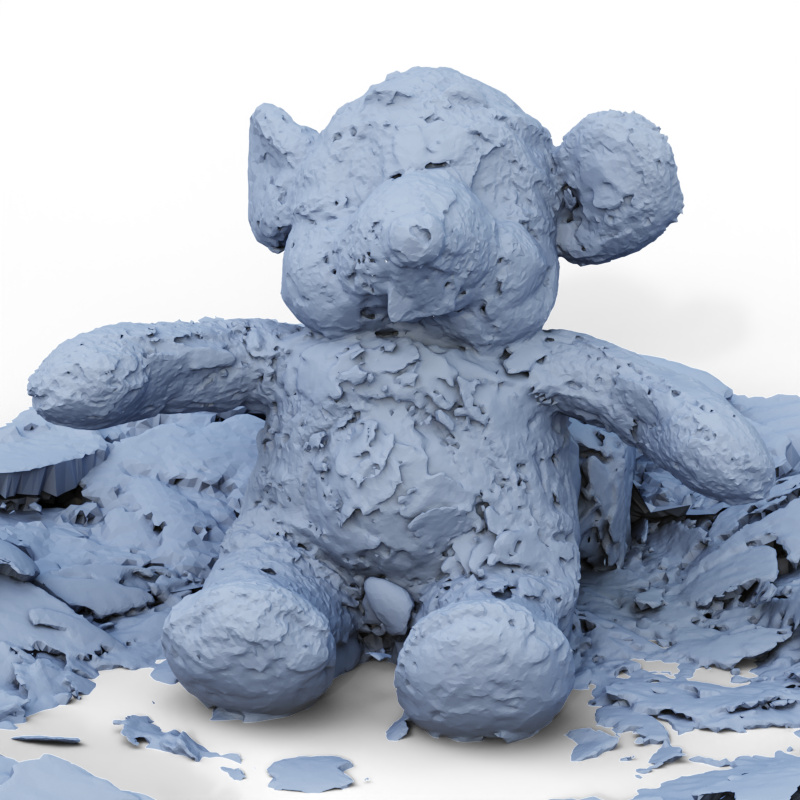}
    \includegraphics[width=0.11\linewidth]{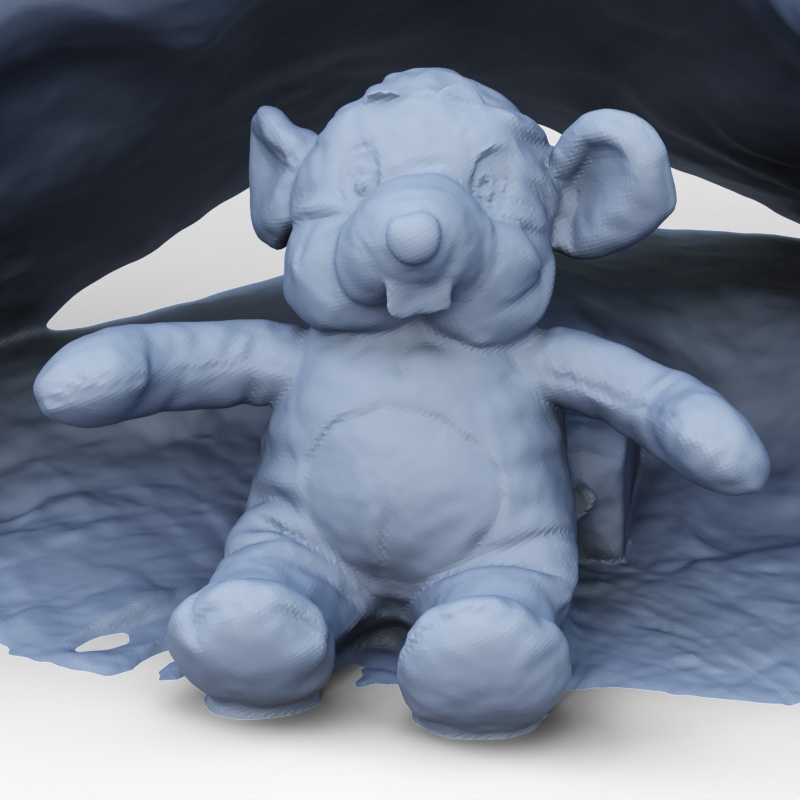}
    \\
    \includegraphics[width=0.11\linewidth]{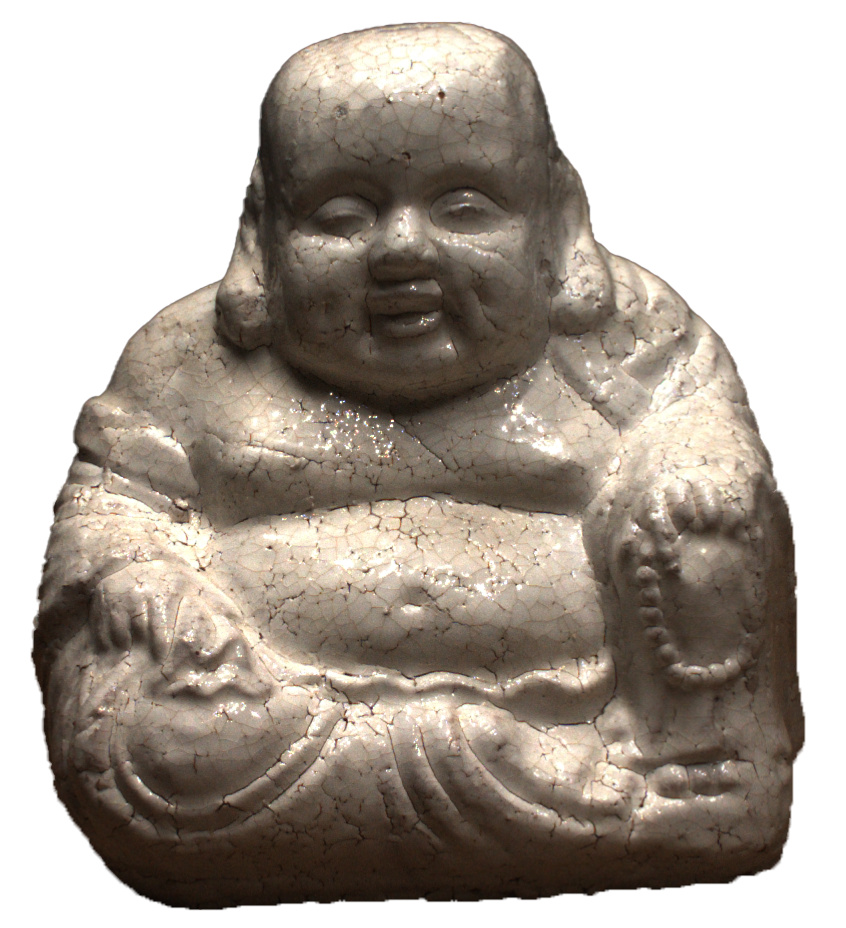}
    \includegraphics[width=0.11\linewidth]{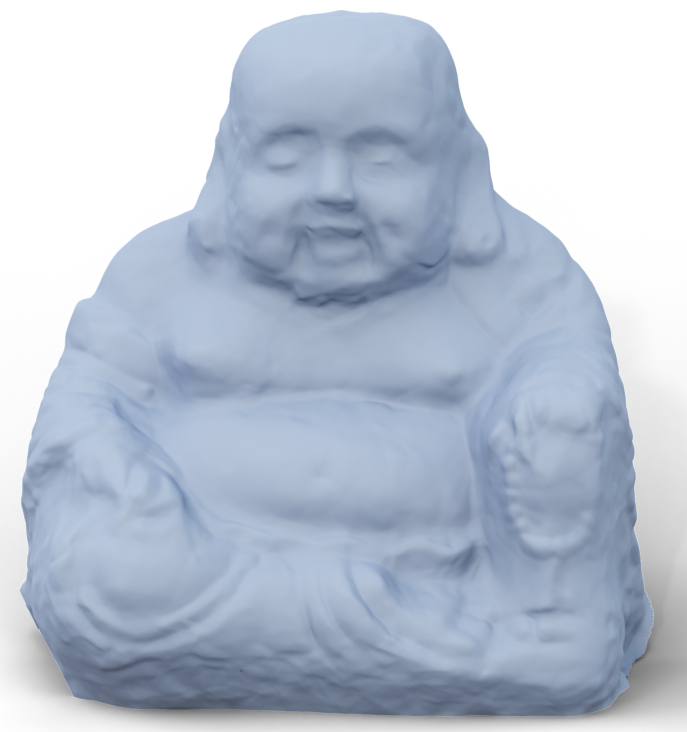}
    \includegraphics[width=0.11\linewidth]{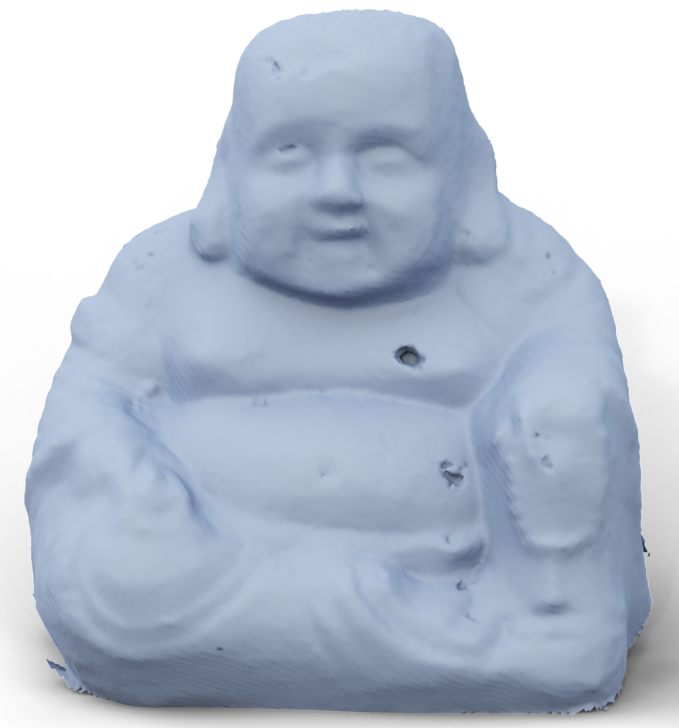}
    \includegraphics[width=0.11\linewidth]{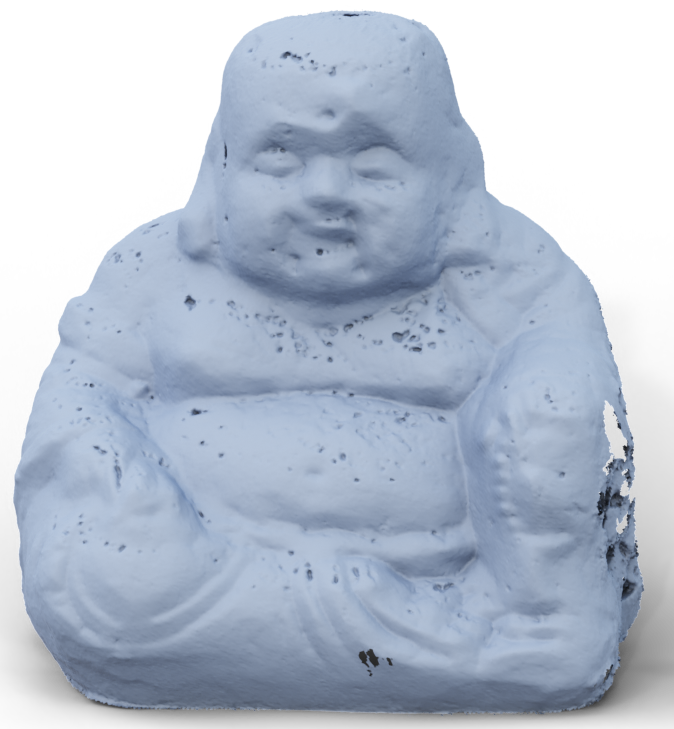}
    \includegraphics[width=0.11\linewidth]{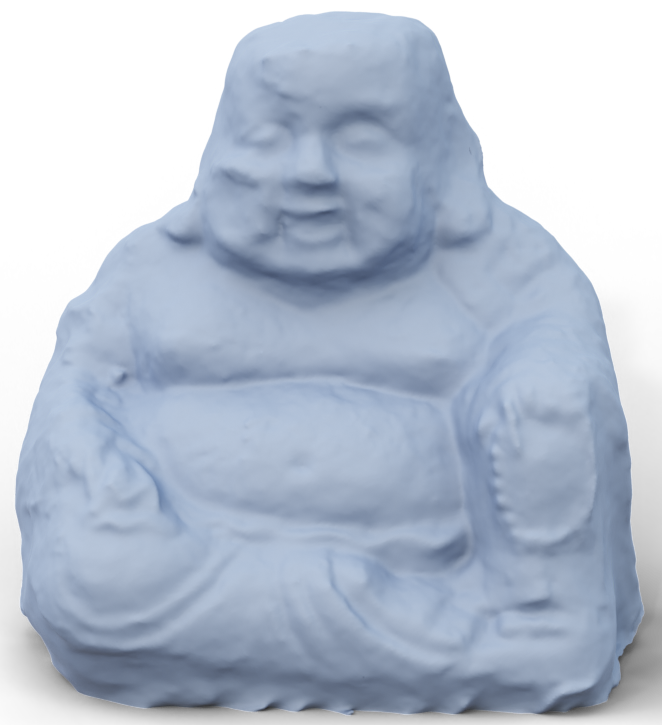}
    \includegraphics[width=0.11\linewidth]{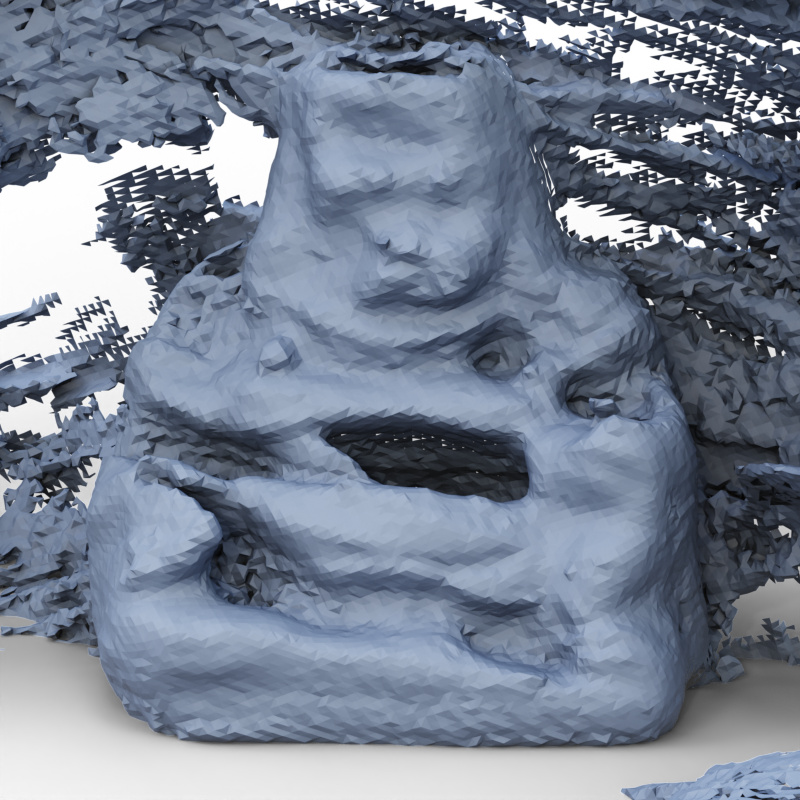}
    \includegraphics[width=0.11\linewidth]{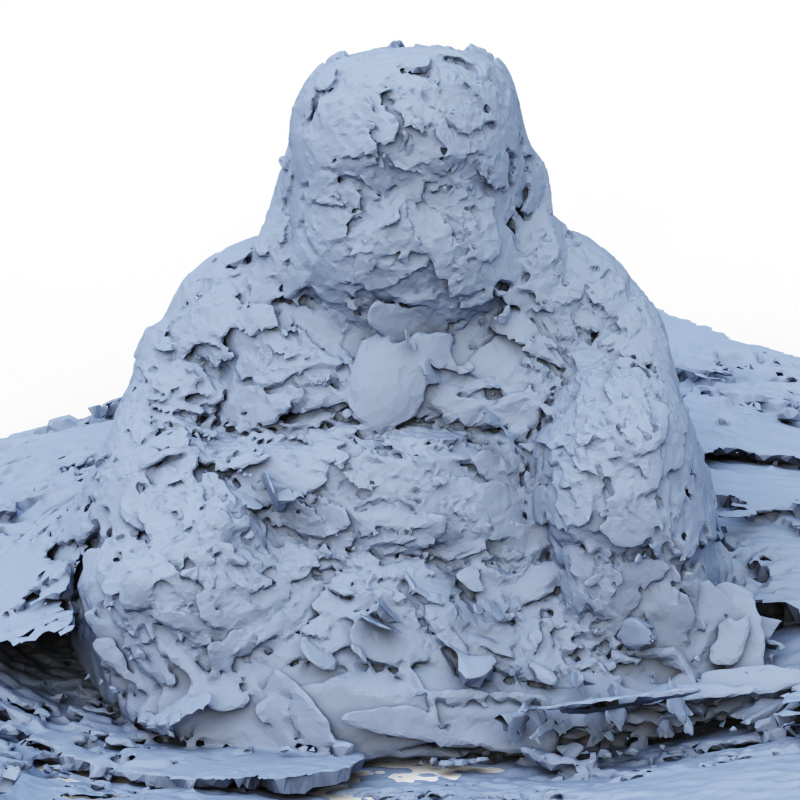}
    \includegraphics[width=0.11\linewidth]{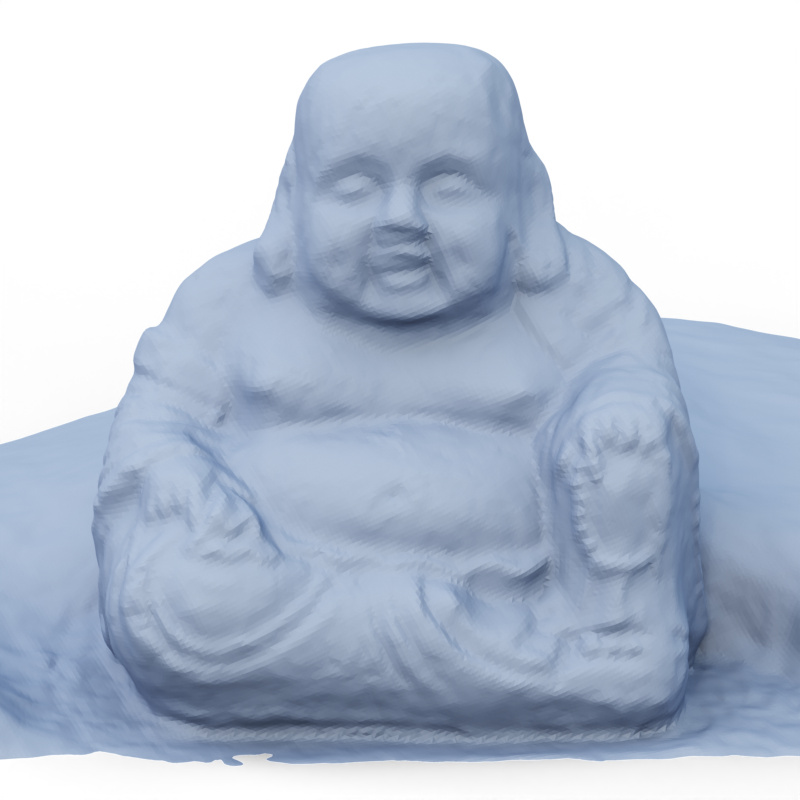}
    \\
    \scriptsize
    \makebox[0.11\linewidth]{Reference}
    \makebox[0.11\linewidth]{Ours}
    \makebox[0.11\linewidth]{2DGS}
    \makebox[0.10\linewidth]{GOF}
    \makebox[0.13\linewidth]{\tiny GaussianSurfels}
    \makebox[0.10\linewidth]{DN-Splatter}
    \makebox[0.11\linewidth]{SuGaR}
    \makebox[0.11\linewidth]{VolSDF}     
    \caption{Visualizations of cropped meshes using scripts from 2DGS~\cite{huang20242d} and GOF~\cite{yu2024gof}.}
    \label{fig:dtu}
\end{figure*}

\subsection{Gaussian Regularization}
\label{subsec:gs_reg}
As previously discussed, the color loss alone is insufficient to constrain Gaussian primitives to the object surface, since semi-transparent Gaussians located off-surface can still contribute to rendering. To ensure that all Gaussian centroids align with the object surface, which is crucial for learning an accurate SDF, we introduce a regularization term that penalizes low-opacity Gaussians. 

Given our focus on opaque objects, we encourage Gaussians to be highly opaque. Rather than fixing the opacity values, as done in prior work~\cite{peng2024rtg}, we employ an entropy-based loss to drive opacities toward binary values (0 or 1):
\begin{equation}
    L_\text{ent} = \lambda_6 \left(-\sum_k o_k \ln(o_k)\right),
\end{equation}
where \( o_k \) is the opacity of the \( k \)-th Gaussian.

During training, we further prune Gaussians whose opacities fall below a predefined threshold, effectively reducing redundancy and enhancing SDF learning stability. Opacity is parameterized using a sigmoid activation $o_k = \sigma(\tilde{o}_k) \in (0, 1)$, where $\tilde{o}_k \in \mathbb{R}$ is the unbounded learnable parameter optimized during training. The entropy loss actively drives $o_k$ toward the binary extremes. This loss is coupled with a hard pruning threshold $\tau = 0.05$. Gaussians with opacities falling below this threshold are periodically removed.

\subsection{Geometry-guided Appearance Modeling}
\label{subsec:app_model}

As noted in previous works~\cite{xu2023deformable, wu2022voxurf}, appearance modeling based solely on spatial position and viewing direction often fails to capture fine-grained geometry. To address this, incorporating geometric cues, such as normals and learned geometric features, has proven effective in neural implicit surface methods~\cite{yariv2020idr}. Following this insight, we use the SDF module $f_{\text{sdf}}$ to predict both the signed distance $s$ and a geometric feature vector $F_{\text{geo}}$ for each Gaussian centroid $\mathbf{p}_k$:
\begin{equation}
    (s, F_\text{geo}) = f_\text{sdf}(\mathbf{p}_k).
\end{equation}
The appearance module $f_\text{rgb}$ then takes as input the centroid position $\mathbf{p}_k$, viewing direction $\mathbf{d}$, the normal vector $\nabla f_\text{sdf}$, and the geometry feature $F_\text{geo}$:
\begin{equation}
    \mathbf{c}_k = f_\text{rgb}\left(\mathbf{p}_k, \mathbf{d}, \nabla f_\text{sdf}, F_\text{geo} \right),
\end{equation}
producing the RGB color $\mathbf{c}_k$ for each Gaussian.

\subsection{Implementation Details}
\label{subsec:implement}

The overall loss function in GSurf is defined as follows.
\begin{equation}
    L_\text{total} = L_\text{gs} + L_\text{sdf} + L_\text{ent}.
\end{equation}
We model the SDF and the appearance with MLPs of 8 and 4 layers, respectively, each with 256 hidden units and ReLU activation. 

The 3D Gaussians are initialized randomly for the OmniObjects3D~\cite{wu2023omniobject3d} dataset, and from sparse, noisy Structure-from-Motion (SfM)~\cite{schoenberger2016sfm} point clouds for the DTU~\cite{jensen2014large} dataset, we delay the activation of SDF learning, beginning SDF training only after the first 500 iterations. To promote robust convergence, the SDF is initialized as a sphere. Both Gaussians and MLPs are optimized using the Adam optimizer~\cite{kingma2014adam}, trained on a single NVIDIA A800 GPU. We adopt the same densification strategies as those used in 2DGS~\cite{huang20242d}. The hyperparameters are empirically set to $ \lambda_1 = 0.1$, $\lambda_2 = \lambda_3 = \lambda_6 = 0.01$, and $\lambda_4 = \lambda_5= 0.05$.

\section{Experimental results}
\subsection{Setup}
\paragraph{Datasets and metrics} We evaluated our method on the DTU dataset~\cite{jensen2014large}, a widely used benchmark for 3D reconstruction, as well as the OmniObject3D-d and OO3D-SL datasets, both of which are subsets of the OmniObject3D dataset~\cite{wu2023omniobject3d}. OmniObject3D-d focuses on object categories such as antiques and ornaments, comprising 8 objects with rich geometric details. OO3D-SL, used in~\cite{li2024real}, contains 24 objects captured under strong lighting conditions, which pose significant challenges for multi-view 3D reconstruction methods. Each object in the OmniObject3D dataset includes approximately 100 images at a resolution of 800$\times$800 pixels, along with a ground-truth mesh. Furthermore, we evaluated our method on the $\alpha$-NeuS~\cite{zhang2024alphaneus} and Ref-NeRF~\cite{verbin2022refnerf} datasets, which include objects with semi-transparent and reflective materials.

We report Chamfer distance (CD) as a measure of geometric accuracy. For models in the OmniObjects3D dataset, we also included normal consistency (NC), which is computed using the available ground-truth meshes. NC is generally more consistent with the perception of surface quality by humans. The evaluation scripts for the DTU and OmniObjects3D datasets are adapted from 2DGS~\cite{huang20242d}, GOF~\cite{yu2024gof}, and the official OmniObjects3D repository~\cite{wu2023omniobject3d}.
\begin{table*}
    \centering    %
    \caption{Quantitative evaluation on the OmniObjects3D-d dataset. We report CD ($\times 10^3 \downarrow$) and Normal Consistency (NC $\uparrow$). OR and GAM stand for opacity regularization and geometry-guided appearance modeling, respectively. 2DGS+NSH refers to the integration of Gaussian splatting with NSH~\cite{wang2023neuralsingular}, a deep learning model for point cloud reconstruction, where the generated Gaussian primitives are fed into the NSH network.
    }
    \label{tab:oo3d-d}
    \resizebox{\textwidth}{!}{\begin{tabular}{c|cc|cc|cc|cc|cc|cc|cc|cc|cc}
        \toprule
        \multirow{3}{*}{Method} & \multicolumn{10}{c|}{Ornaments} & \multicolumn{6}{c|}{Antique} & \multicolumn{2}{c}{Mean}\\
        & \multicolumn{2}{c}{1} & \multicolumn{2}{c}{3} & \multicolumn{2}{c}{4} & \multicolumn{2}{c}{5} & \multicolumn{2}{c|}{8} & \multicolumn{2}{c}{16} & \multicolumn{2}{c}{19} & \multicolumn{2}{c|}{21} & \\
        \cline{2-19}
        & CD & NC & CD & NC& CD & NC& CD & NC& CD & NC& CD & NC& CD & NC& CD & NC &CD & NC\\
        \midrule
        VolSDF~\cite{yariv2021volume}& 12.13 & 0.950 & 4.32 & 0.992 & 14.00&0.939 &9.11&0.959 & 22.89&0.903 & 29.45 & 0.938 & 11.60 & 0.906 & 12.56 & 0.932 & 14.50 & \underline{0.939}\\
        2DGS~\cite{huang20242d} & 8.27 &0.954 & 19.63&0.924 & 8.74&0.964 & 5.82&0.962 & 32.74&0.813 & 10.86&0.939 & 11.68&0.853 & 12.11&0.940 & 13.73 & 0.918\\
        GOF~\cite{yu2024gof} & 7.90&0.956 & 4.66&0.992 & 12.16&0.961 & 5.97&0.966 & 44.08&0.790 & 23.71&0.898 & 15.44&0.834 &21.76&0.966 & 16.95 & 0.920\\
        GSurfels~\cite{dai2024gsurfel}& 11.13&0.950 & 7.57&0.980 & 9.55&0.970 & 8.05&0.962& 13.95&0.886 & 11.84&0.943 & 9.28&0.918 & 9.36&0.962 & \underline{10.09} & \textbf{0.946} \\
        2DGS+NSH~\cite{wang2023neuralsingular} & 11.26&0.922 & 12.85&0.936 & 16.12&0.905 & 9.68&0.942 & 11.27&0.893 & 23.13&0.901 & 15.37&0.860 & 33.18&0.873 & 16.60 & 0.904 \\
        Ours & 7.04&0.959 & 3.11&0.978 & 8.34&0.967 & 6.35&0.960 & 18.04&0.856 & 14.99&0.922 & 8.46&0.882 & 5.47&0.956 & \textbf{8.97} & \underline{0.934} \\
        \midrule
        Ours (\small w/o GAM)& 7.83&0.938 & 6.73&0.964 & 9.43&0.963 & 7.46&0.952 & 20.56&0.812 & 17.88&0.918 & 10.43&0.886 & 5.97& 0.962 & 10.78 & 0.924\\
        Ours (w/o OR)& 12.50&0.917 & 25.20&0.920 & 27.18&0.869 & 21.92&0.888 & 27.19&0.831 & 28.13&0.845 & 18.83&0.828 & 13.72&0.924 & 21.83 & 0.877\\
        \bottomrule
    \end{tabular}}
\end{table*}
\paragraph{Baselines} We compared our method with the latest 3DGS-based reconstruction approaches, including 2DGS~\cite{huang20242d}, GOF~\cite{yu2024gof}, SuGaR~\cite{guedon2024sugar}, DN-Splatter~\cite{turkulainen2024dnsplatter}, and GaussianSurfels~\cite{dai2024gsurfel}. For GaussianSurfels~\cite{dai2024gsurfel}, we observed that its pre-trained normal prediction model often generated inaccurate normals, occasionally with completely flipped orientations, when applied to models from the OmniObjects3D dataset. Therefore, we disabled the pre-trained normal estimation during our evaluations on this dataset.

\begin{figure}
    \centering
    \includegraphics[width=0.975\linewidth]{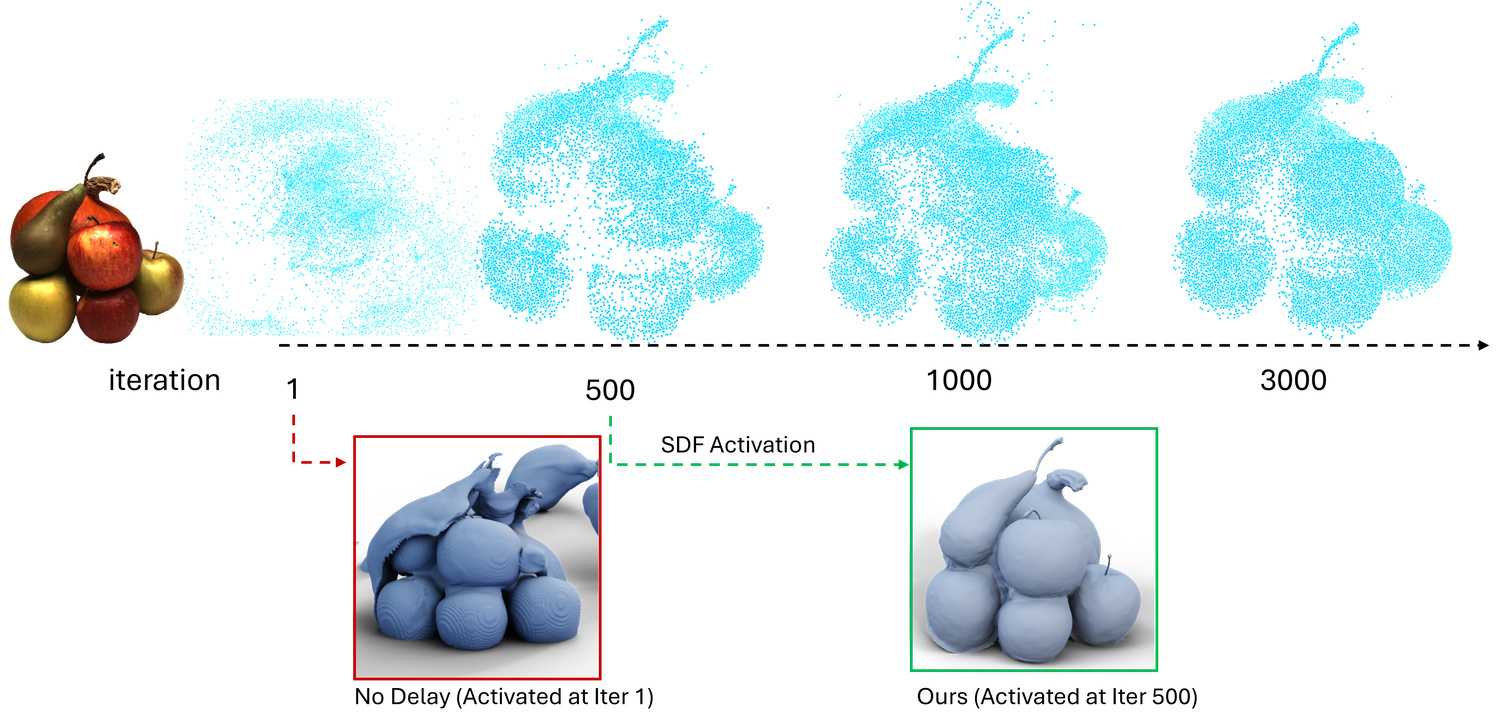}
    \caption{ Impact of the delayed SDF activation strategy. Our 500-iteration delay provides the SDF with high-confidence spatial anchors, avoiding early optimization collapse while maintaining training efficiency.}
    \label{fig:init_gs}
    \vspace{-0.2in}
\end{figure}

\subsection{Comparisons}
\paragraph{Training time} Our method achieves competitive training efficiency, requiring only 40 minutes to 1.3 hours on a single NVIDIA A800 GPU. Prior methods that integrate Gaussian splatting with SDFs typically require 2-16 hours of training, for example, 2 hours for GSDF~\cite{yu2024gsdf} and up to 16 hours for NeuSG~\cite{chen2023neusg}.

\paragraph{Mitigating Cold-Start Divergence.} At initialization (Iteration 1, \Cref{fig:init_gs}), Gaussian primitives derived from initial point clouds exhibit a highly chaotic and noisy spatial distribution. Supervising the continuous implicit field with these unaligned, raw centroids would inherently pull the zero-level set away from the true surface, leading to severe geometric distortion and optimization collapse.

To circumvent this cold-start problem, we propose a 500-iteration delayed SDF activation strategy. During this initial warm-up phase, the multi-view photometric loss, tightly coupled with our entropy-based opacity regularization, operates independently to rapidly prune severe outliers. As demonstrated in the primitive evolution visualization (\Cref{fig:init_gs}), this provides the implicit network with high-confidence spatial anchors from the very first step of geometric learning, effectively averting early optimization divergence. While extending the delay to 1,000 or 3,000 iterations yields marginally denser point clouds, our empirical evaluations indicate that 500 iterations provide the optimal trade-off between securing geometric stability and maintaining overall training efficiency.

\begin{figure*}[!htbp]
    \centering
    \begin{minipage}{0.65\linewidth}
        \centering
        {\footnotesize \textbf{Antique 016}}\\[0.1cm]
        \includegraphics[width=0.17\linewidth]{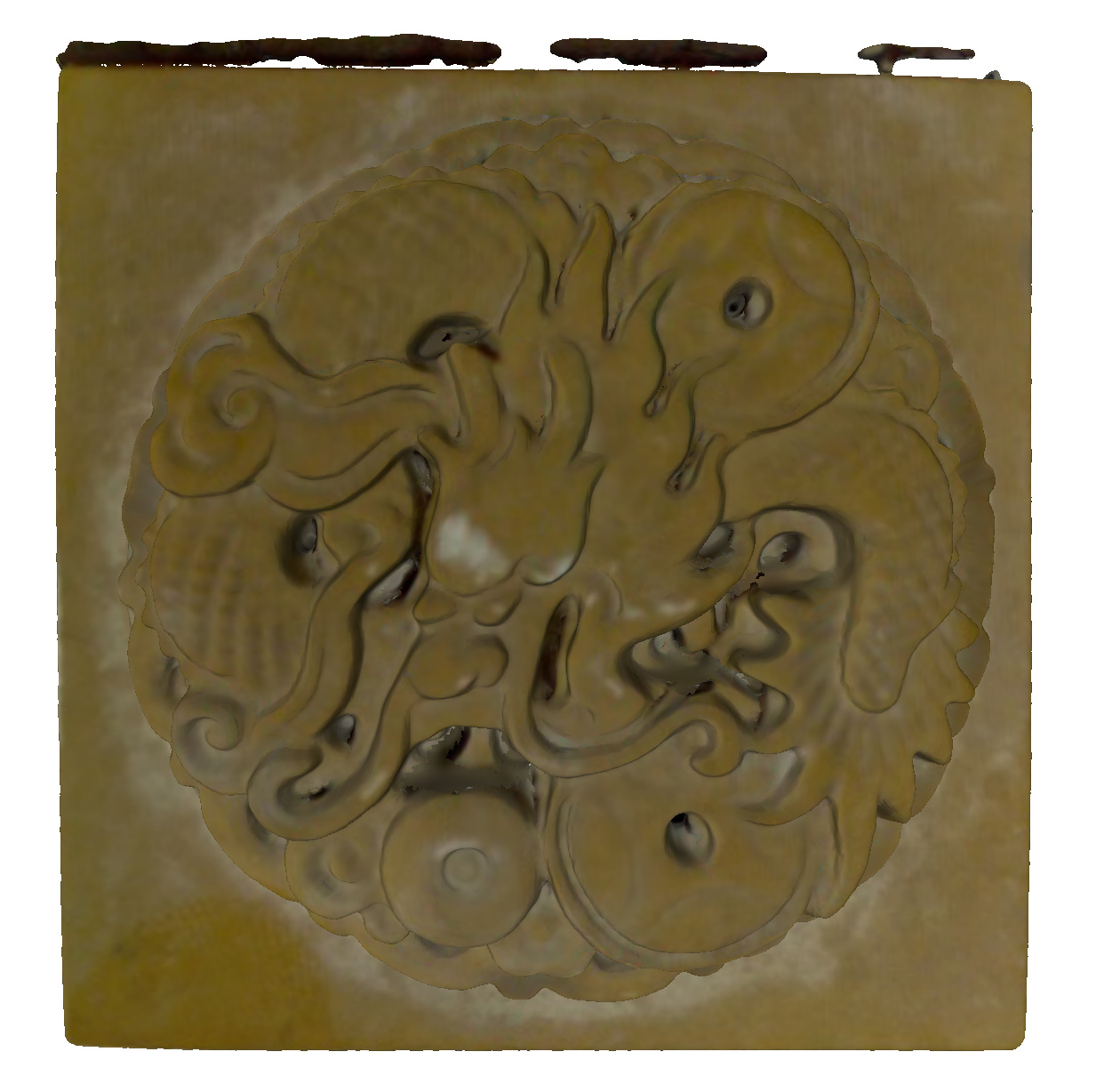}
        \includegraphics[width=0.17\linewidth]{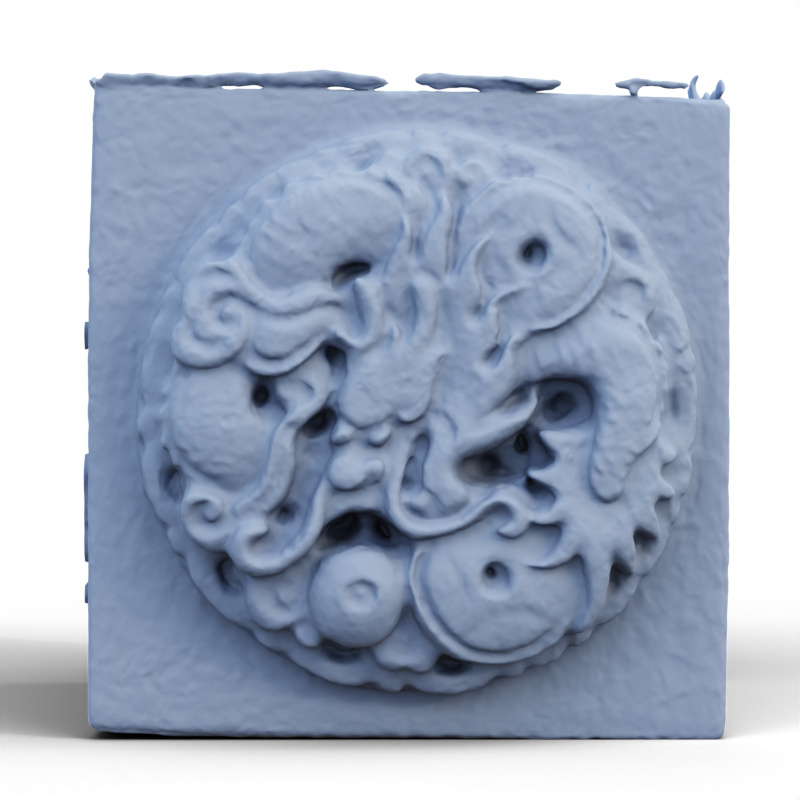}
        \includegraphics[width=0.17\linewidth]{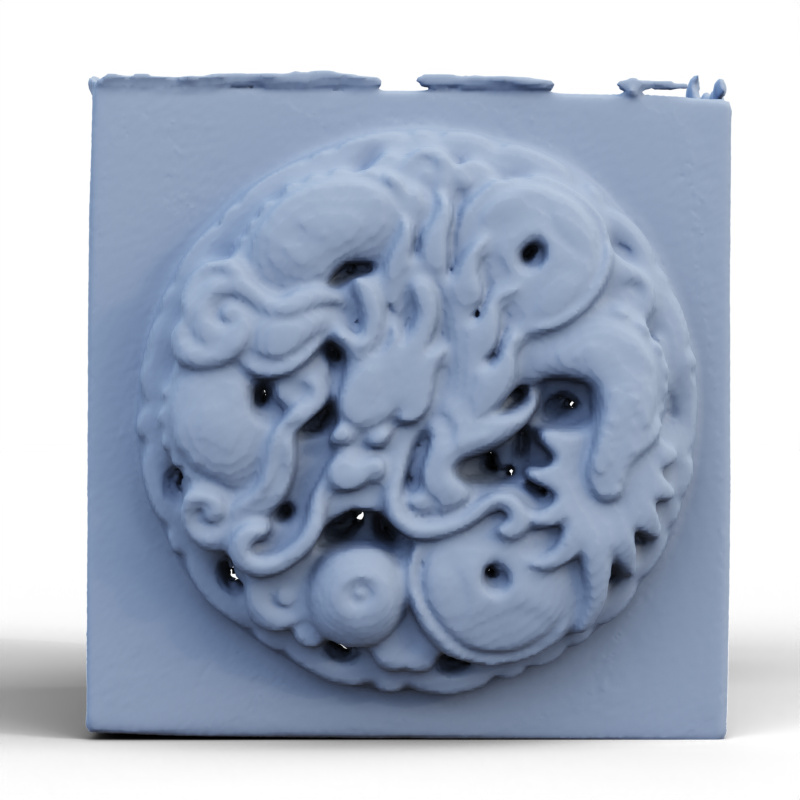}
        \includegraphics[width=0.17\linewidth]{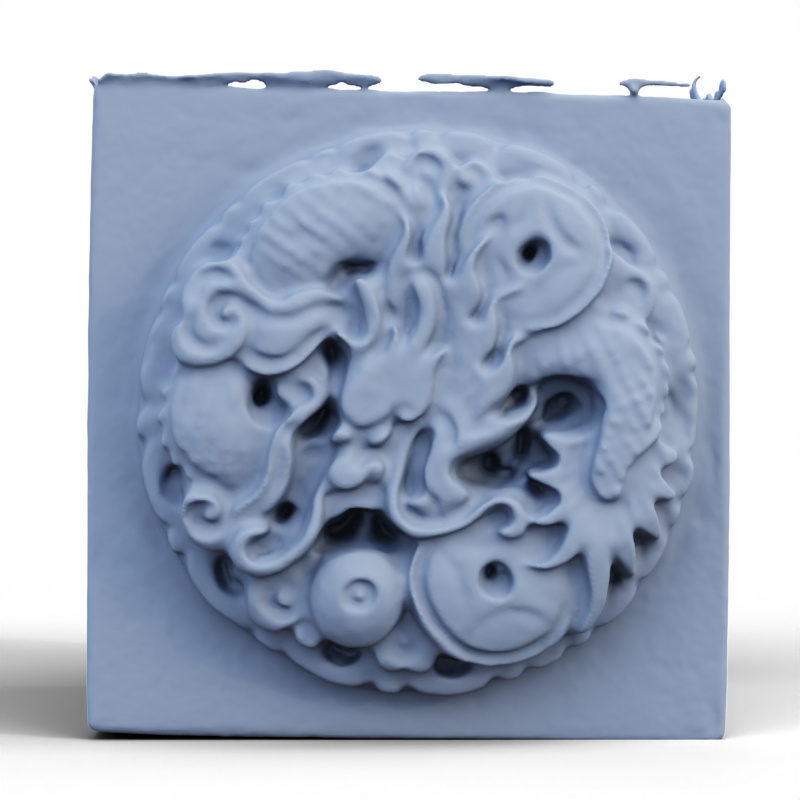}
        \includegraphics[width=0.17\linewidth]{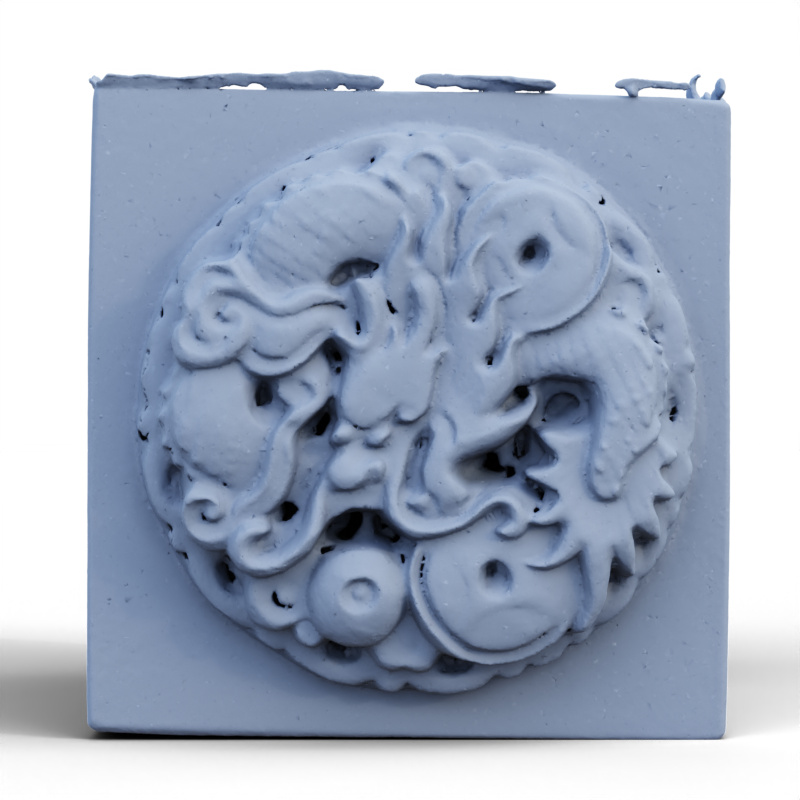}\\
        {\tiny
        \makebox[0.17\linewidth]{}
        \makebox[0.17\linewidth]{44,827}
        \makebox[0.17\linewidth]{64,711}
        \makebox[0.17\linewidth]{123,826}
        \makebox[0.17\linewidth]{54,160}\\[0.3cm]
        }
        
        {\footnotesize \textbf{Table 019}}\\[0.1cm]
        \includegraphics[width=0.17\linewidth]{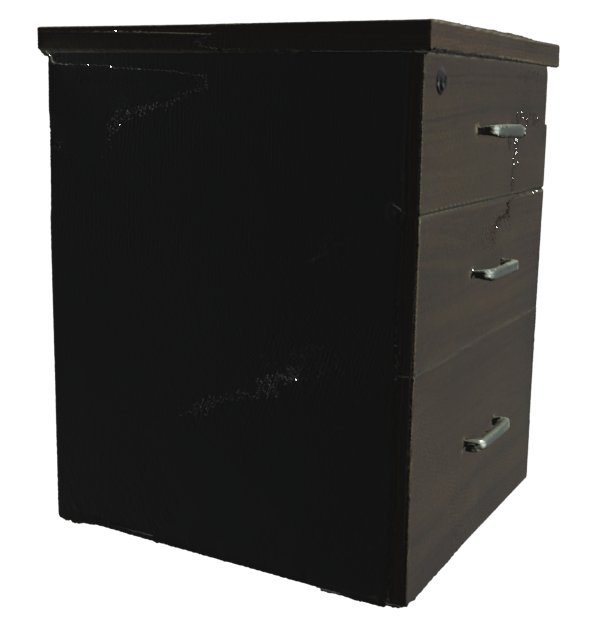}
        \includegraphics[width=0.17\linewidth]{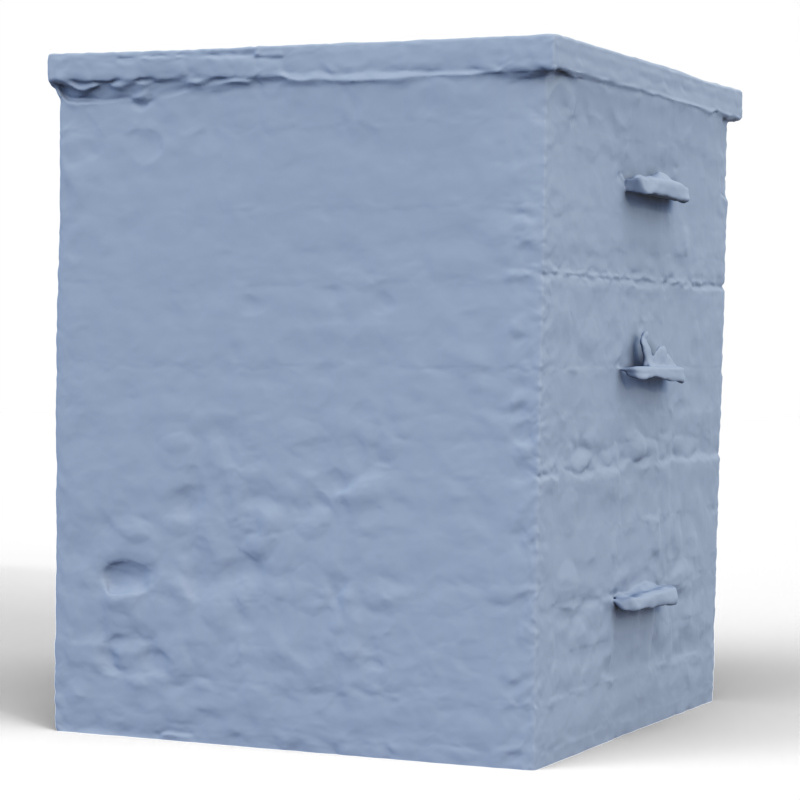}
        \includegraphics[width=0.17\linewidth]{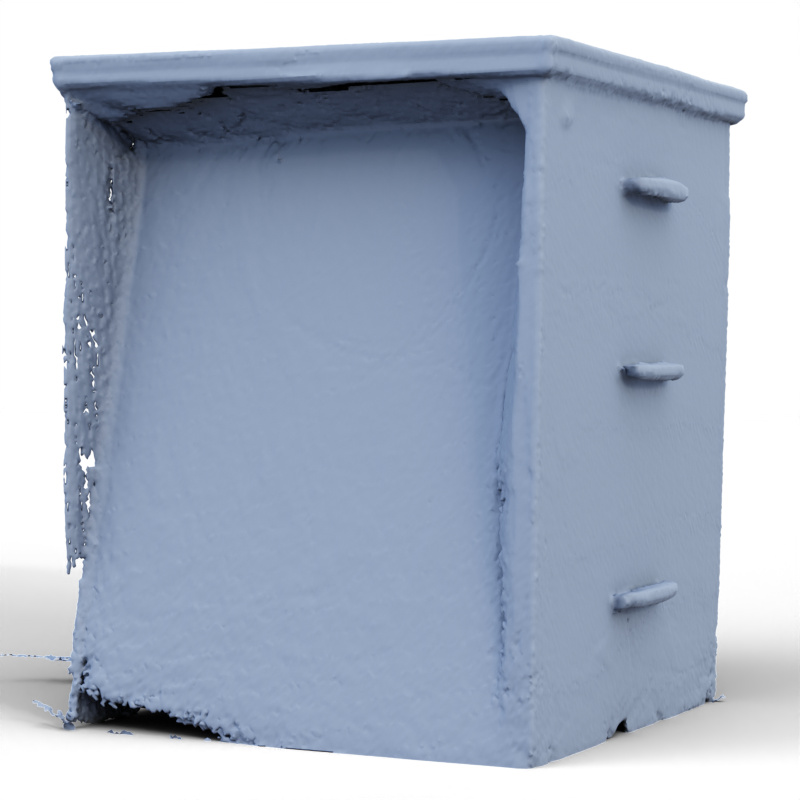}
        \includegraphics[width=0.17\linewidth]{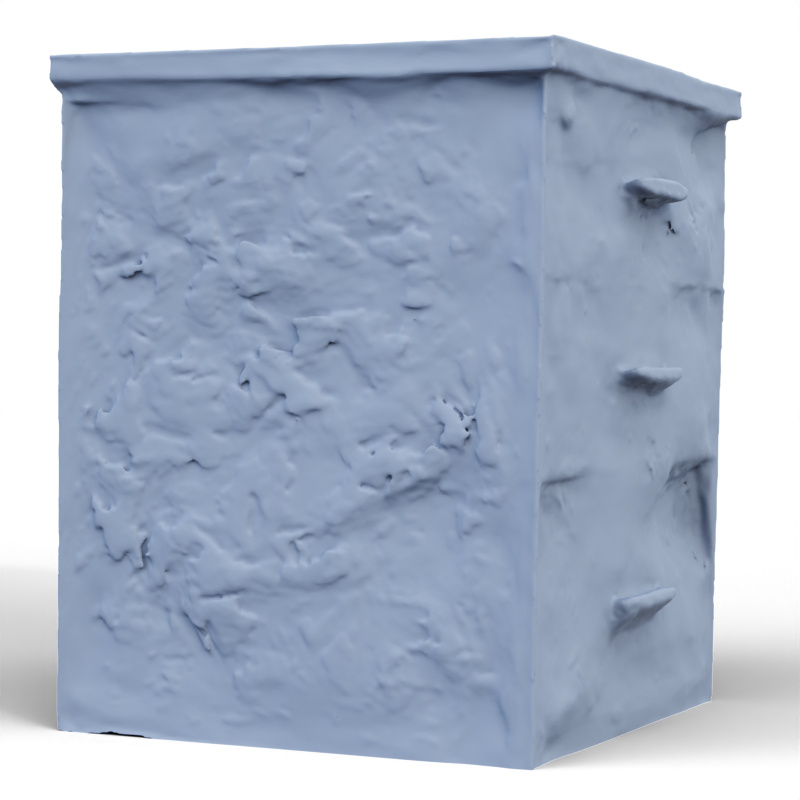}
        \includegraphics[width=0.17\linewidth]{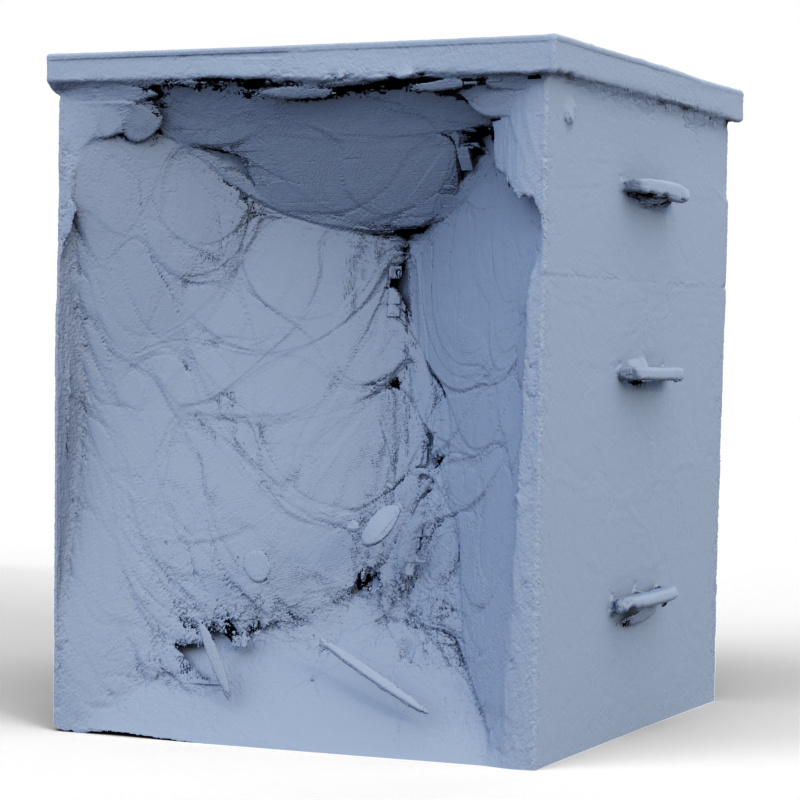}\\
        {\tiny
        \makebox[0.17\linewidth]{}
        \makebox[0.17\linewidth]{39,191}
        \makebox[0.17\linewidth]{26,098}
        \makebox[0.17\linewidth]{43,063}
        \makebox[0.17\linewidth]{32,471}\\
        \makebox[0.17\linewidth]{Reference}
        \makebox[0.17\linewidth]{Ours}
        \makebox[0.17\linewidth]{2DGS}
        \makebox[0.17\linewidth]{GaussianSurfels}
        \makebox[0.17\linewidth]{GOF}\\
        }
    \end{minipage}
    \hfill
    \begin{minipage}{0.33\linewidth}
        \centering
        \includegraphics[width=\linewidth]{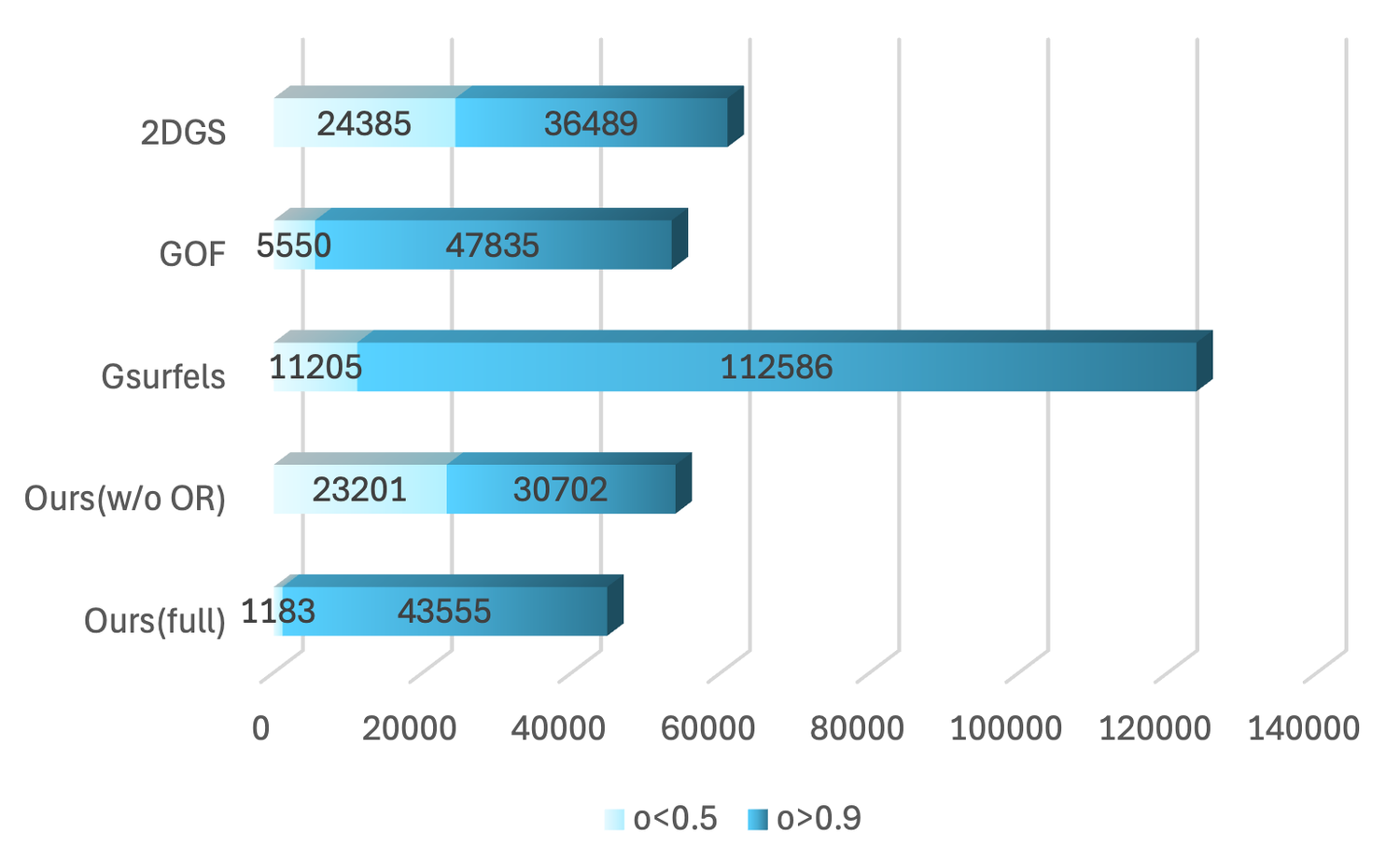}
        {\footnotesize \textbf{Opacity Distribution}}
    \end{minipage}
    
    \caption{Comparison on the Antique 016 and Table 019 models from the OmniObjects3D dataset. The number shown below each image indicates the number of Gaussians used by each method. Bar charts illustrate the distribution of low- and high-opacity Gaussians on the Antique 016 model. Our entropy-based opacity regularization significantly reduces the low-opacity Gaussians while maintaining comparable or better reconstruction quality.}
    \label{fig:oo3dd}
\end{figure*}
\paragraph{Results on DTU and OmniObjects3D-d} As shown in~\Cref{fig:dtu}, our method produces more robust geometry reconstruction than GS-based baselines. On Scan 63 (first row), it produces smoother surfaces and preserves fine details such as the apple stem, while other methods struggle with noise in reflective regions. We note that Chamfer distance in~\Cref{tab:dtu} should not be the sole evaluation metric on DTU due to (1) noise and holes in the ground-truth point clouds and (2) potential bias from cropping, which retains boundary regions contributing to higher CD. Accurate thresholding for cropping is difficult to define manually. 

VolSDF~\cite{yariv2021volume} requires over 10 hours of training per model on OmniObjects3D due to the high cost of volume rendering. While 2DGS and GOF offer faster training, they often produce fragmented surfaces and yield higher CD as reported in~\Cref{tab:oo3d-d}. GaussianSurfels~\cite{dai2024gsurfel} better handles holes, common in depth fusion pipelines~\cite{huang20242d}, but at the cost of using significantly more Gaussian primitives than our method (see~\Cref{fig:oo3dd}).
\begin{figure*}[!htbp]
    \centering
    \includegraphics[width=0.95\linewidth]{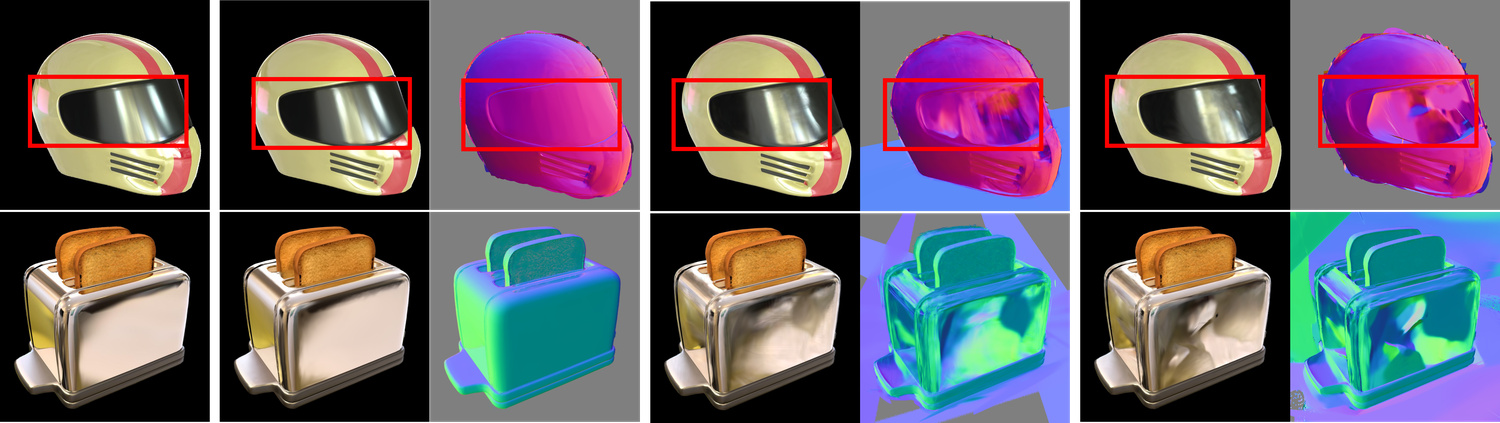}\\
    \makebox[0.12\linewidth]{\scriptsize Reference}
    \makebox[0.24\linewidth]{\small Ours}
    \makebox[0.24\linewidth]{\small GOF}
    \makebox[0.24\linewidth]{\small 2DGS}\\
    \includegraphics[width=0.11\linewidth]{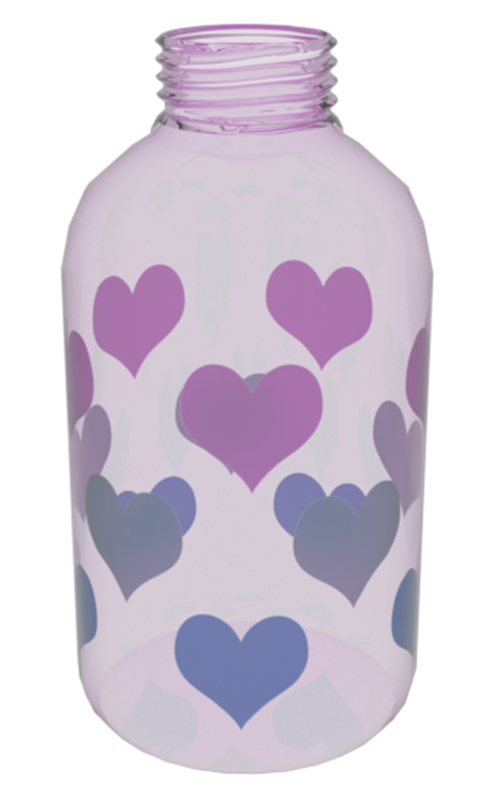}
    \includegraphics[width=0.11\linewidth]{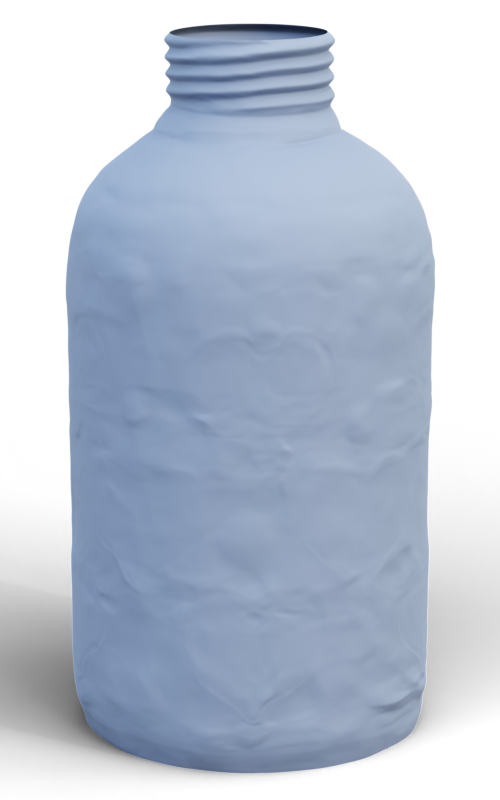}
    \includegraphics[width=0.11\linewidth]{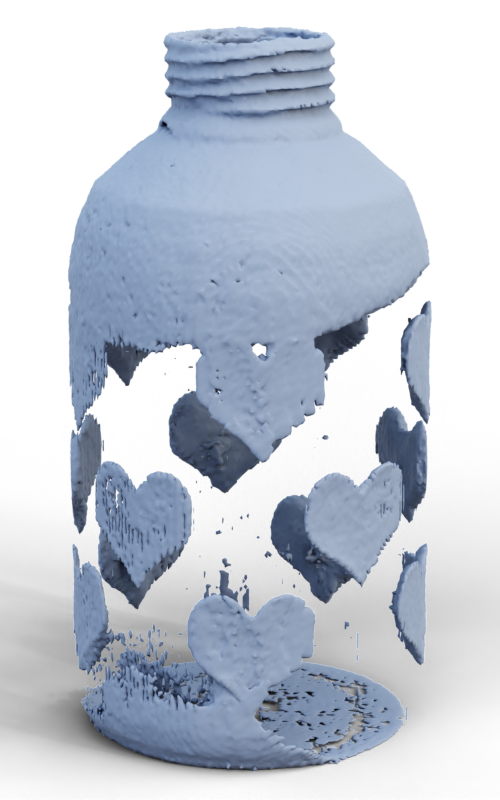}
    \includegraphics[width=0.11\linewidth]{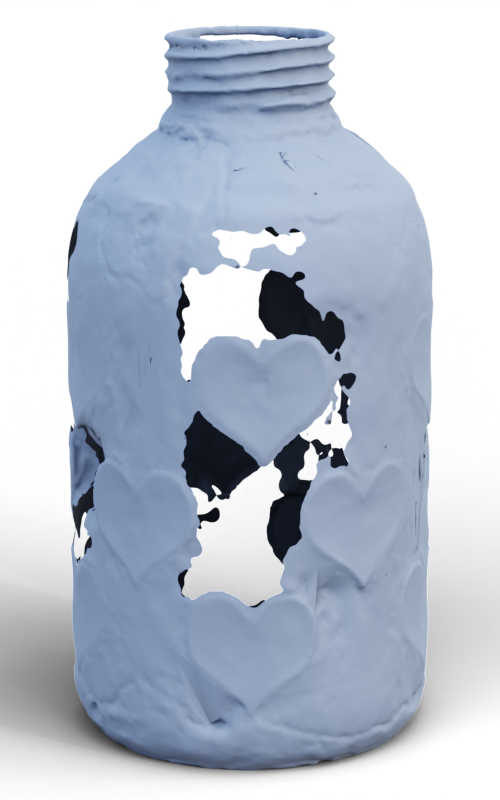}
    \includegraphics[width=0.11\linewidth]{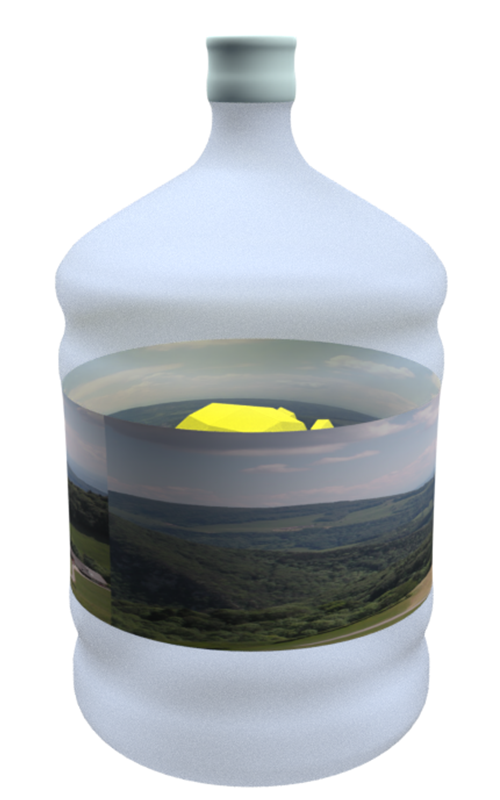}
    \includegraphics[width=0.11\linewidth]{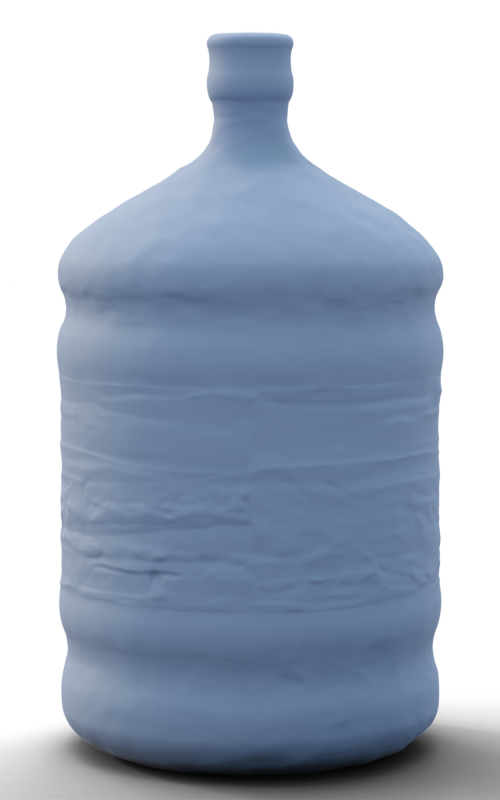}
    \includegraphics[width=0.11\linewidth]{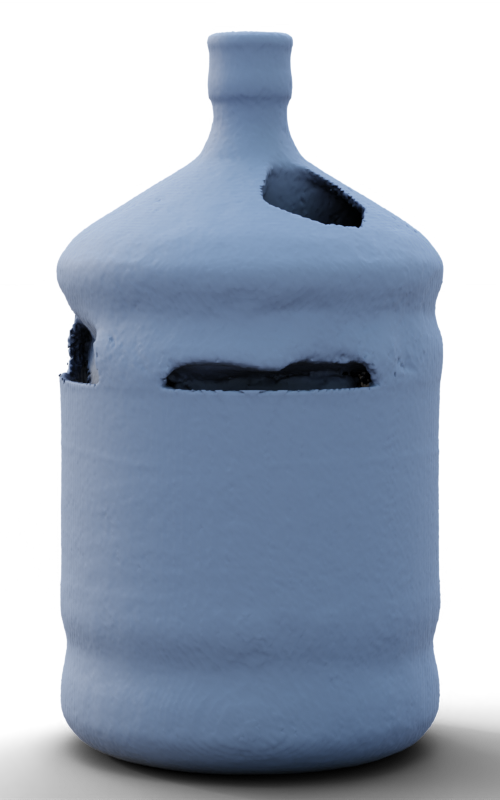}
    \includegraphics[width=0.11\linewidth]{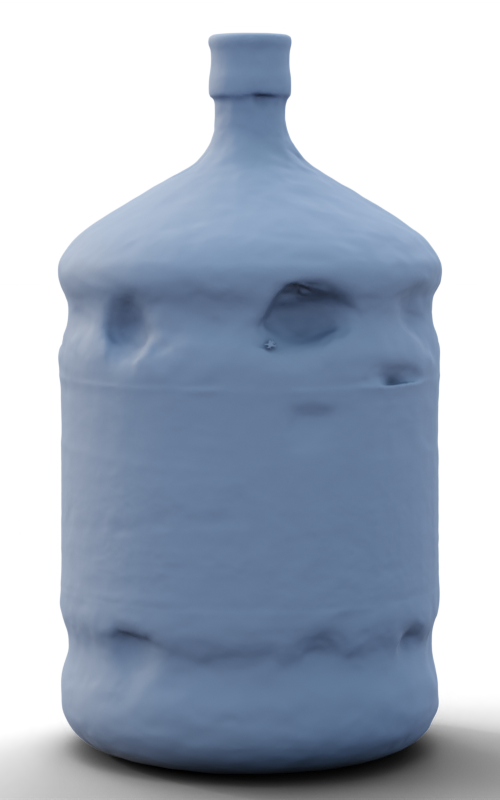}\\
    \makebox[0.11\linewidth]{\scriptsize Ref.}
    \makebox[0.11\linewidth]{\small Ours}
    \makebox[0.11\linewidth]{\small 2DGS}
    \makebox[0.12\linewidth]{\scriptsize G-Surfels}
    \makebox[0.11\linewidth]{\scriptsize Ref.}
    \makebox[0.11\linewidth]{\small Ours}
    \makebox[0.11\linewidth]{\small 2DGS}
    \makebox[0.11\linewidth]{\scriptsize G-Surfels}

  \caption{Visual comparison on the $\alpha$-NeuS dataset~\cite{zhang2024alphaneus} for semi-transparent objects and RefNeRF~\cite{verbin2022refnerf} for reflective objects.}
    \label{fig:trans}
\end{figure*}

\paragraph{Opacity distribution}
\Cref{fig:oo3dd} (bottom) shows the opacity distribution across Gaussians. In 2DGS, 37.6\% of Gaussians have opacity below 0.5. Our method produces only 2.6\% transparent Gaussians and achieves the highest proportion of opaque Gaussians, with 97.1\% having opacity greater than 0.9, compared to 56.3\% for 2DGS, 88.3\% for GOF, and 90.9\% for GaussianSurfels. This supports our design, which relies on reliable SDF learning from well-localized, opaque Gaussian centroids. GOF filters Gaussians based on opacity contribution, while GaussianSurfels applies a bell-shaped exponential regularization. In contrast, our entropy-based approach yields better compactness and higher reconstruction quality.

\paragraph{Results on OO3D-SL, $\alpha$-NeuS, and RefNeRF}
These datasets feature models with strong lighting conditions, as well as semi-transparent and reflective materials. As shown in \Cref{fig:teaser} and \Cref{fig:oo3dd}, lighting variation has minimal impact on our method due to the stability provided by integrating SDF with Gaussian splatting. In \Cref{fig:trans}, GSurf reconstructs semi-transparent and reflective surfaces more smoothly than other methods. TSDF fusion~\cite{huang20242d} and Poisson reconstruction~\cite{dai2024gsurfel} rely on potentially unreliable depth maps and often fail in such scenarios. Our method avoids this by supervising the continuous SDF directly via the centroids of Gaussians. Following UniSDF~\cite{wang2024unisdf}, we also use reflective viewpoints to model the view-dependent appearance of reflective materials. In reflective scenes such as the toaster exterior, our approach produces more realistic rendering compared to 2DGS and GOF, which struggle with rendering artifacts.
\begin{figure*}[!htbp]
    \centering
    \makebox[0.145\linewidth]{\small Reference} \hfill
    \makebox[0.145\linewidth]{\small GS-Pull} \hfill
    \makebox[0.145\linewidth]{\small Ours} \hfill
    \makebox[0.165\linewidth]{\small Reference} \hfill
    \makebox[0.165\linewidth]{\small GS-Pull} \hfill
    \makebox[0.165\linewidth]{\small Ours} \\[0.8mm]
    
    \includegraphics[width=0.145\linewidth]{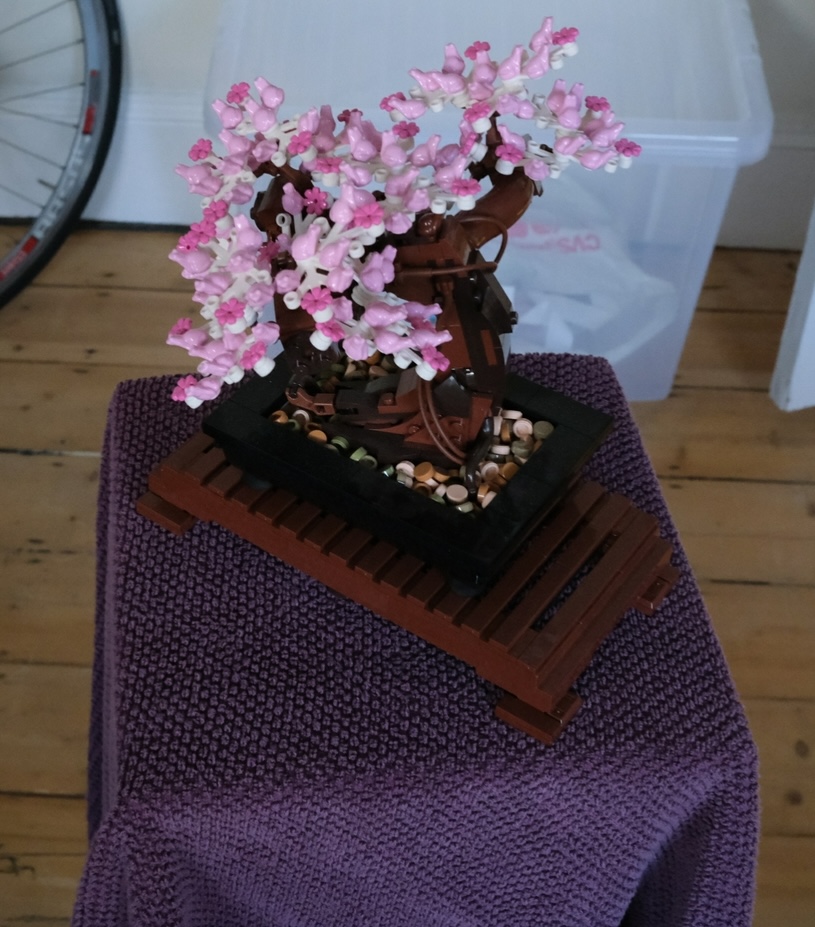}\hfill
    \includegraphics[width=0.145\linewidth]{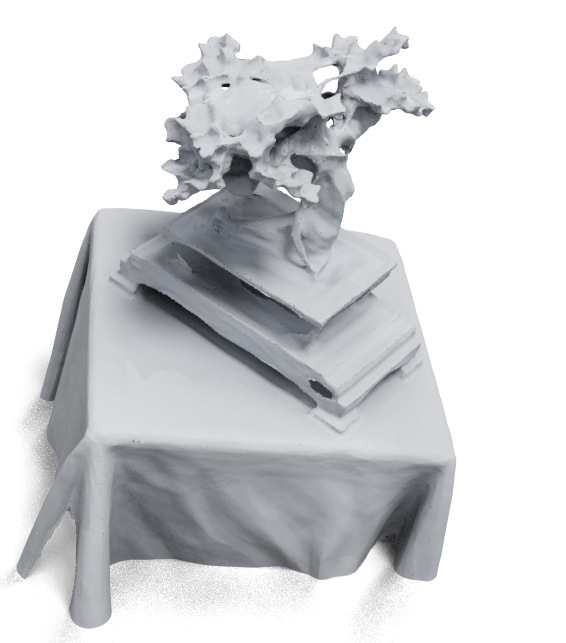}\hfill
    \includegraphics[width=0.145\linewidth]{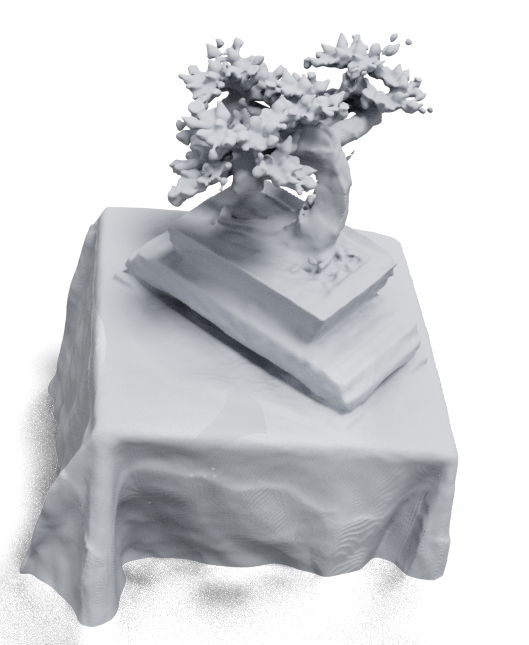}\hfill
    \includegraphics[width=0.165\linewidth]{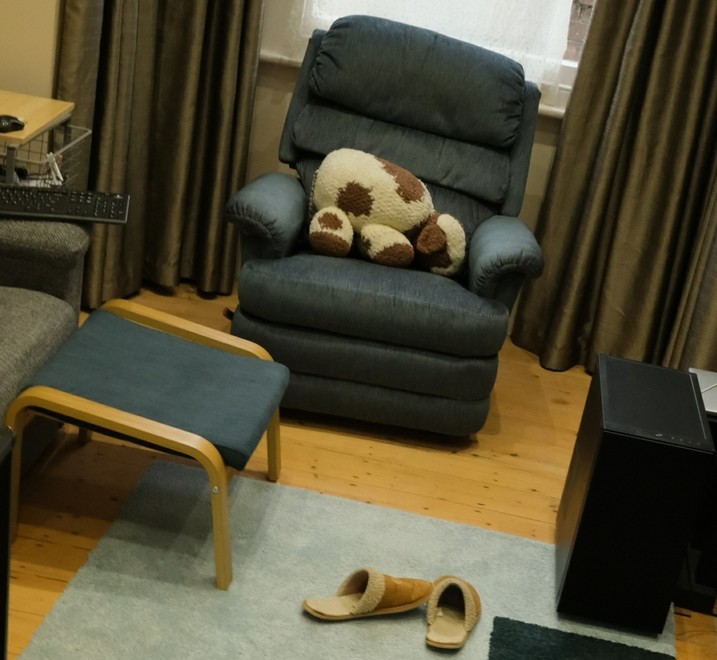}\hfill
    \includegraphics[width=0.165\linewidth]{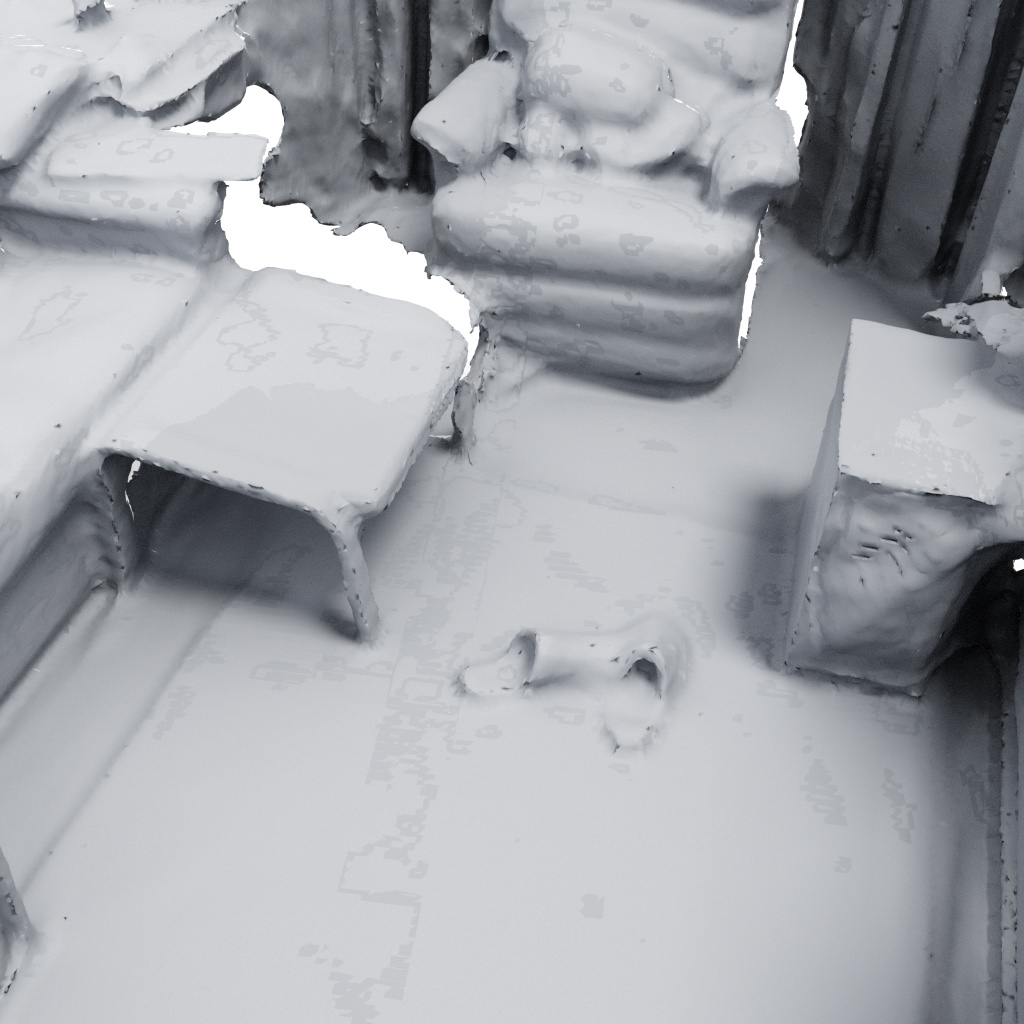}\hfill
    \includegraphics[width=0.165\linewidth]{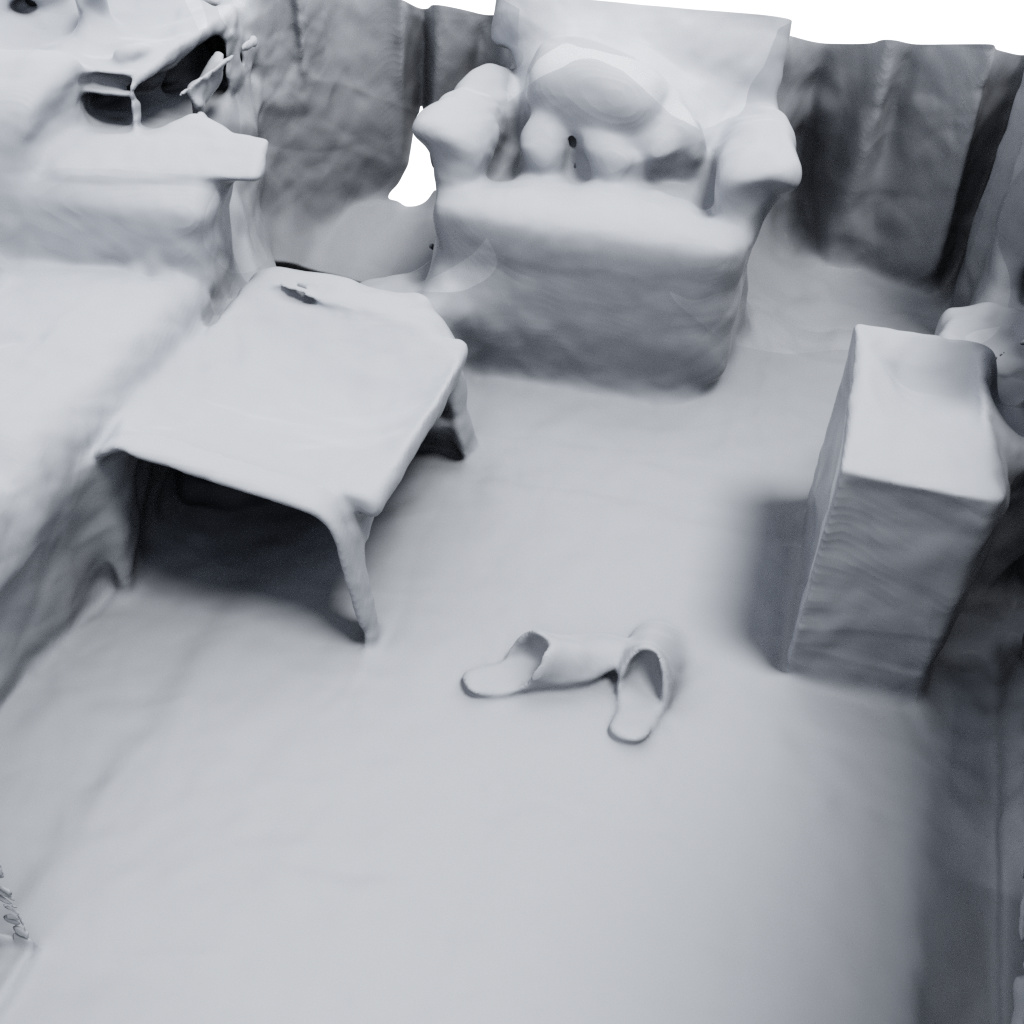}\\
    
    \vspace{3mm} %
    \makebox[0.145\linewidth]{\small Reference} \hfill
    \makebox[0.145\linewidth]{\small GS-Pull} \hfill
    \makebox[0.145\linewidth]{\small Ours} \hfill
    \makebox[0.165\linewidth]{\small Reference} \hfill
    \makebox[0.165\linewidth]{\small GS-Pull} \hfill
    \makebox[0.165\linewidth]{\small Ours} \\[0.8mm]
    
    \includegraphics[width=0.145\linewidth]{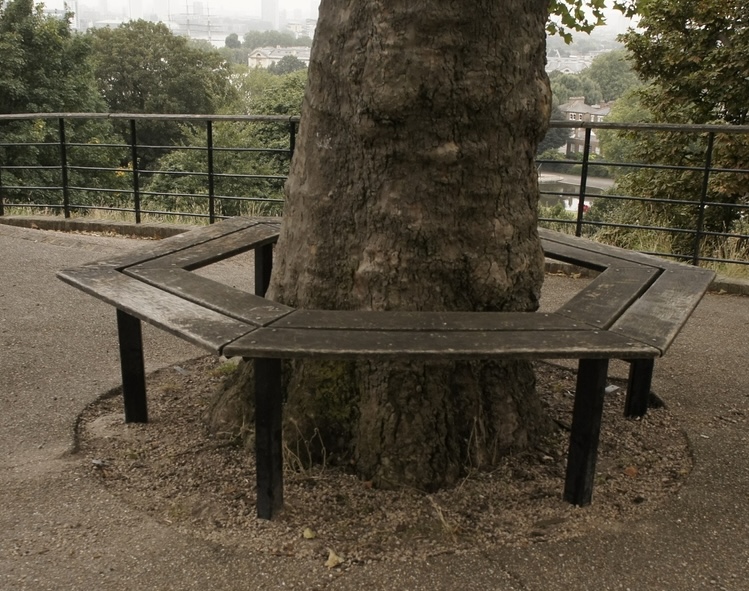}\hfill
    \includegraphics[width=0.145\linewidth]{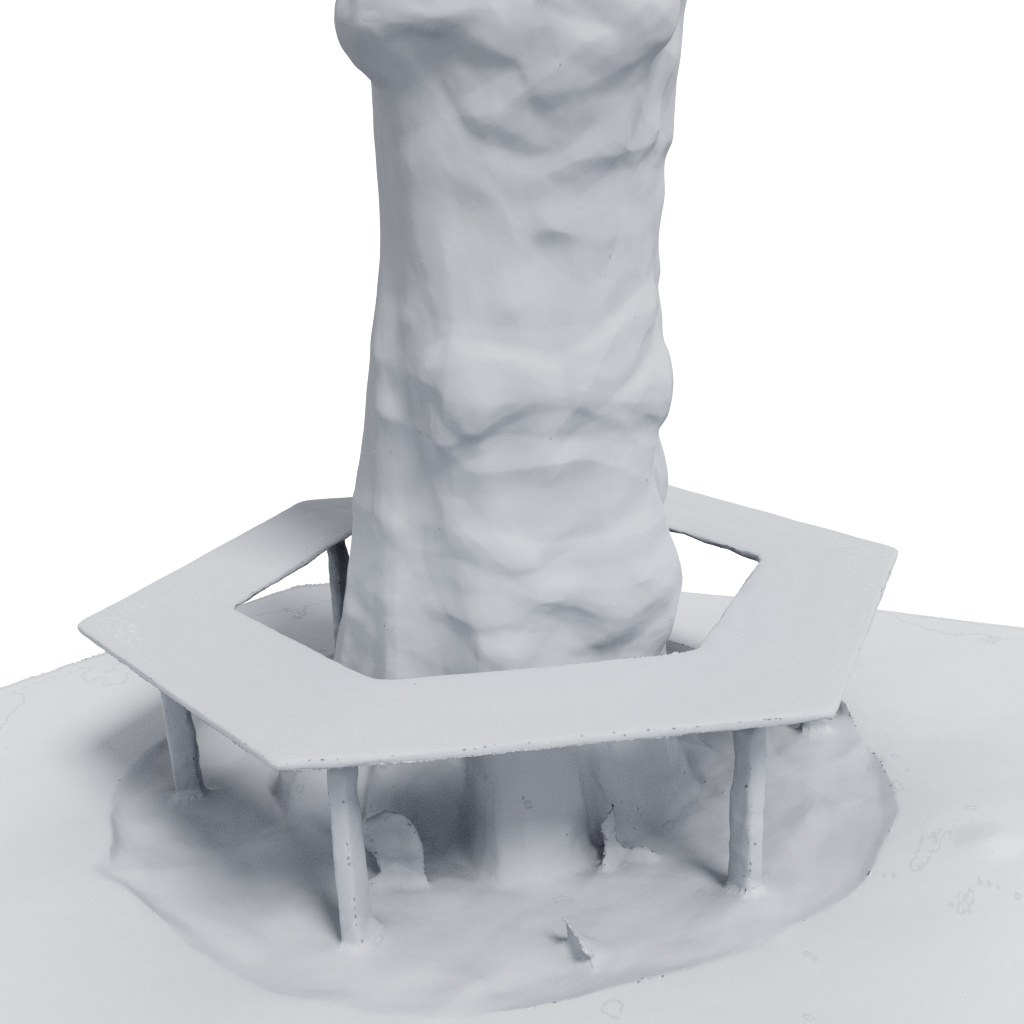}\hfill
    \includegraphics[width=0.145\linewidth]{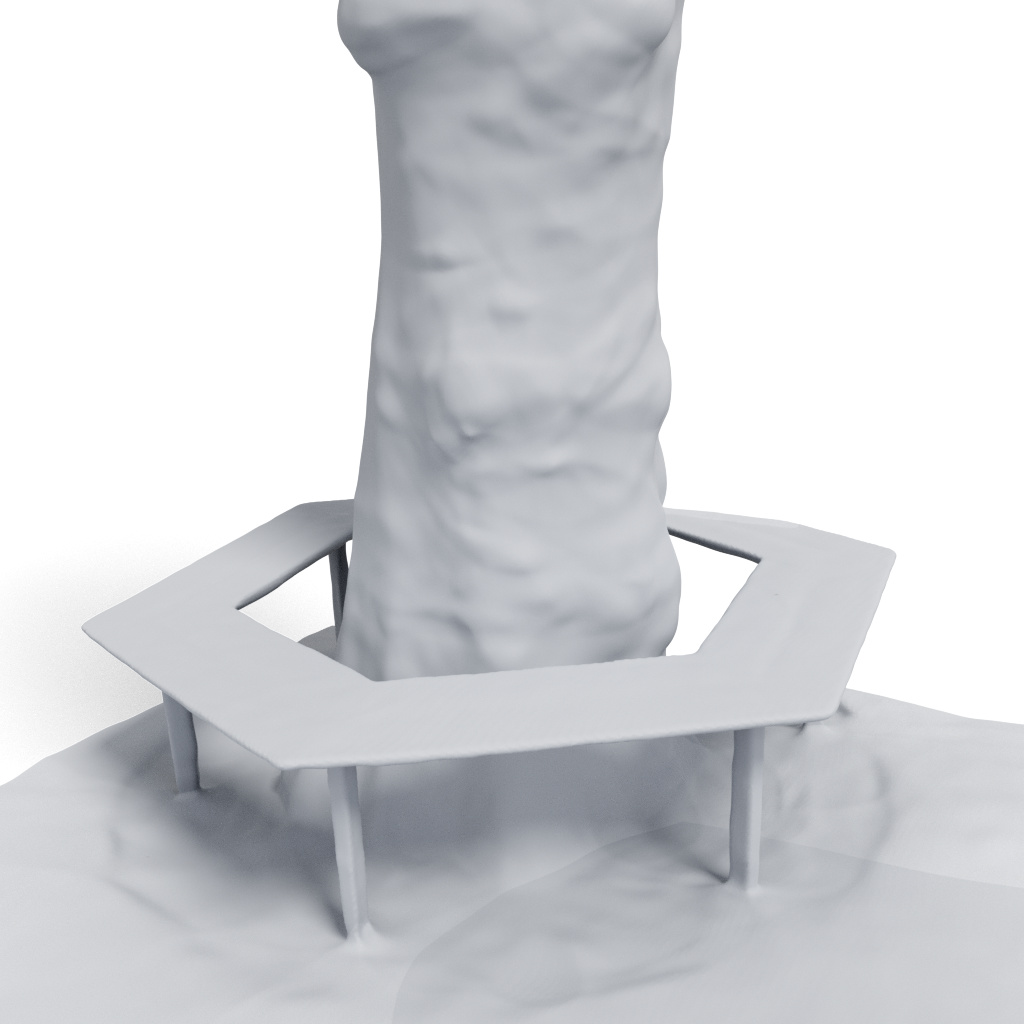}\hfill
    \includegraphics[width=0.165\linewidth]{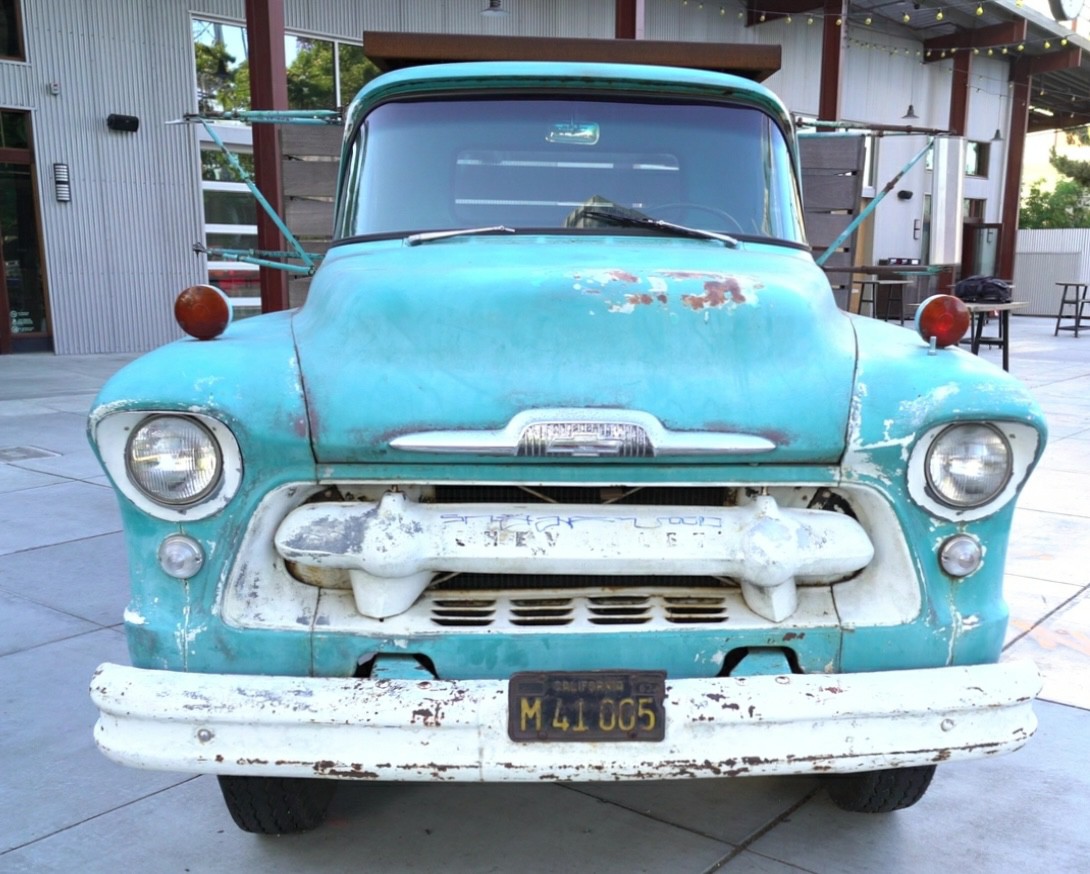}\hfill
    \includegraphics[width=0.165\linewidth]{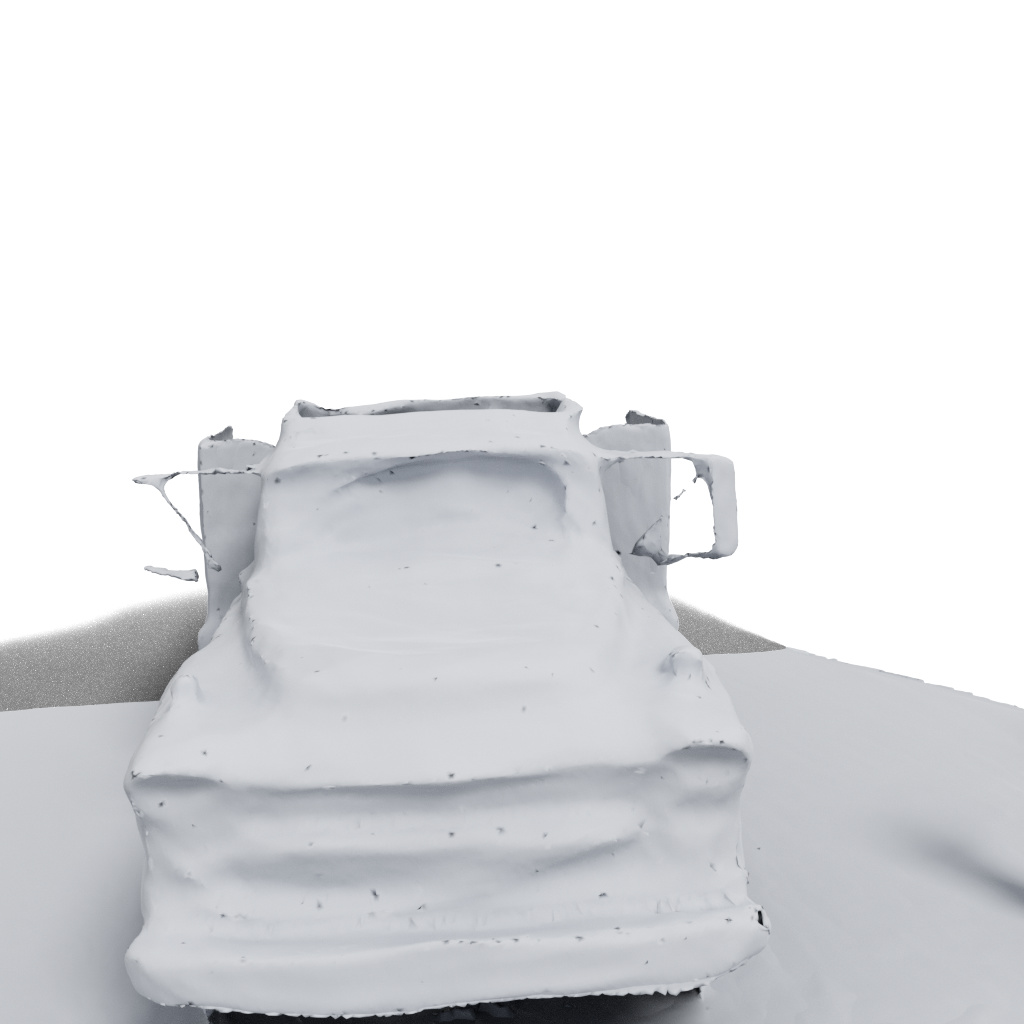}\hfill
    \includegraphics[width=0.165\linewidth]{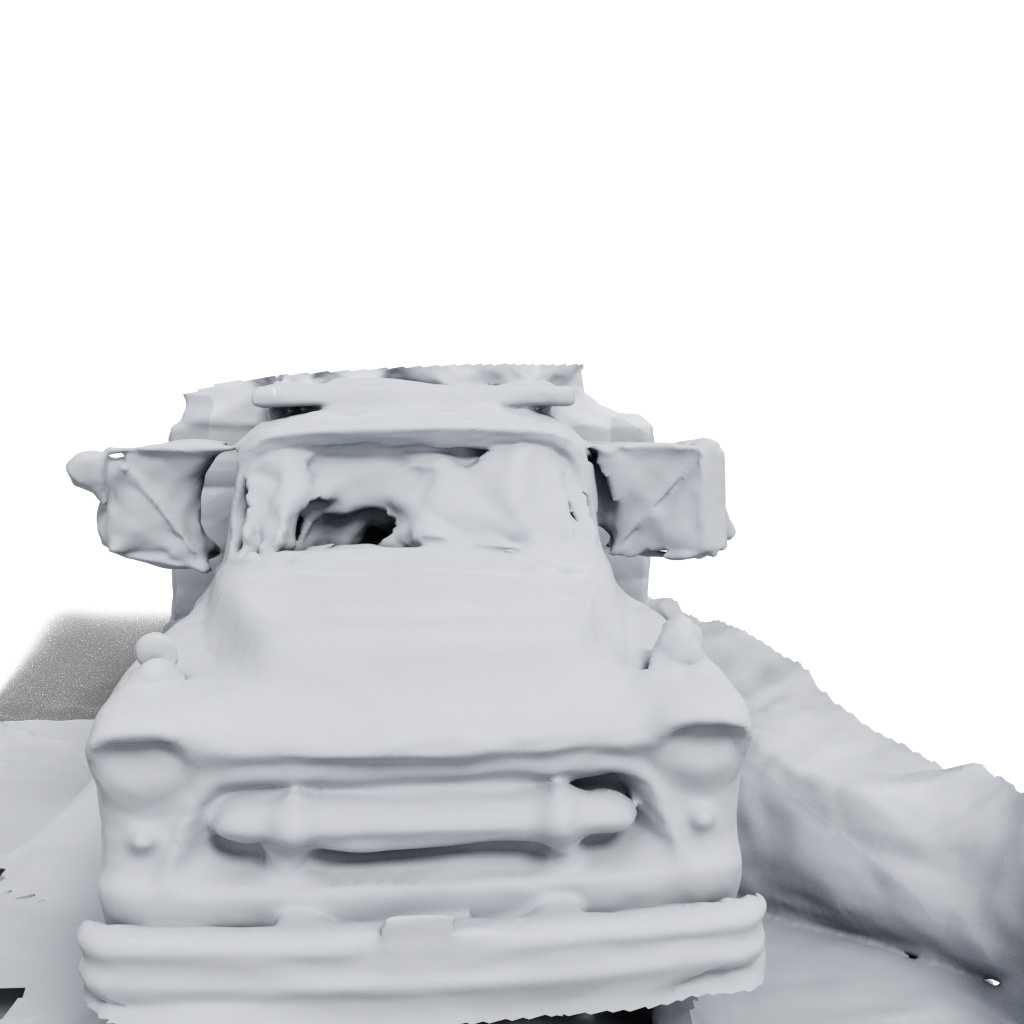}\\[0.8mm]
    \caption{Visual comparison with GS-Pull~\cite{zhang2024gspull} on the MipNeRF 360~\cite{barron2022mipnerf360} and Tank \& Temple ~\cite{knapitsch2017tanks} datasets, featuring scene-level data.}
    \label{fig:tnt}
\end{figure*}
\paragraph{Scene-level reconstructions}
For indoor and outdoor scene data, our method captures fine structural details, such as the bonsai in \Cref{fig:tnt}, and produces more coherent and less fragmented surfaces. Compared to the state-of-the-art GS-Pull~\cite{zhang2024gspull}, our approach offers improved geometric fidelity and greater robustness.

\begin{figure*}[!htbp]
    \centering
    \includegraphics[width=0.13\linewidth]{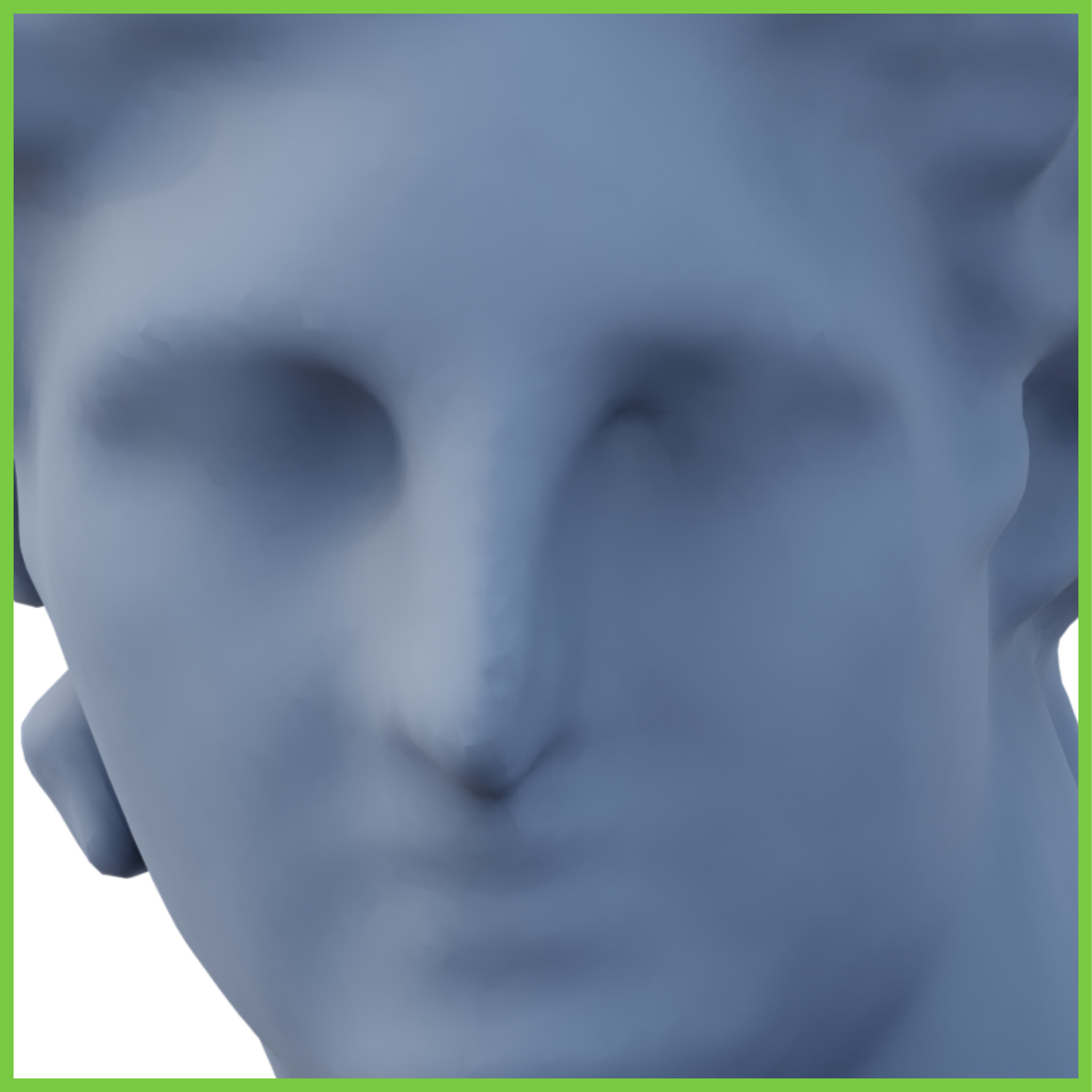}
    \includegraphics[width=0.13\linewidth]{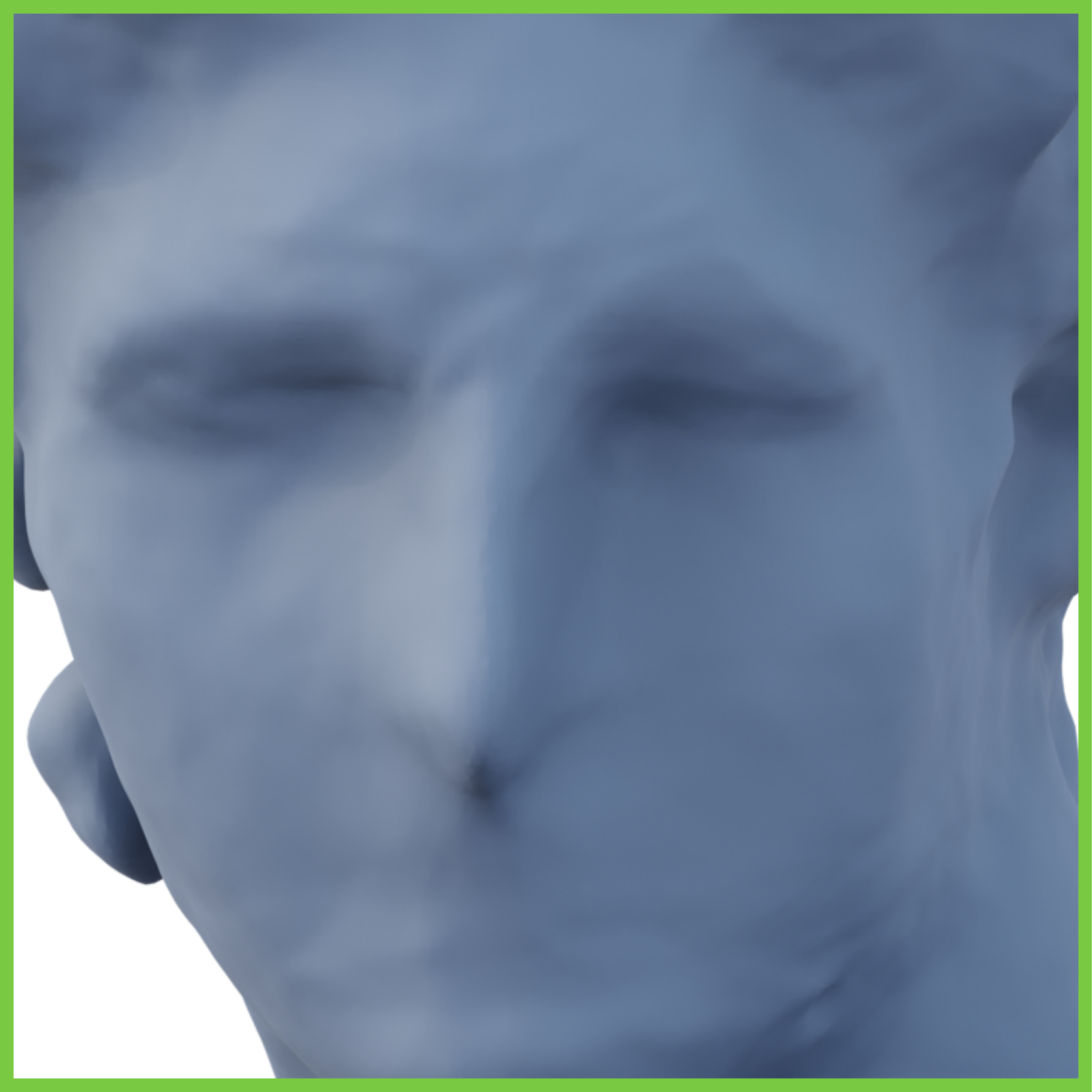}
    \includegraphics[width=0.13\linewidth]{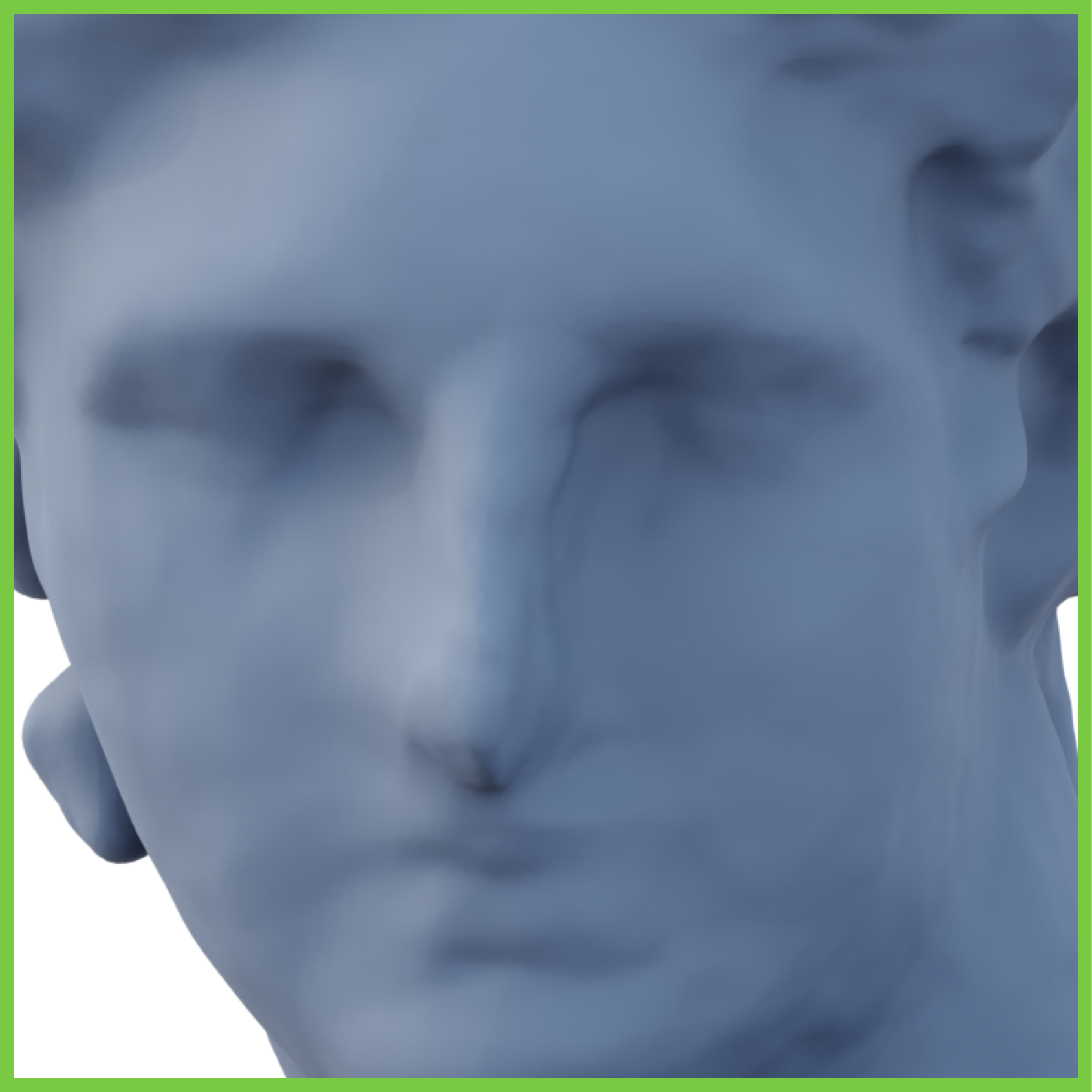}
    \includegraphics[width=0.13\linewidth]{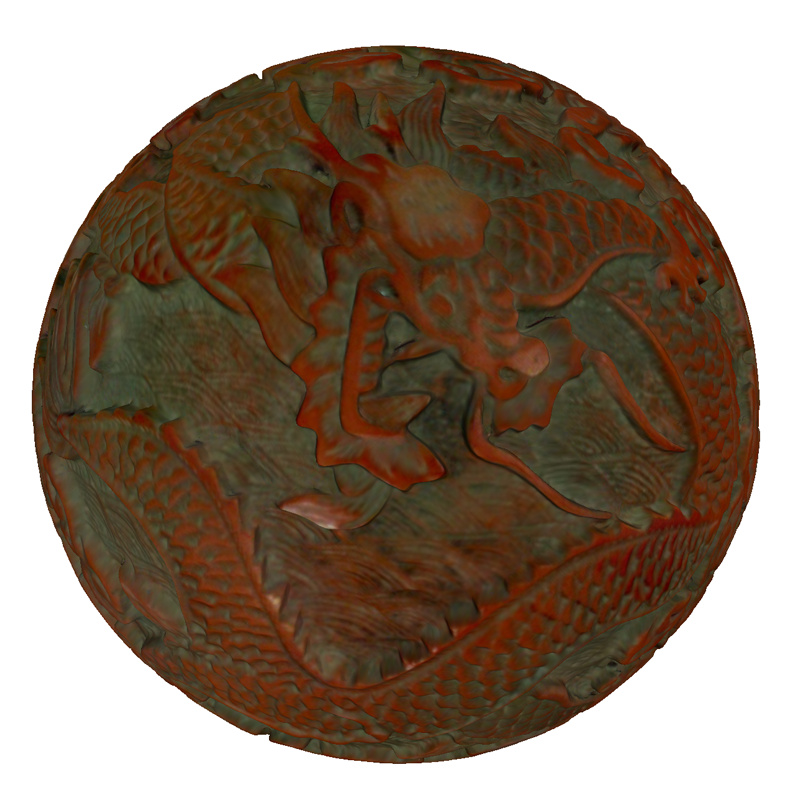}
    \includegraphics[width=0.13\linewidth]{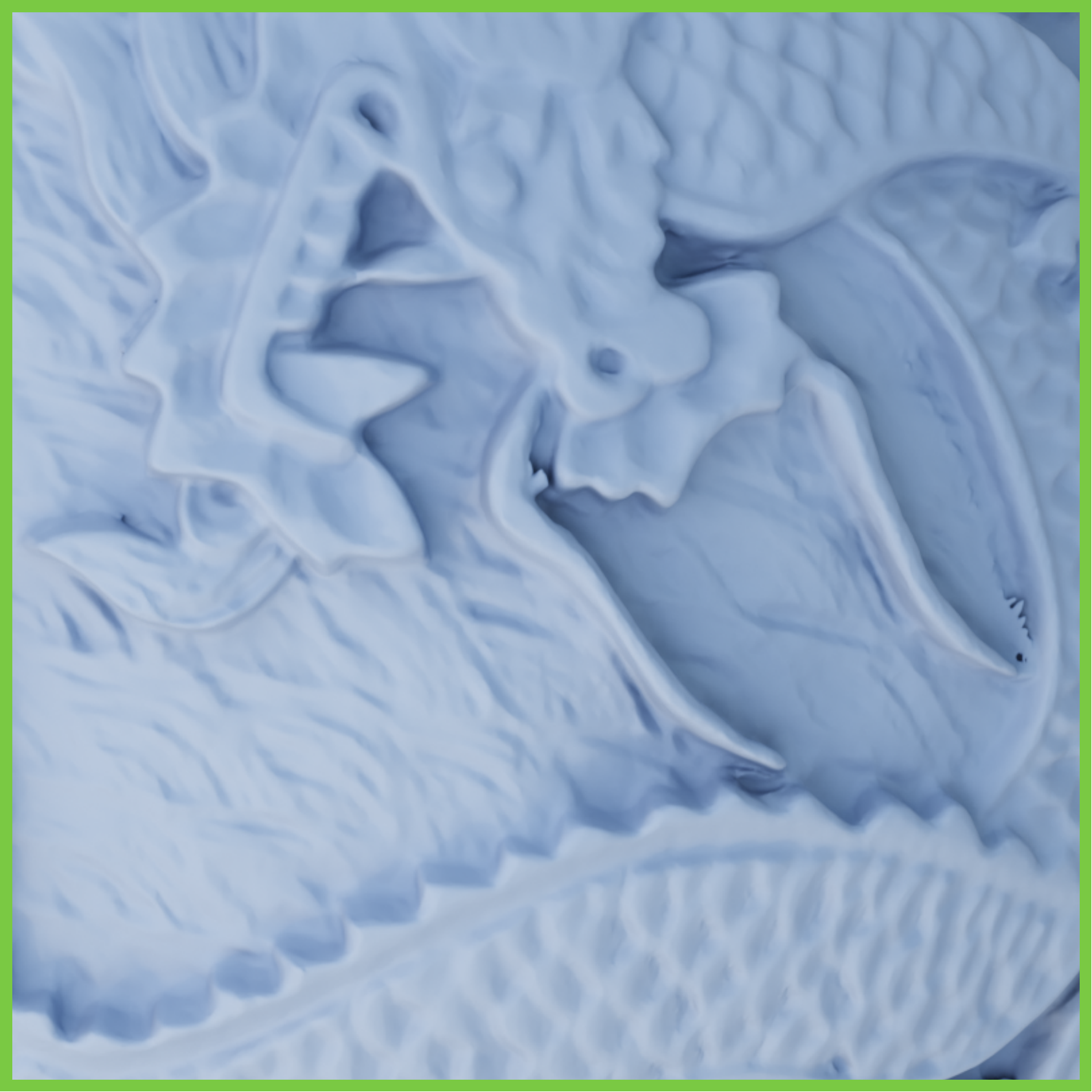}
    \includegraphics[width=0.13\linewidth]{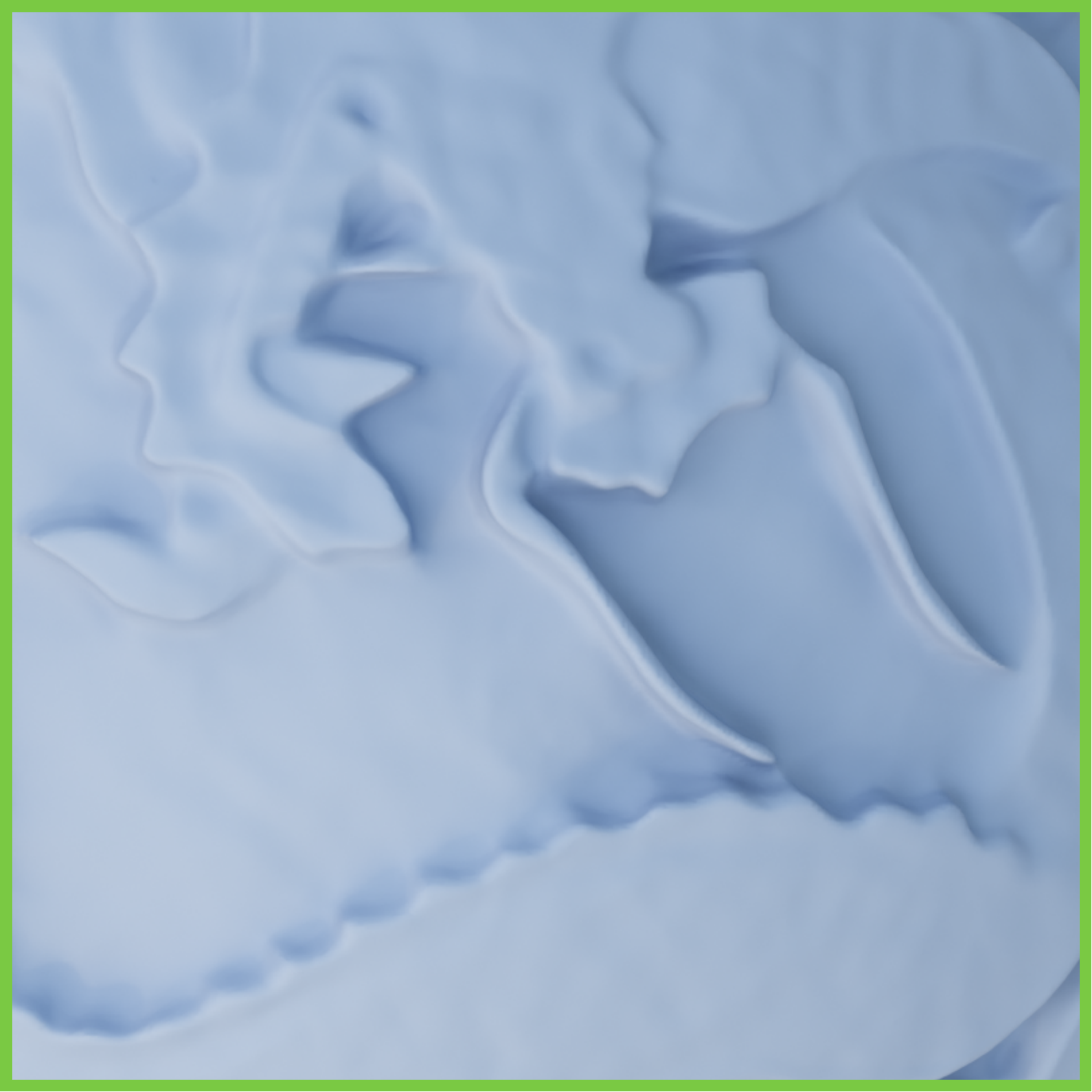}
    \includegraphics[width=0.13\linewidth]{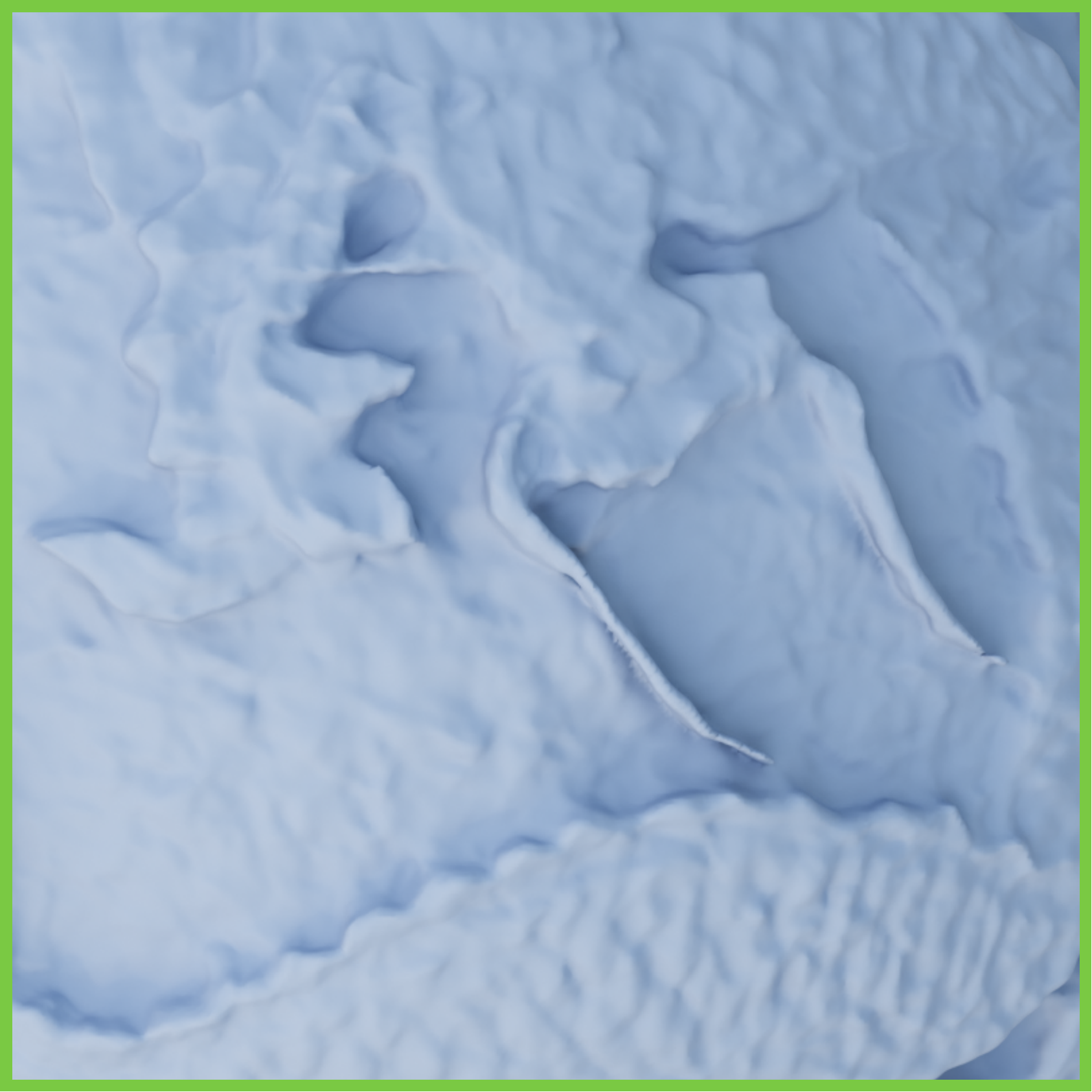}
    \\
    \includegraphics[width=0.13\linewidth]{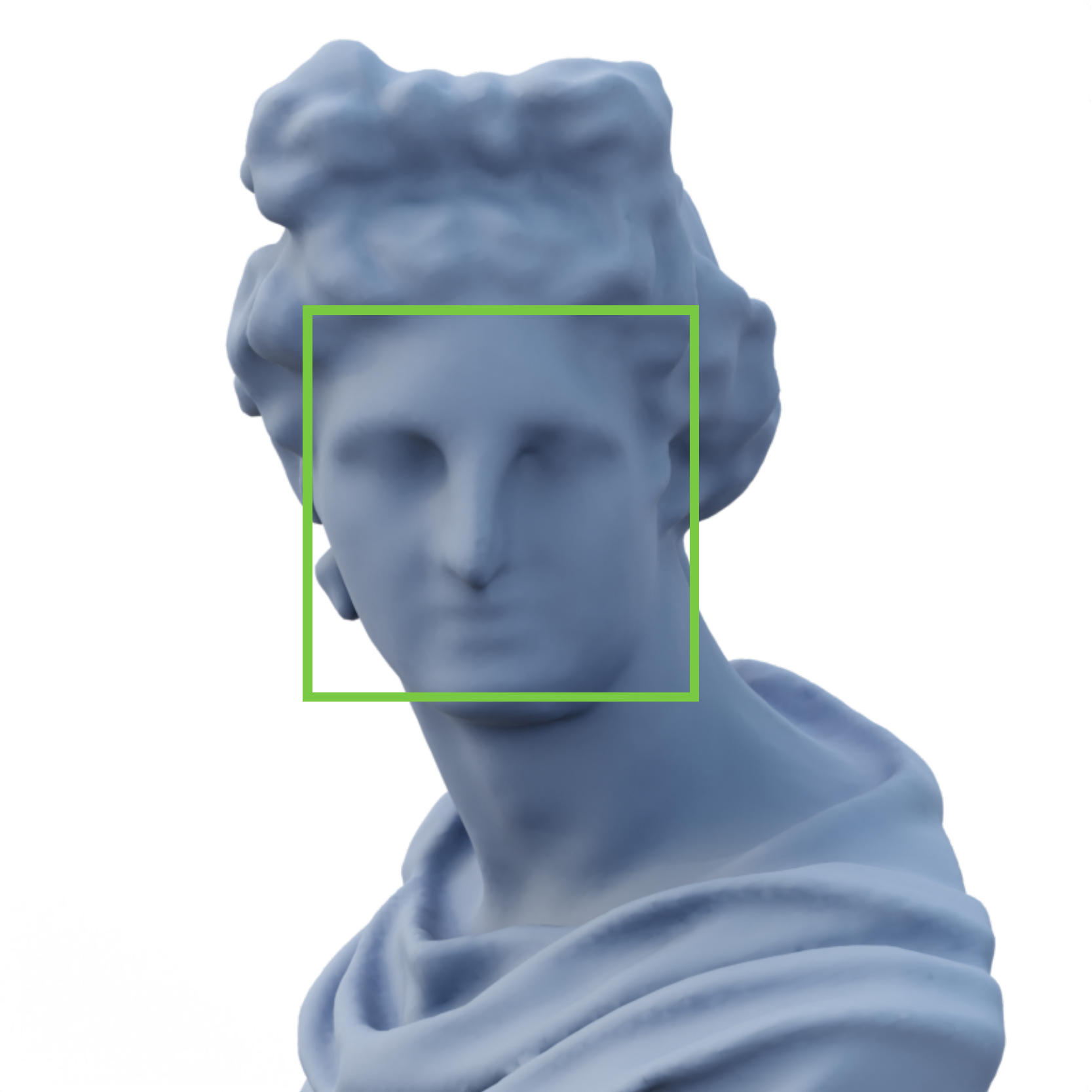}
    \includegraphics[width=0.13\linewidth]{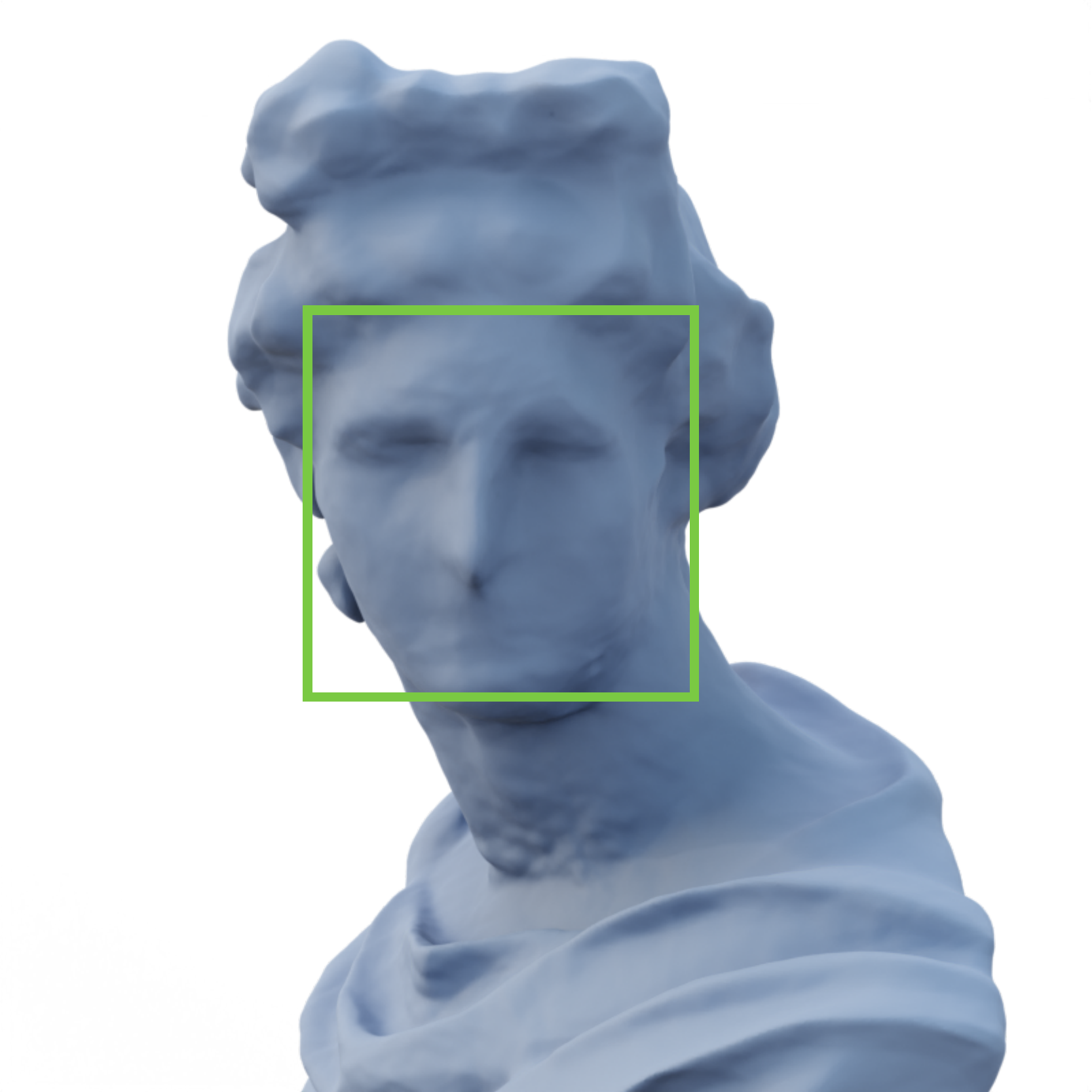}
    \includegraphics[width=0.13\linewidth]{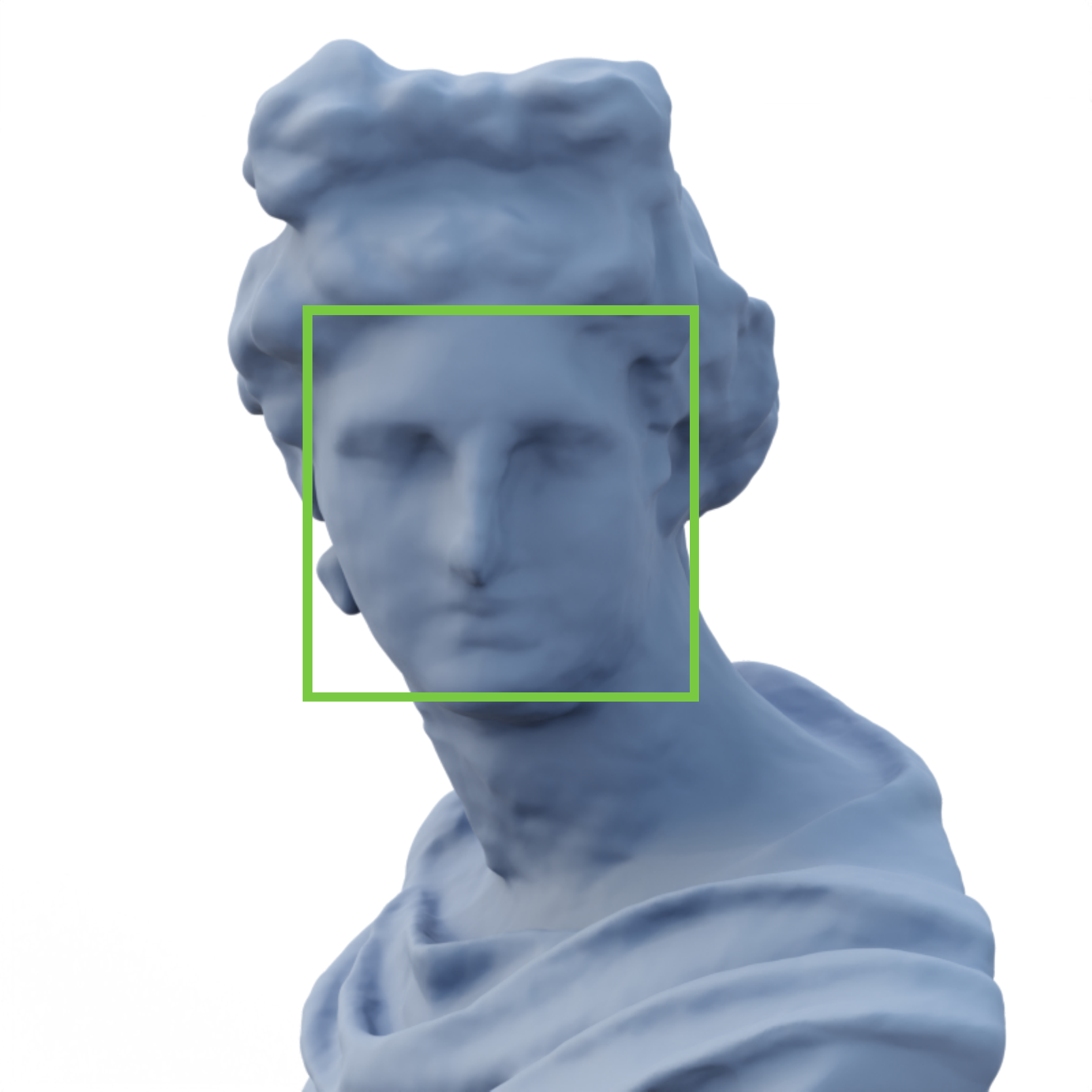}
    \includegraphics[width=0.13\linewidth]{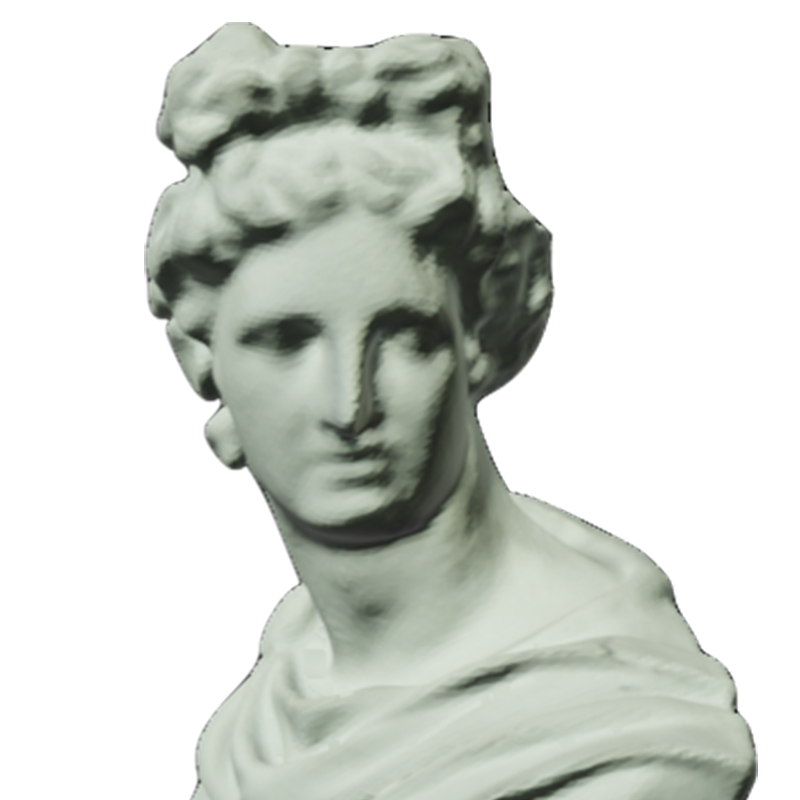}
    \includegraphics[width=0.13\linewidth]{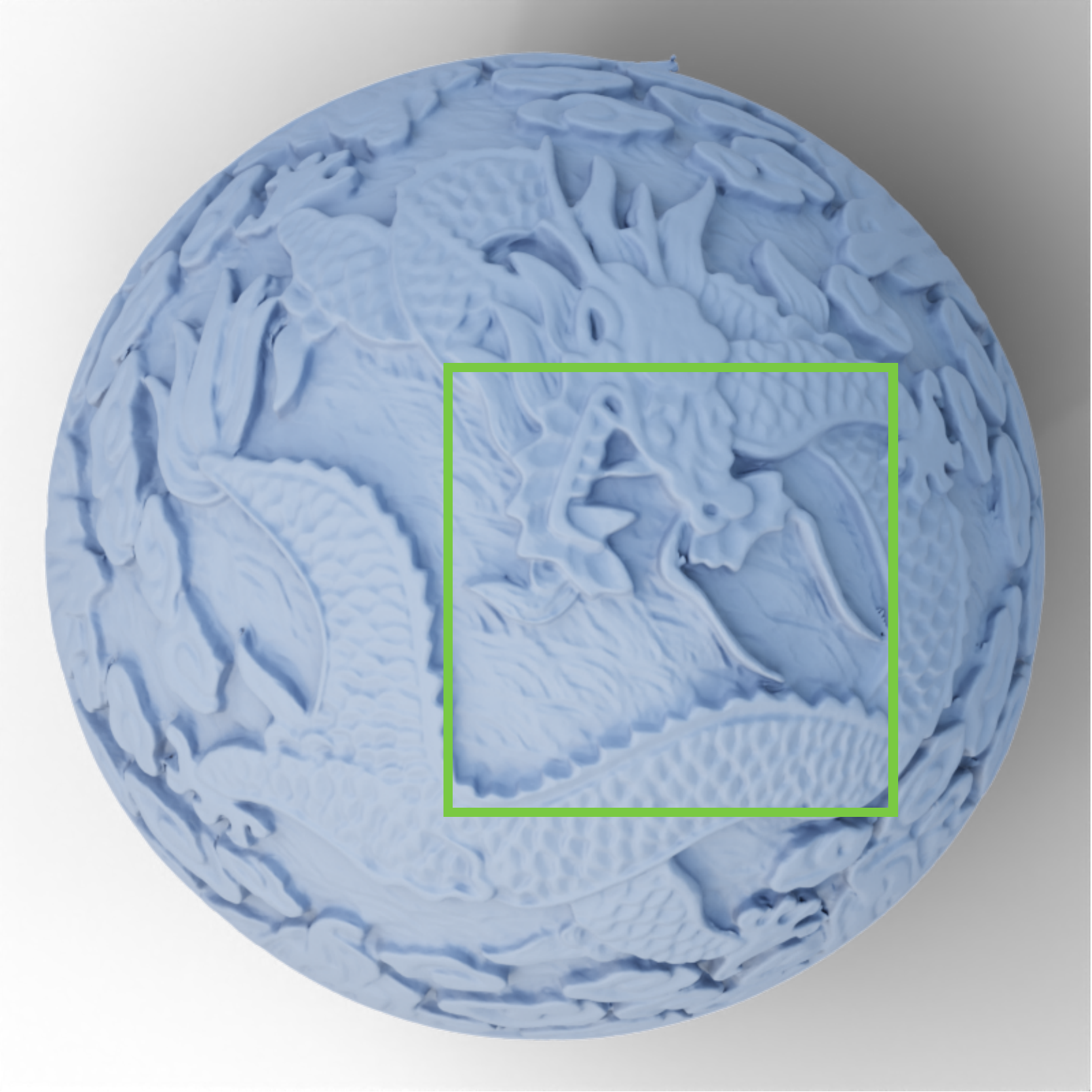}
    \includegraphics[width=0.13\linewidth]{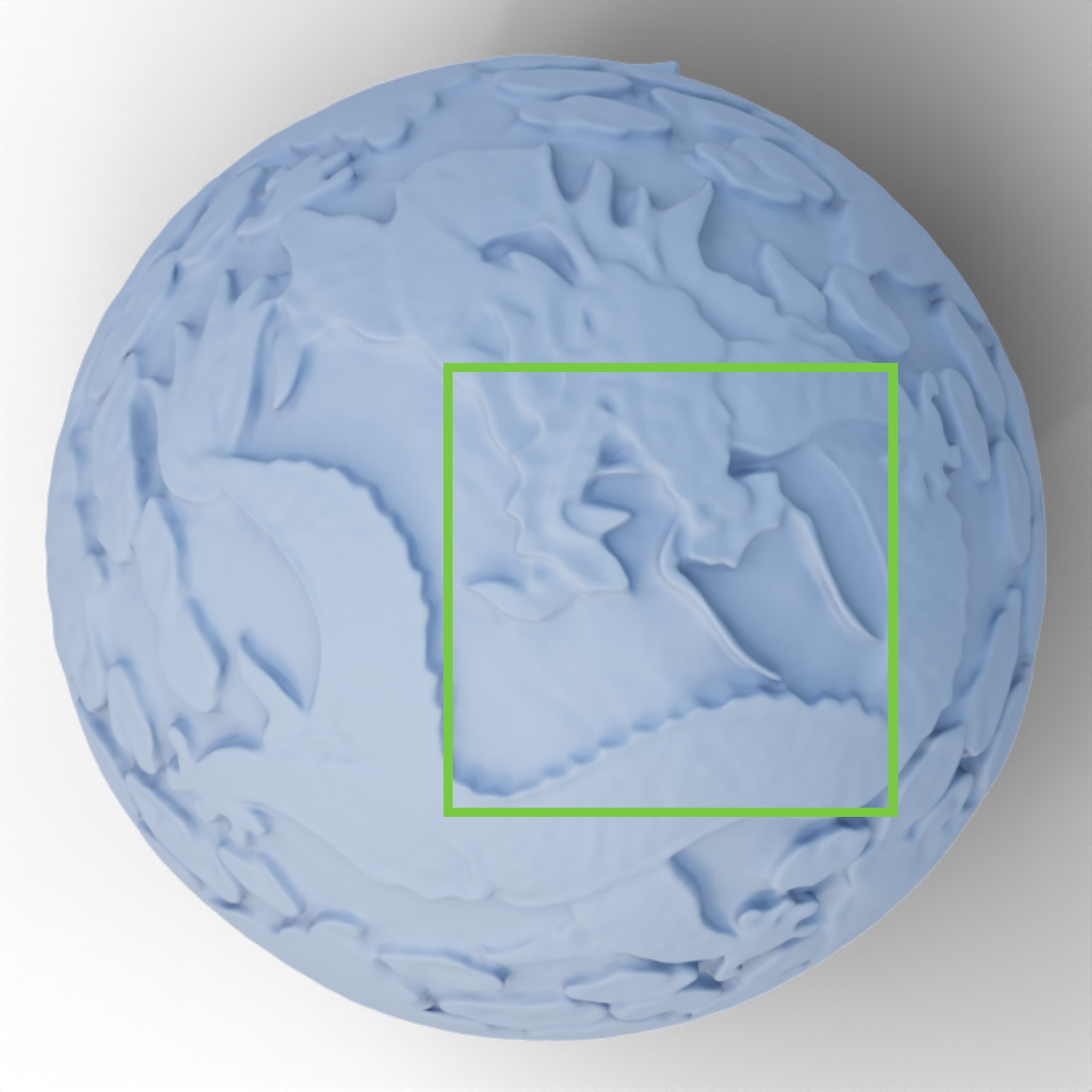}
    \includegraphics[width=0.13\linewidth]{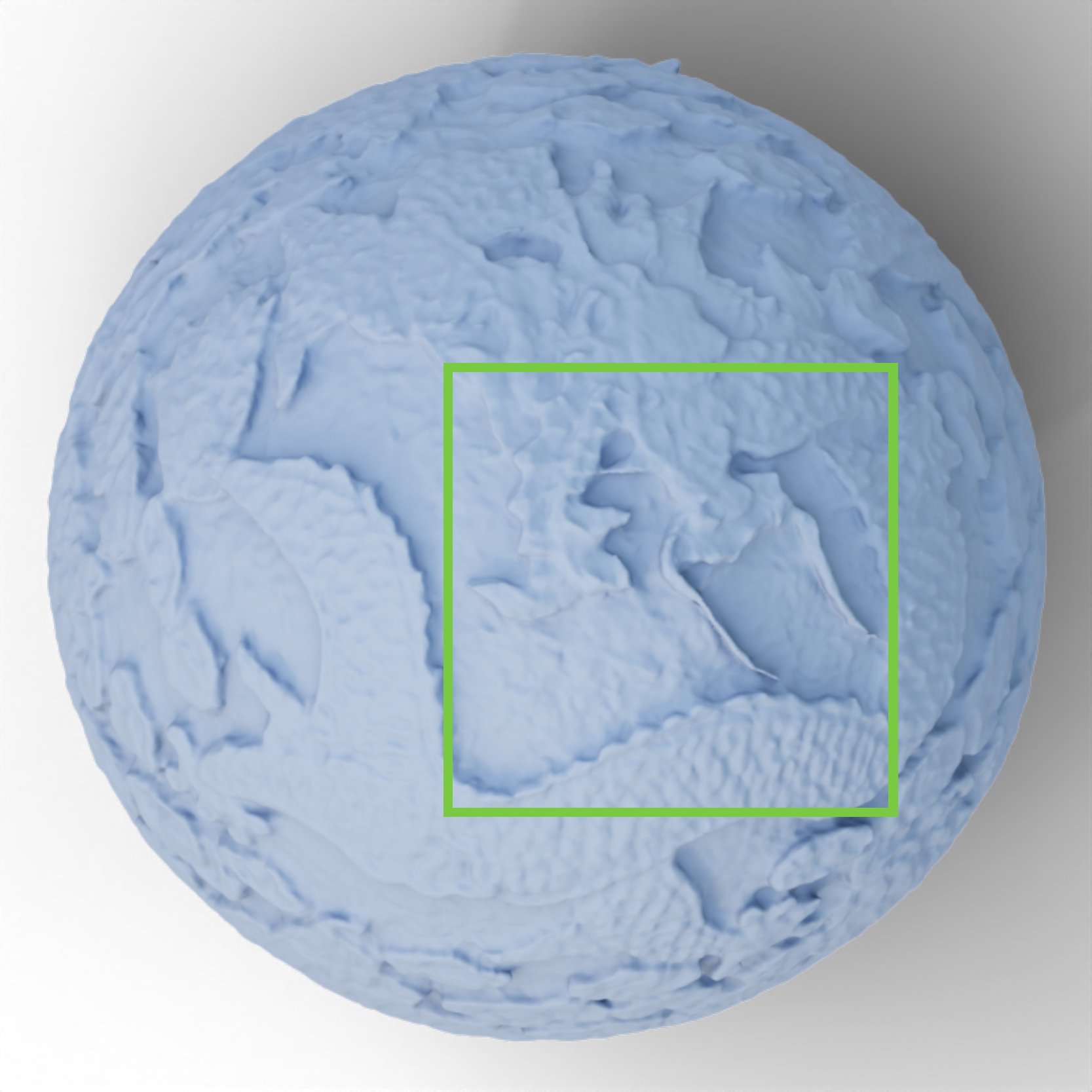}
    \\
    \makebox[0.13\linewidth]{\scriptsize GT}
    \makebox[0.13\linewidth]{\scriptsize w/o GAM}
    \makebox[0.13\linewidth]{\scriptsize w/ GAM}
    \makebox[0.13\linewidth]{\scriptsize Reference}
    \makebox[0.13\linewidth]{\scriptsize GT}
    \makebox[0.13\linewidth]{\scriptsize w/o GAM}
    \makebox[0.13\linewidth]{\scriptsize w/ GAM}
    \\  
    \caption{Ablation study on geometry-guided appearance modeling.}
    \label{fig:aba-rgb}
\end{figure*}
\subsection{Ablation Studies}
\label{subsec:abalation}
\paragraph{Opacity regularization} Low-opacity Gaussians contribute little to rendering, but severely disrupt the SDF module, leading to distorted and inaccurate geometry. Therefore, reducing the number of low-opacity Gaussians is crucial for effective integration of SDF and Gaussian splatting. As shown in~\Cref{tab:oo3d-d} and~\Cref{fig:oo3dd}, our entropy-based opacity regularization reduces the total number of Gaussians (from 58,148 to 44,827) and increases their opacity, encouraging better alignment with the SDF's zero-level set.

Without opacity regularization, the centroids of Gaussians become unreliable and tend to shift slightly during training, introducing instability and noise in the SDF learning process. This results in a higher CD when the opacity is unconstrained, as shown in~\Cref{tab:oo3d-d}.

\paragraph{Geometry-guided appearance modeling} Removing geometric features from appearance modeling leads to a slight degradation in both geometry reconstruction and rendering quality, as shown in Tables~\ref{tab:oo3d-d} and~\ref{tab:aba_render}. \Cref{fig:aba-rgb} shows that spherical harmonics fail to capture fine geometric details. By incorporating geometry features and surface normals, GSurf improves geometry reconstruction by using richer structural cues. This also improves rendering quality, particularly for reflective surfaces, as demonstrated in~\Cref{fig:trans}.
\begin{table}[htbp]
    \centering
    \small
    \caption{ Ablation studies on OmniObject3D. OR and GAM
stand for opacity regularization and geometry-guided appearance modeling.}  
\resizebox{\linewidth}{!}{
    \begin{tabular}{c|c|c|c|c|c}
    \toprule
         & Ours(full) & w/o GAM & w/o OR  & 2DGS & GOF \\
         \midrule
    PSNR $\uparrow$   & 40.06 & 39.19 & 37.12 &39.24 & 40.02\\
    SSIM $\uparrow$  & 0.986  & 0.984 & 0.972 &0.982 & 0.983 \\
    LPIPS $\downarrow$ & 0.025 & 0.031 & 0.039 &0.027 &  0.026\\
    \bottomrule
    \end{tabular}
    }
    \label{tab:aba_render}
\end{table}

\begin{figure}
    \centering
    \includegraphics[width=0.95\linewidth]{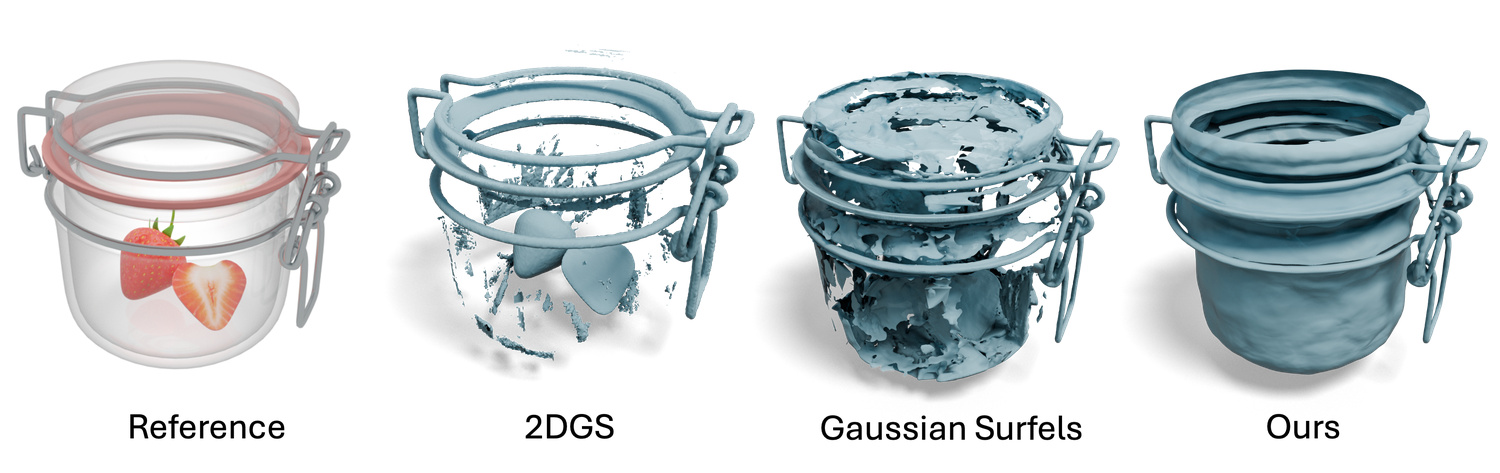}
    \caption{Visual comparison of surface extraction on a semi-transparent object. Accurate depth estimation on refractive glass is highly ill-posed. 2DGS completely fails to reconstruct the glass envelope, capturing only fragments of the internal strawberries. GaussianSurfels extracts a chaotic, fragmented shell. Due to our binarization strategy ($o_k \to 1$), GSurf introduces a geometric bias that robustly extracts a clean, structurally sound, opaque outermost envelope.}
    \label{fig:semi_jar}
\end{figure}
\paragraph{Limitations} We acknowledge the limitations of our method when reconstructing scenes with genuinely translucent structures (e.g., the glass jar). By mathematically forcing the opacity $o_k \to 1$ via entropy regularization, our method treats semi-transparent volumes as equivalent opaque outer shells. To minimize the photometric loss, the geometry-conditioned appearance MLP is forced to ``bake'' the visual information of internal volumetric transmissions (e.g., the strawberries inside the jar) into high-frequency, view-dependent textures directly onto this solid boundary.

As a result, suppressing semi-transparent Gaussians introduces a geometric bias: the method does not model internal structures and depth correctly. However, as shown in \Cref{fig:semi_jar}, accurately reconstructing transparent refractive objects remains a significant challenge for standard splatting-based surface extraction pipelines. Because multi-layered refractive ray paths severely disrupt linear depth estimation, baselines such as 2DGS completely miss the glass envelope (only partially capturing internal items), while GaussianSurfels produces severely fragmented, noisy geometry. In this highly ill-posed context, GSurf provides a predictable and structurally advantageous failure mode by robustly outputting a clean, outermost envelope.

\section{Conclusion}
\label{sec:conclusion}

We presented GSurf, a single-branch framework that couples continuous SDF learning with discrete Gaussian splatting for efficient multi-view surface reconstruction. By supervising the SDF directly from Gaussian centroids, regularizing opacity to remove low-confidence primitives, and using SDF-derived features for appearance modeling, GSurf improves surface completeness and fine geometric detail without requiring an additional volume-rendering branch. Experiments on object-level and scene-level datasets demonstrate that the method is competitive with, and often improves upon, existing GS-based reconstruction approaches, especially in challenging cases involving strong lighting, reflective materials, and semi-transparent objects. Moreover, GSurf is computationally more efficient than existing GS+SDF approaches, requiring less training time and producing fewer Gaussians while maintaining or exceeding reconstruction quality. These results validate the effectiveness of bridging continuous and discrete representations and highlight the potential of hybrid, multi-level frameworks for 3D reconstruction in real-world applications.

\section*{Acknowledgment} This project was partially supported by the Ministry of Education, Singapore, under its Academic Research Fund Grant (RT19/22).

{
    \small
    \bibliographystyle{ieeenat_fullname}
    \bibliography{main}
}

\end{document}